\documentclass[twoside,11pt]{article}
\usepackage{jair, theapa, rawfonts}
\usepackage{amsmath}
\usepackage{amssymb}
\usepackage{IEEEtrantools}
\usepackage{subfigure}
\usepackage{tikz}
\usetikzlibrary{shapes.symbols}
\usepackage{fontawesome}
\usepackage{multirow}
\usepackage{bbm}
\usepackage{amsthm}
\usepackage[utf8]{inputenc}
\usepackage[T1]{fontenc}

\newcommand{\dd}{\mathop{}\!\mathrm{d}}

\usepackage{algorithm}
\usepackage{algorithmic}

\jairheading{79}{2024}{515-567}{06/2023}{02/2024}
\ShortHeadings{Kernel Affine Hull Machine}
{Kumar, Moser, \& Fischer}

\firstpageno{515}

\newtheorem{problem}{Problem}
\newtheorem{theorem}{Theorem}
\newtheorem{definition}{Definition}
\newtheorem{remark}{Remark}
\newtheorem{result}{Result}
\begin{document}

\title{On Mitigating the Utility-Loss in Differentially Private Learning:  A New Perspective by a Geometrically Inspired Kernel Approach}

\author{\name Mohit Kumar \email mohit.kumar@uni-rostock.de \\
       \addr Faculty of Computer Science and Electrical Engineering \\
       University of Rostock, Germany\\
        and\\
       \addr Software Competence Center Hagenberg GmbH\\   
       A-4232 Hagenberg, Austria
       \AND
       \name Bernhard A. Moser \email bernhard.moser@scch.at \\
         \addr Institute of Signal Processing \\
         Johannes Kepler University Linz, Austria \\
                 and\\
                \addr Software Competence Center Hagenberg GmbH\\
                 A-4232 Hagenberg, Austria
         \AND
         \name Lukas Fischer \email lukas.fischer@scch.at  \\
         \addr  Software Competence Center Hagenberg GmbH\\
                 A-4232 Hagenberg, Austria
}


\maketitle

\begin{abstract}
Privacy-utility tradeoff remains as one of the fundamental issues of differentially private machine learning. This paper introduces a geometrically inspired kernel-based approach to mitigate the accuracy-loss issue in classification. In this approach, a representation of the affine hull of given data points is learned in Reproducing Kernel Hilbert Spaces (RKHS). This leads to a novel distance measure that hides privacy-sensitive information about individual data points and improves the privacy-utility tradeoff via significantly reducing the risk of membership inference attacks. The effectiveness of the approach is demonstrated through experiments on MNIST dataset, Freiburg groceries dataset, and a real biomedical dataset. It is verified that the approach remains computationally practical. The application of the approach to federated learning is considered and it is observed that the accuracy-loss due to data being distributed is either marginal or not significantly high.
\end{abstract}

\section{Introduction}
\label{Introduction}

 Privacy-preserving machine learning is the central topic of this study. Differential privacy~\cite{DBLP:journals/fttcs/DworkR14} is a standard framework to quantify the degree to which the data privacy of each individual in the dataset is preserved while releasing the output of any statistical analysis algorithm. Differential privacy, being a property of an algorithm's data access mechanism, automatically provides protection against arbitrary privacy-leakage risks. The goal of protecting sensitive information (that is embedded in training data) from any leakage through machine learning models has been addressed within the framework of differential privacy~\shortcite{Abadi:2016:DLD:2976749.2978318,Phan:2016:DPP:3015812.3016005}. The classical approach for designing differentially private algorithms is {\em output perturbation}, where the idea is to perturb the function output via adding noise calibrated to the global \emph{sensitivity} of the function~\shortcite{10.1007/11681878_14}. A common form of output perturbation is the {\em Gaussian mechanism}, where Gaussian noise calibrated to the $L2$ sensitivity is added. Differential privacy has been defined for functions and functional data~\shortcite{10.5555/2567709.2502603}. Specifically for functions in RKHS generated by the covariance kernel of the Gaussian process, the correct noise level is established by the sensitivity of the function in the RKHS norm~\cite{10.5555/2567709.2502603}. The iterative nature of machine learning algorithms causes a high cumulative privacy loss and thus a high amount of noise need to be added to compensate for the privacy loss. A \emph{moments accountant} method~\cite{Abadi:2016:DLD:2976749.2978318}, based on the properties of a \emph{privacy loss} random variable, has been suggested to keep track of the privacy loss incurred by successive iterations for composition analysis. The moments accountant method can be combined with the use of privacy amplification effect of subsampling to deal with the iterative algorithms~\shortcite{article}.

An obvious effect of adding noise into an algorithm for preserving differential privacy is the loss in algorithm's accuracy. As differential privacy remains immune to any post-processing of released output, the output data can be denoised using statistical estimation theory~\cite{DBLP:conf/icml/BalleW18}. It is not surprising that efforts have been made to optimize the privacy-accuracy tradeoff~\shortcite{Geng2018OptimalNM,DBLP:conf/icml/BalleW18,doi:10.1137/09076828X,Gupte:2010:UOP:1807085.1807105,7345591,7093132,7353177}. Previously, the studies~\shortcite{Kumar/IWCFS2019,KUMAR202187} have derived the probability density function of noise that minimizes the expected noise magnitude together with satisfying the sufficient conditions for $(\epsilon,\delta)-$differential privacy. Given $N$ number of $p-$variate data points (represented by a matrix $Y \in \mathbb{R}^{N \times p}$), any computational algorithm  operating on the data matrix $Y$ can be represented by a mapping, $alg: \mathbb{R}^{N \times p} \rightarrow Range(alg)$. The input perturbation method achieves the $(\epsilon,\delta)-$differential privacy of $alg$ via adding a random noise matrix $V \in \mathbb{R}^{N \times p}$ to $Y$ such that the following inequality holds good:         
 \begin{IEEEeqnarray}{rCl}
\label{eq_260520231423} Pr\{ alg(Y+V) \in \mathcal{O} \} & \leq & \exp(\epsilon) Pr\{ alg(Y^{\prime} + V) \in \mathcal{O} \} + \delta
\end{IEEEeqnarray}     
for any measurable set $\mathcal{O} \subseteq  \{ alg(Y + \mathrm{V})\; | \;  Y \in \mathbb{R}^{N \times p}, \mathrm{V} \in \mathbb{R}^{N \times p}  \}$ and for \emph{neighboring} matrices pair $(Y,Y^{\prime})$. Previously, the noise distribution (from which each element of noise matrix $V$ is independently sampled), that achieves differential privacy inequality (\ref{eq_260520231423}) with the minimum possible noise magnitude, has been derived~\cite{Kumar/IWCFS2019} using an entropy based approach. The optimal expected noise magnitude is given as~\cite{Kumar/IWCFS2019}:
\begin{IEEEeqnarray}{rCl}
\label{eq_optimal_noise_magnitude_epsilon_delta_privacy} E_{f_{\mathrm{v}_j^i}^*}\left[|v|\right] & = & (1-\delta) \frac{d}{\epsilon},
\end{IEEEeqnarray}  
where $d \in \mathbb{R}_{+}$ is a scalar defining the adjacency between $Y$ and $Y^{\prime}$, and $\mathrm{v}_j^i$ is the $(i,j)-$th element of noise matrix $V$ with its probability density function as $f_{\mathrm{v}_j^i}(v)$. It follows from (\ref{eq_optimal_noise_magnitude_epsilon_delta_privacy}) that despite an optimization, a low value of privacy-loss bound $\epsilon$ requires a large amount of noise leading to a considerable loss in the accuracy of a subsequent machine learning algorithm operating on the noise added data. 

\begin{figure}
\centerline{\subfigure[original data samples and corresponding differentially private approximations using optimal noise adding mechanism ($\epsilon = 1$)]{\includegraphics[width=0.46\textwidth]{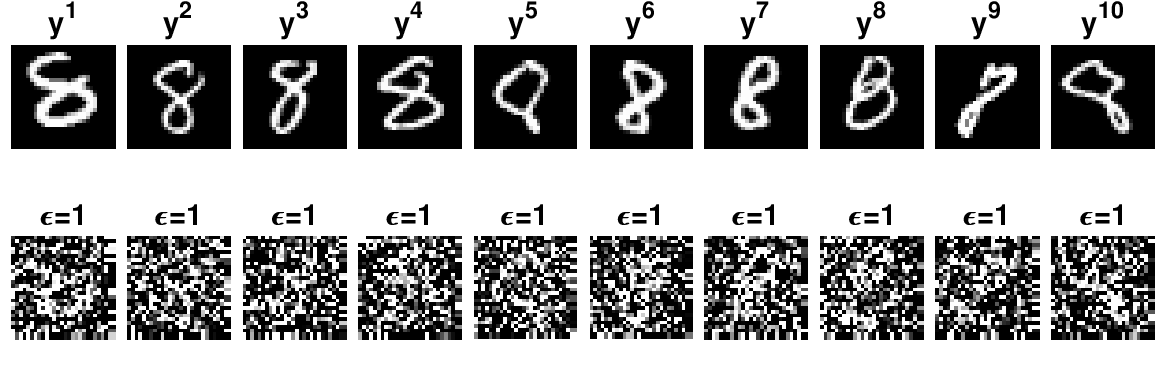}\label{fig_300520231642}} \hfil \subfigure[original data samples and corresponding differentially private fabricated data samples ($\epsilon = 1$)]{\includegraphics[width=0.46\textwidth]{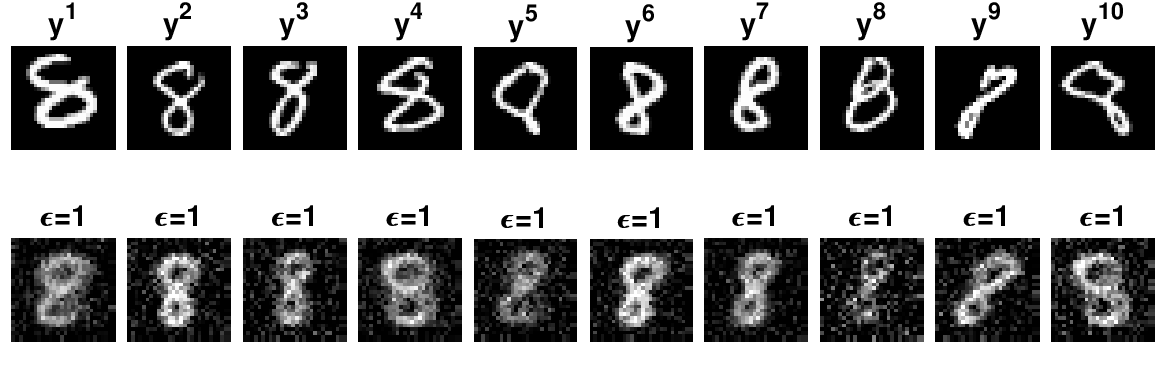}\label{fig_300520231641}}  }
\caption{An example of the data fabrication by means of a geometric model ensuring the geometric modeling error of fabricated data samples not to exceed that of original data samples without compromising on the value of privacy-loss bound $\epsilon$.}
\label{fig_3005202316}
\end{figure}

To mitigate the effect of noise, the flexibility of defining computational algorithm $alg$ in (\ref{eq_260520231423}) can be leveraged without compromising on the privacy-loss bound $\epsilon$. Specifically, a model of the geometric structure induced by noise added data points can be integrated in the definition of $alg$ for a \emph{smoothing}. The $alg$ can be defined as a composition of a smoothing and the machine learning algorithm:
 \begin{IEEEeqnarray}{rCl}
alg & := & machine\_learning \circ smoothing.
\end{IEEEeqnarray}       
Here, the $smoothing$ is based on a model (that represents the geometric structure induced by the noise added data points) ensuring that the smoothing leads to the fabrication of new data points which are not only differentially private but also their geometric modeling error does not exceed that of original data points. Fig.~\ref{fig_3005202316} provides an example of the data fabrication by means of such a geometric model. 
The central problem of this study is stated in the following: 
\begin{problem}[Central Problem]\label{problem_central}
To mitigate the accuracy-loss issue of differential privacy, the post-processing property of differential privacy is leveraged for fabricating new data samples by means of a geometric model ensuring the geometric modeling error of fabricated data samples not to be larger than that of original data samples while simultaneously achieving the privacy-loss bound.
\end{problem}
\begin{remark}[Motivation]\label{rem_motivation}
To our best knowledge, the state-of-the-art does not address Problem~\ref{problem_central}. It requires an approach to learn the representation of geometric structure induced by a finite set of data points. Encouraged by the fact that kernel-based solutions can be computed analytically and analyzed using a broad range of mathematical techniques, the approach opted in this study to address Problem~\ref{problem_central} is of learning in Reproducing Kernel Hilbert Spaces (RKHS) the representation of data points to design a geometrically inspired model such that the model output range defines a bounded geometric structure in the affine hull of given data samples. 
\end{remark}

Kernels have been widely used in machine learning~\shortcite{Ghojogh2021ReproducingKH,10.1214/009053607000000677} and can be scaled up for their applicability in large scale scenarios~\shortcite{NIPS2017_05546b0e}. Not only the parallels between the properties of deep neural networks and kernel methods have been established~\shortcite{pmlr-v80-belkin18a}, but also deep kernel machines have been introduced~\shortcite{pmlr-v51-wilson16,10.1007/s10044-020-00933-1}. Kernel autoencoders are effective models for representation learning. The kernel formulation of an autoencoder has been considered in \cite{7477690} from a hashing perspective. A deep autoencoder, that aligns the latent code with a user-defined kernel matrix to learn similarity-preserving data representations, has been suggested~\shortcite{KAMPFFMEYER2018816}. Further, a kernel autoencoder based on the composition of mappings from vector-valued reproducing kernel Hilbert spaces has been studied \shortcite{added1}. Recently, a fuzzy theoretic approach to kernel based wide and conditionally deep autoencoders has been introduced \shortcite{8888203,9216097,ZHANG2022128,KUMAR20211,ZHANG2023120145,10.1007/978-3-030-87101-7_13,10.1007/978-3-030-87101-7_14,10012502}, wherein analytical solutions are derived for the learning of models using variational optimization technique. This approach has been further extended to privacy-preserving learning under a differential privacy framework~\cite{kumar2023differentially,KUMAR202187,10.1145/3386392.3399562}. As an alternative to the SVM, the idea of affine hull large margin classifier has been investigated~\shortcite{CEVIKALP20103160}. Although kernel methods have been studied~\shortcite{pmlr-v28-jain13,chaudhuri2011differentially,8765749} under differential privacy, no previous study has considered geometrically inspired kernel methods to mitigate the accuracy-loss issue of differential privacy. \emph{State of the art lacks geometrically inspired kernel machines for scalable learning solutions that remain accurate even after providing differential privacy guarantee.}

\begin{figure}
\centerline{\subfigure[first example]{\includegraphics[width=0.33\textwidth]{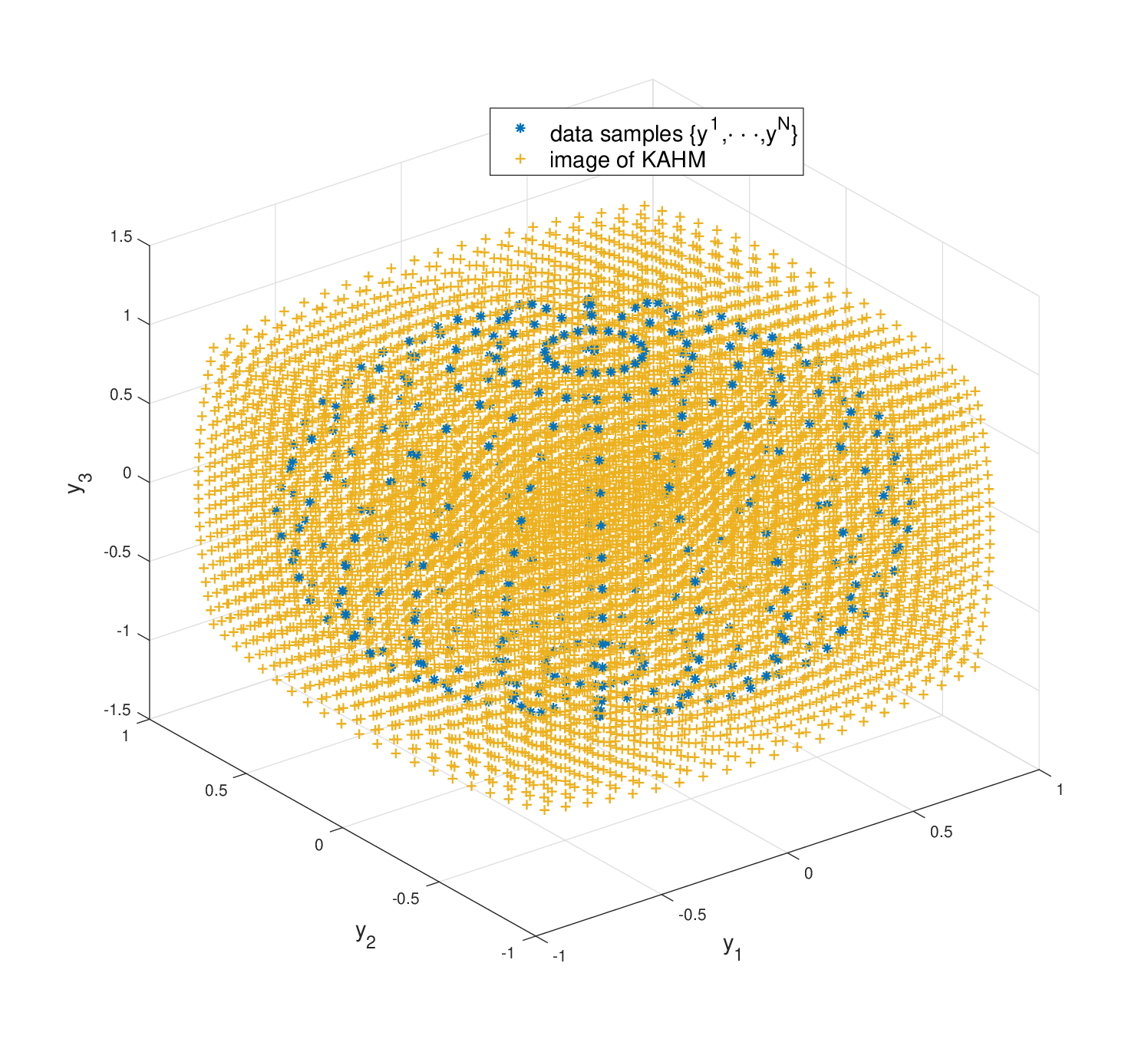}\label{kahm_example_3d_1}} \hfil 
\subfigure[second example]{\includegraphics[width=0.33\textwidth]{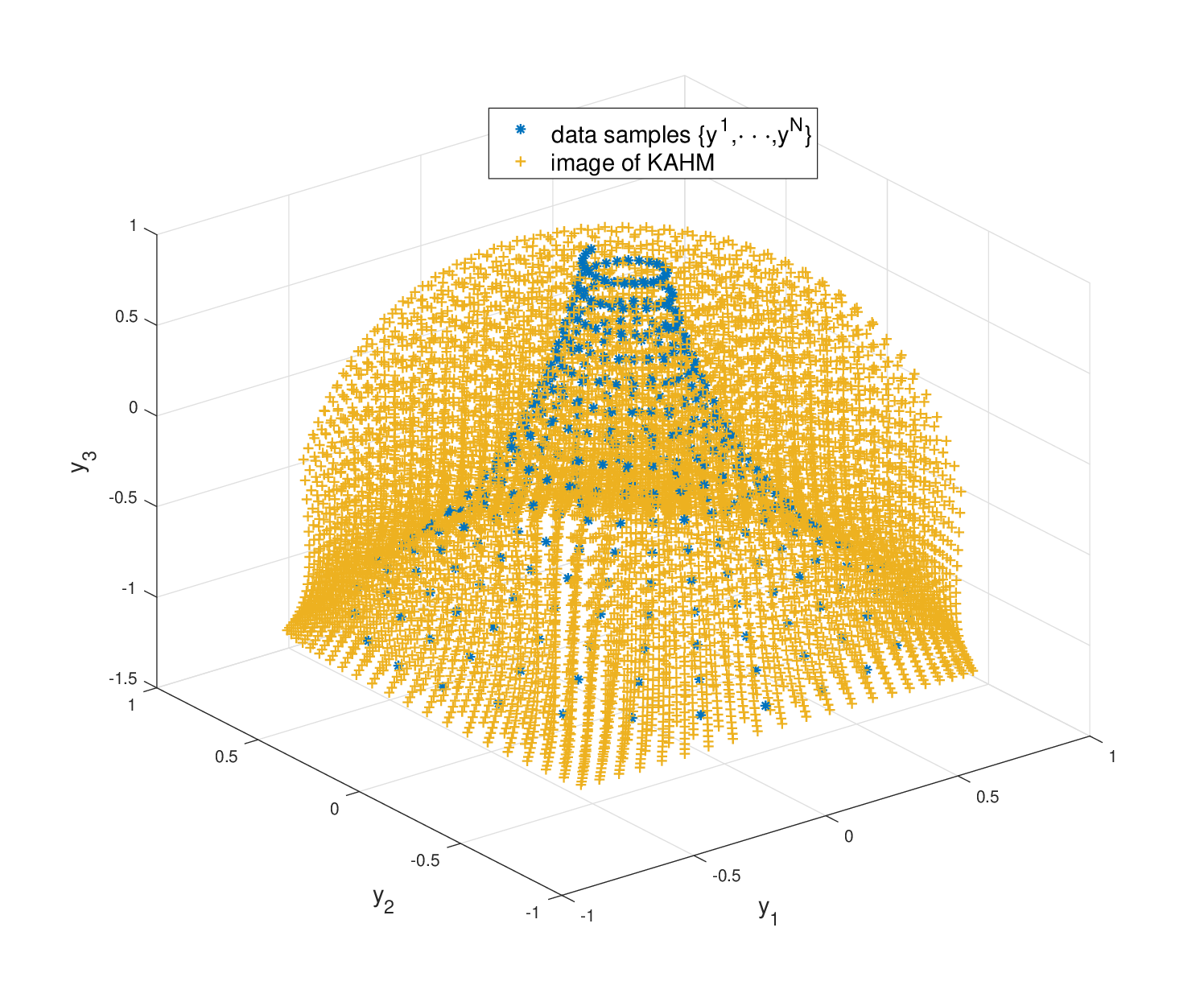}\label{kahm_example_3d_2}} \hfil \subfigure[third example]{\includegraphics[width=0.33\textwidth]{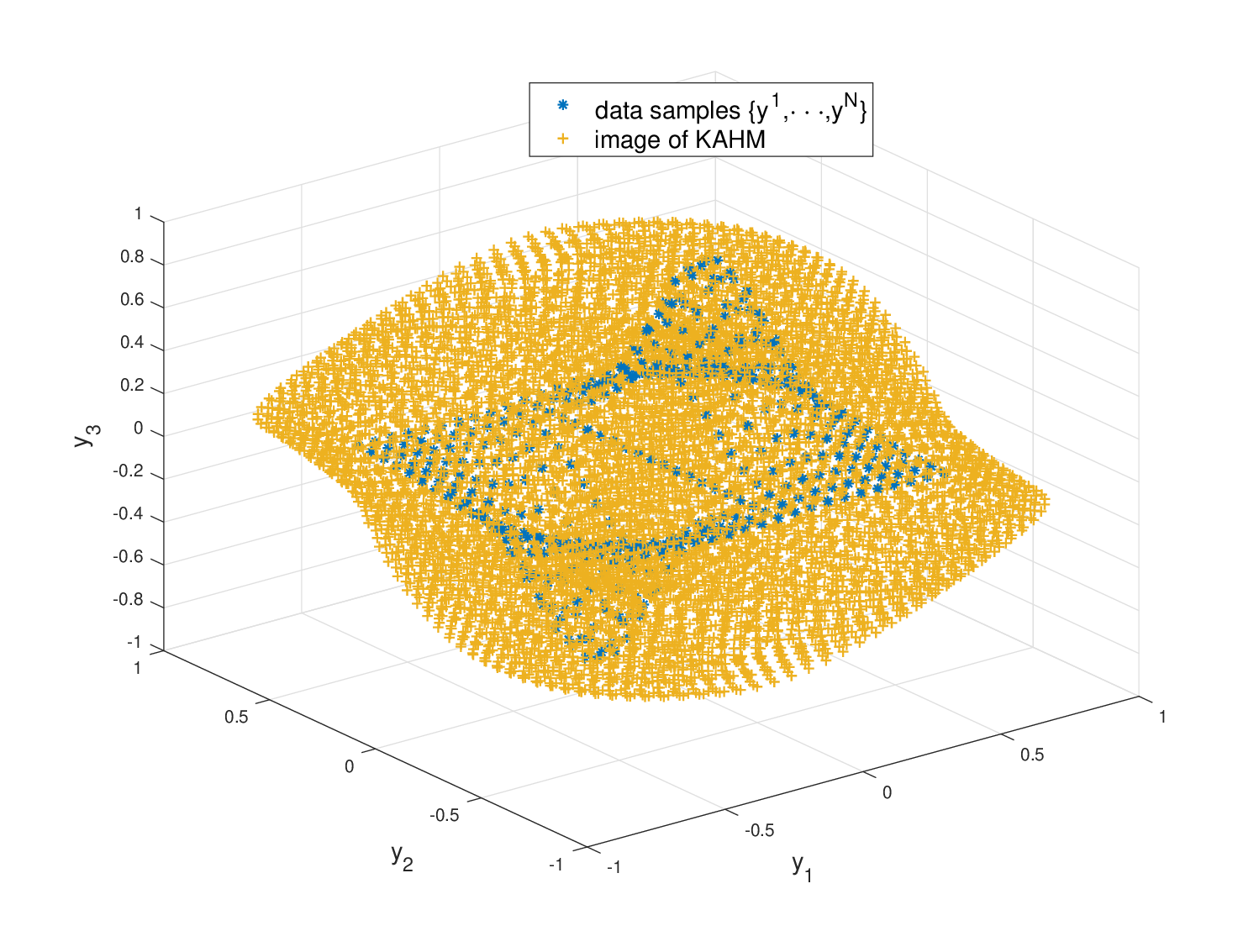}\label{kahm_example_3d_3}}  }
\caption{A few examples of 3-dimensional samples $\{y^1,\cdots,y^N \}$ and geometric structures in $\mathrm{aff}(\{y^1,\cdots,y^N \})$ defined by the images of corresponding KAHMs.}
\label{fig_KAHM_examples_3d}
\end{figure}

\begin{figure}
\centerline{\subfigure[data samples and images of corresponding KAHMs]{\includegraphics[width=0.33\textwidth]{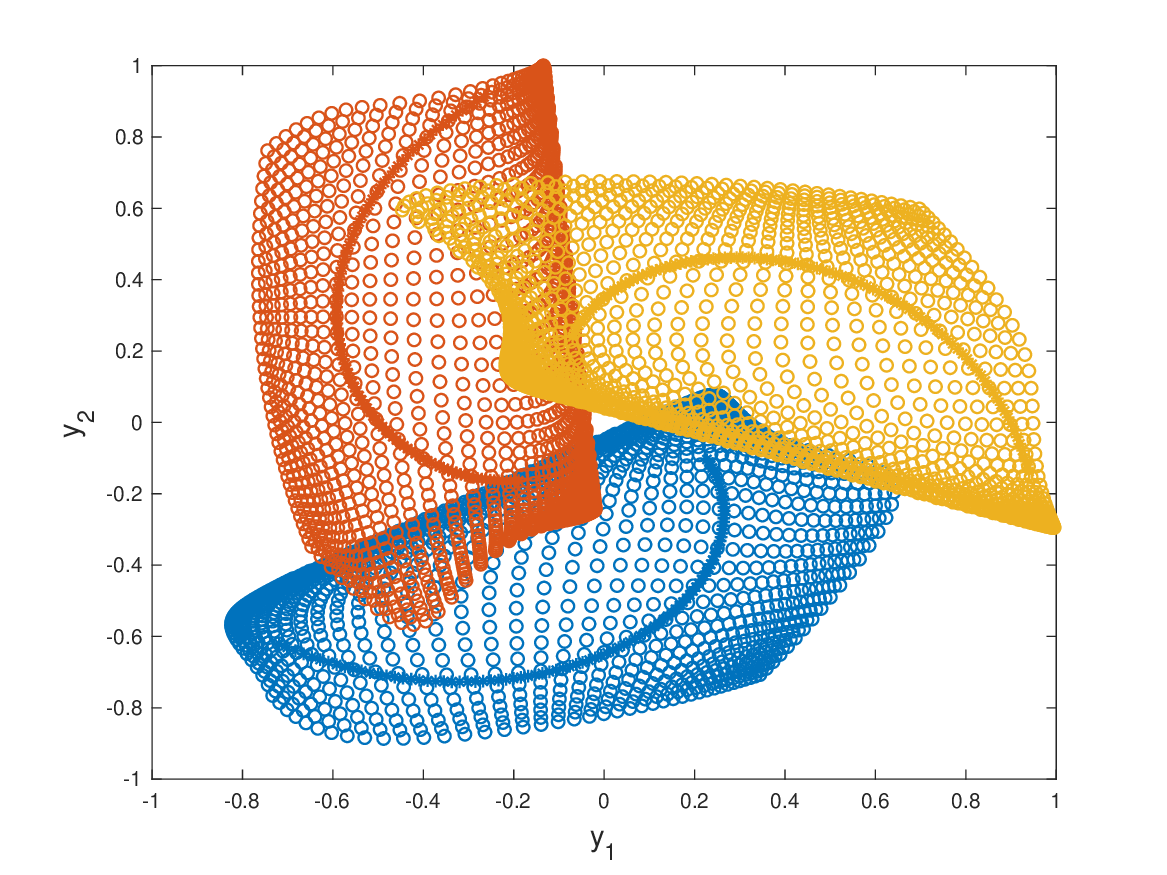}\label{fig_demo_dp_fabricated_data2D_1}} \hfil 
\subfigure[differentially private fabricated data samples and images of corresponding KAHMs]{\includegraphics[width=0.33\textwidth]{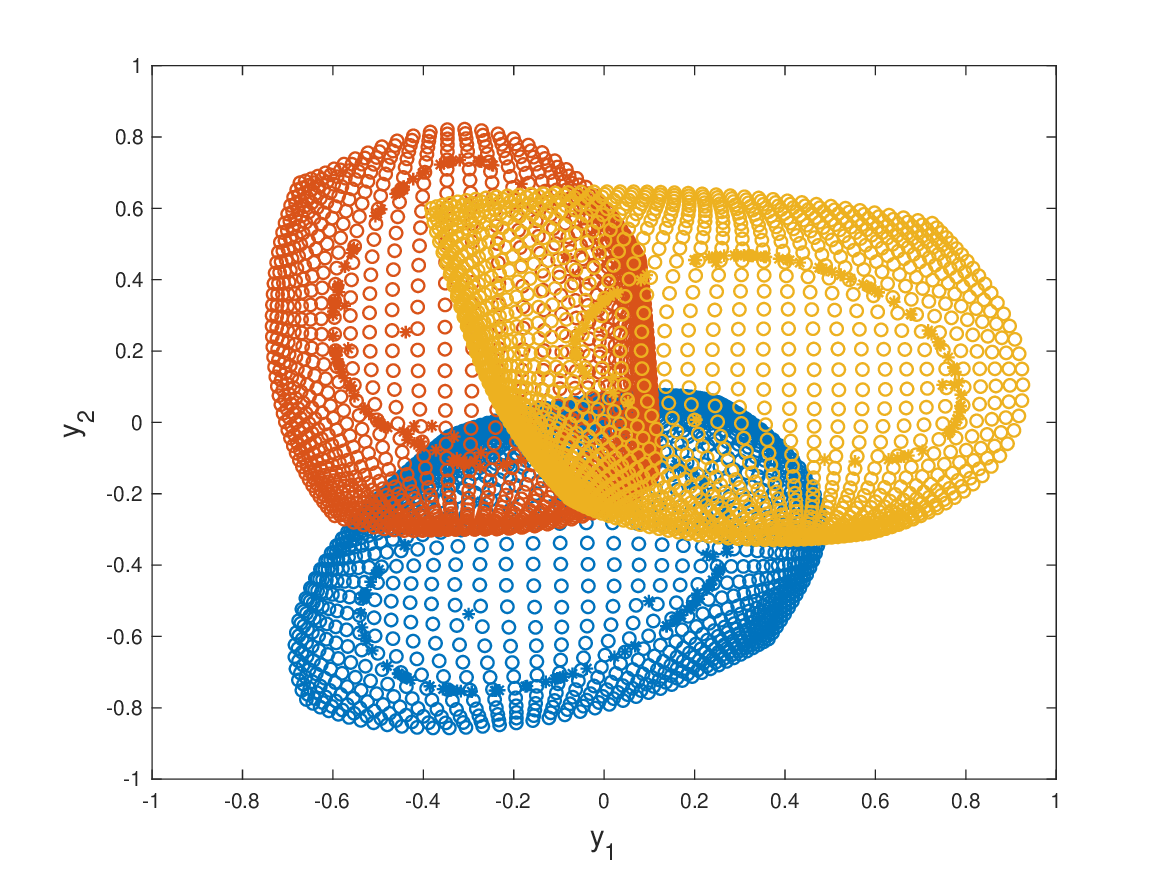}\label{fig_demo_dp_fabricated_data2D_2}} \hfil 
\subfigure[decision boundaries determined by KAHM based differentially private classifier with fabricated data]{\includegraphics[width=0.33\textwidth]{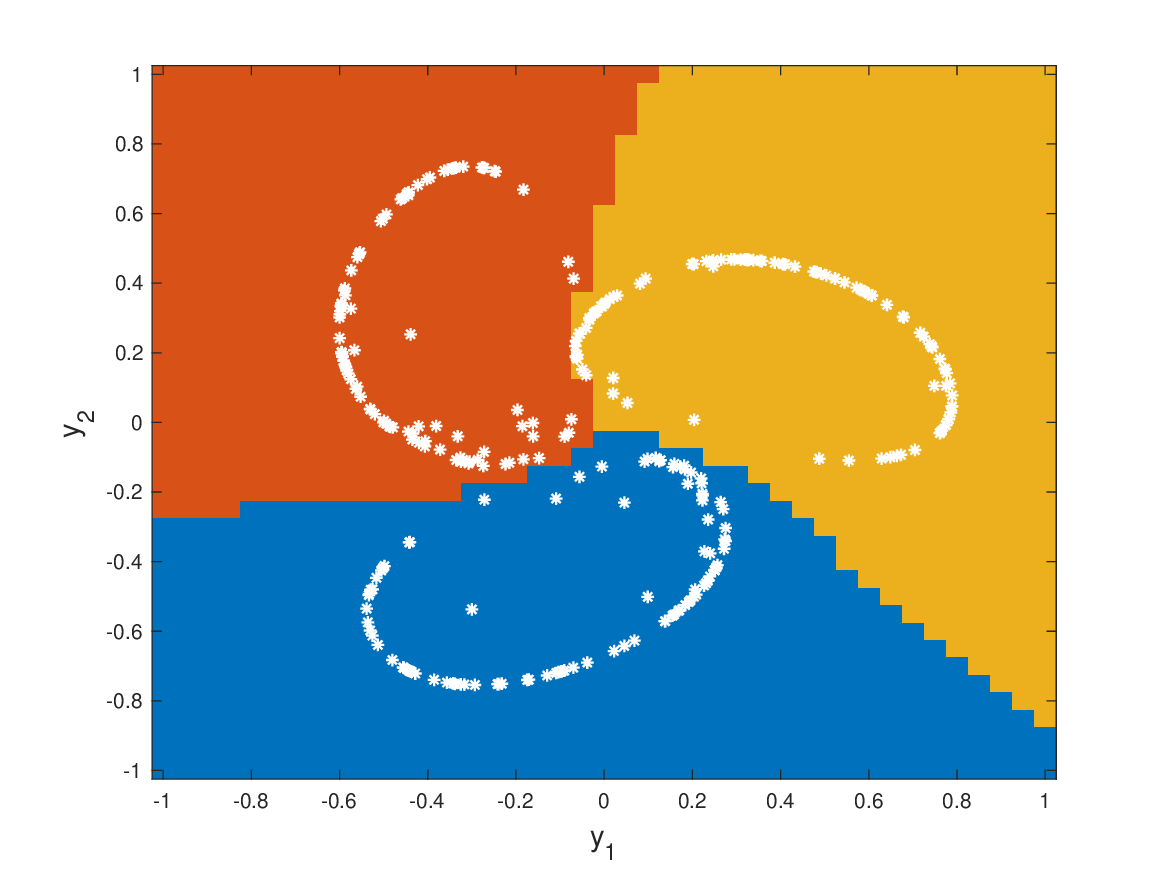}\label{fig_demo_dp_fabricated_data2D_3}}}
\caption{An example of differentially private classifier built with a 2-dimensional fabricated dataset with 3 classes.}
\label{fig_demo_dp_fabricated_data2D}
\end{figure}

This study solves Problem~\ref{problem_central} via making the following contributions (C1-C7):
\paragraph{Kernel Affine Hull Machines (C1):} For given distinct data points $(y^i)_{i= 1, \ldots, N}$ in some vector space we study the sets of the affine form 
\begin{IEEEeqnarray}{rCl}
\mathcal{L} & = & \left \{y =  \left(w^1/\sum_{i=1}^N w^i\right) y^1+ \cdots + \left(w^N/\sum_{i=1}^N w^i\right) y^N \; \mid \; w^i \in \mathbb{R} \right \},
 \end{IEEEeqnarray}  
and ask for reasonable conditions on the real-valued scalars $(w^{i})_i$ to serve our purpose of representing the geometric structure induced by data points. First of all, in our approach $(w^{i})_i$ are considered to be functions in a RKHS. By postulating that indicator functions (specifically, their RKHS approximations) define scalar-valued functions $(w^{i})_i$, the set $\mathcal{L}$ actually can be identified by functions defining a subset in RKHS that represents our data points. This way we introduce the concept of Kernel Affine Hull Machine (KAHM) to learn kernel-based representation of multivariate scattered data as in the following:

Let $n,p,N$ be the positive integers and $\mathcal{X} \subset \mathbb{R}^n$ be a region. Let $\mathcal{H}_k(\mathcal{X})$ be the reproducing kernel Hilbert space of functions from $\mathcal{X}$ to $\mathbb{R}$ for a reproducing kernel $k : \mathcal{X} \times \mathcal{X} \rightarrow \mathbb{R}$. For a finite set of ordered pairs $\{(x^i,y^i) \in \mathcal{X} \times \mathbb{R}^p \; \mid \; i \in \{1,\cdots,N \}\}$ such that $\{ x^1,\cdots,x^N\}$ are pairwise distinct points, a point $y^i$ can be represented using indicator functions as
\begin{IEEEeqnarray}{rCl}
  y^i & = &   \mathbbm{1}_{\{x^1\}}(x^i) \: y^1 + \cdots + \mathbbm{1}_{\{x^N\}}(x^i)\: y^N,
 \end{IEEEeqnarray} 
 where $\mathbbm{1}_{\{x^i\}}$ is the indicator function of the set $\{ x^i\}$. We approximate the indicator function $\mathbbm{1}_{\{x^i\}}$ through a function in $ \mathcal{H}_{k}(\mathcal{X})$ that fits to the ordered pairs $\{ \left(x^j, \mathbbm{1}_{\{x^i\}}(x^j)\right) \; \mid \; j \in \{1,\cdots,N \} \}$. The function in RKHS approximating $\mathbbm{1}_{\{x^i\}}$ is given as the solution of the following kernel regularized least squares problem: 
   \begin{IEEEeqnarray}{rCl}
 \label{eq_220320231124}h^i & = & \arg \; \min_{f \in \mathcal{H}_k(\mathcal{X})} \; \left( \sum_{j=1}^N \left |\mathbbm{1}_{\{x^i\}}(x^j) - f(x^j) \right |^2 + \lambda \left \| f \right \|^2_{\mathcal{H}_k(\mathcal{X})} \right),\; \lambda \in \mathbb{R}_+,
  \end{IEEEeqnarray}    
where $\left \| f \right \|_{\mathcal{H}_k(\mathcal{X})} := \sqrt{\left < f, f \right>_{\mathcal{H}_k(\mathcal{X})}}$ is the norm induced by the inner product on $\mathcal{H}_k(\mathcal{X})$. The fact that $h^i $ is an approximation of $\mathbbm{1}_{\{x^i\}}$ (i.e. the value $h^i(x)$ represents kernel-smoothed ``membership'' of $x$ to the set $\{ x^i \}$) allows introducing a model based on the affine combination of $y^1,\cdots,y^N$ as in the following: 
   \begin{IEEEeqnarray}{rCl}
A(x) & = & \frac{h^1(x)}{\sum_{i=1}^N h^i(x)}\:y^1 + \cdots + \frac{h^N(x)}{\sum_{i=1}^N h^i(x)}\: y^N. 
   \end{IEEEeqnarray}    
Let $\mathrm{aff}(\{y^1,\cdots,y^N \})$ denote the affine hull of $\{y^1,\cdots,y^N\}$. The function $A:\mathcal{X} \rightarrow \mathrm{aff}(\{y^1,\cdots,y^N \})$ is referred to as kernel affine hull machine, since it maps a point $x \in \mathcal{X}$ onto the affine hull of $\{y^1,\cdots,y^N\}$ via learning representation of $x^1,\cdots,x^N$ through functions in reproducing kernel Hilbert space. The image of $A$, 
   \begin{IEEEeqnarray}{rCCCl}
A[\mathcal{X}] & := & \{ A(x) \; \mid \; x \in \mathcal{X} \} & \subset & \mathrm{aff}(\{y^1,\cdots,y^N \}),
 \end{IEEEeqnarray}  
defines a geometric structure in $\mathrm{aff}(\{y^1,\cdots,y^N \})$. Fig.~\ref{fig_KAHM_examples_3d} displays a few examples of 3-dimensional samples and geometric structures defined by KAHMs' images.    
\paragraph{Regularization Parameter for Kernel Regularized Least Squares (C2):} Since indicator functions are approximated via solving a regularized least squares problem, the kernel regularized least squares problem is revisited in a deterministic setting with focus on the determination of regularization parameter. A reasonable choice for regularization parameter is to set it larger than the mean-squared-error on training samples. With this choice, the problem of determining regularization parameter can be reduced to an equivalent problem of finding the unique fixed point of a real-valued positive function. An iterative scheme, together with the mathematical proof of convergence, is provided to find the fixed point and thus to determine the regularization parameter.  
\paragraph{Boundedness of KAHM and Distance Function (C3):} The KAHM mapping is a bounded function and thus the image of KAHM defines a bounded region in the affine hull of data samples. The boundedness of KAHM on data space is proven via deriving upper bounds on the Euclidean norm of KAHM output. The KAHM induces a function on data space, referred to as \emph{distance function}, which is defined on a data point as equal to the distance between that point and its image under KAHM. The distance of an arbitrary point from its image (by the KAHM onto the affine hull of given data samples) is a measure of the distance between that arbitrary point and the given data samples. This is proven via deriving upper bounds on the ratio of these two distances. 
\paragraph{KAHM Compositions for Data Representation Learning and Classification (C4):} The KAHM could serve as the building block for deep models. A nested composition of KAHMs, referred to as \emph{Conditionally Deep Kernel Affine Hull Machine}, is considered for data representation learning. The conditionally deep KAHM discovers layers of increasingly abstract data representation with lowest-level data features being modeled by first layer and the highest-level by end layer. Further, a parallel composition of conditionally deep KAHMs, referred to as \emph{Wide Conditionally Deep Kernel Affine Hull Machine}, is considered to efficiently learn the representation of big data. Similarly to the KAHM, both conditional deep KAHM and wide conditionally deep KAHM induce the distance function with value on a point indicating the distance of the point from data samples. This property of the distance function is leveraged to build a KAHM based classifier via modeling the region of each class through a separate KAHM based composition. 
\paragraph{Membership-Inference Score for KAHM Based Classifier (C5):} Since the KAHM based classifier assigns a class-label to a data point based on the closeness of the point to the training data samples of that class, there is a possibility of an inference of the membership of a data point to the set of training data samples. To evaluate the potential of KAHM induced distance function in inferring the membership of a data point to the training dataset, a score, referred to as \emph{membership-inference score}, is defined for evaluating the risk of {\em membership inference attack}. The membership-inference score is defined as the $L2$ distance between density of probability distribution on values of the distance function at training data points and the density of probability distribution on distance function values at test data points. 
\paragraph{Differentially Private Data Fabrication for Classification (C6):} To ensure that KAHM based classifier keeps the privacy of training data protected, an optimal differentially private noise adding mechanism~\cite{Kumar/IWCFS2019} is applied on training data samples. The noise added training data samples are smoothed through a transformation such that the error in KAHM modeling of smoothed data is not larger than the error in KAHM modeling of original data. It is shown that the error in KAHM modeling of smoothed data can be reduced to an arbitrary low value. The smoothed data samples, guaranteeing not only the differential privacy but also the geometric modeling error not to be larger than that of original data samples, serve as the \emph{fabricated} data. The fabricated data samples are finally used to build the KAHM based differentially private classifier. The advantage of using fabricated data for classification is that fabricated data leads to a considerable reduction in the risk of membership inference attack with relatively much smaller loss of accuracy. Hence, the accuracy-loss issue of differential privacy is mitigated. Fig.~\ref{fig_demo_dp_fabricated_data2D} provides an example of differentially private classifier built with a 2-dimensional fabricated dataset with 3 classes. 
\paragraph{Application to Differentially Private Federated Learning (C7):} The different KAHMs built independently using different datasets can be combined together using the KAHM induced distance function. This allows introducing a federated learning scheme that combines together the local privacy-preserving KAHM based classifiers to build a global classifier. A significant feature of the scheme is that the evaluation of global classifier requires only locally computed distance measures. 

The relation of the current study with previous works is confined to the following three points: 1) The wide and conditionally deep architecture consisting of the composition of kernel based models follows from~\cite{8888203,9216097,KUMAR20211,ZHANG2022128,10.1007/978-3-030-87101-7_14}, wherein a kernel based variational fuzzy model (motivated by measure theoretic basis~\cite{10.1007/978-3-030-87101-7_13}) is used. In contrast, the current study explores geometrically inspired kernel affine hull machines. 2) The input perturbation method (where noise is added to original data to achieve $(\epsilon,\delta)-$differential privacy of any subsequent computational algorithm) was earlier considered in~\cite{KUMAR202187,10.1145/3386392.3399562,kumar2023differentially}. However, the current study complements the input perturbation method with a transformation to mitigate the accuracy-loss issue of differential privacy. 3) The current study follows the federated learning architecture of~\cite{KUMAR202187,10.1145/3386392.3399562,10012502} with the difference that instead of fuzzy attributes, the KAHM induced distance measures are applied to aggregate the distributed local models for federated learning. 

The significance and novelties of the contributions have been highlighted in Table~\ref{table_contributions} and Table~\ref{table_novelty} respectively.  
\begin{table}[h!]
\renewcommand{\arraystretch}{1.3}
\centering
\caption{The significance of the contributions.}
\label{table_contributions}
{\footnotesize %
\begin{tabular}{r|l}
\hline 
& \bfseries Significance   \\
\hline \hline
C1 & Representations learning in RKHS for defining a geometric structure in the affine hull of data samples \\ \hline
C2 & Determination of the regularization parameter for kernel regularized least squares  \\  \hline
C3 & $\begin{array}{l} \mbox{Because of the boundedness of KAHM, the distance of an arbitrary point from its KAHM image}\\ \mbox{is a measure of the distance between that arbitrary point and the given data samples} \end{array}$    \\ \hline
C4 & $\begin{array}{l} \mbox{KAHM compositions learn geometrically inspired representations at varying abstraction level} \\ \mbox{facilitating classification via modeling the region of each class through a separate composition} \end{array}$     \\ \hline
C5 & Evaluation of the risk of membership inference attack on KAHM based classifier     \\ \hline
C6 & $\begin{array}{l} \mbox{Differentially private data fabrication to mitigate the accuracy-loss issue associated with}\\ \mbox{the differentially private classifier} \end{array}$    \\ \hline
C7 & Application to differentially private federated learning  \\ \hline
\hline
\end{tabular}}
\end{table}
\begin{table}[h!]
\renewcommand{\arraystretch}{1.3}
\centering
\caption{The novelties in the contributions.}
\label{table_novelty}
{\footnotesize %
\begin{tabular}{r|l}
\hline 
& \bfseries Novelty   \\
\hline \hline
C1 & The concept of KAHM (Definition~\ref{def_affine_hull_model}) is novel.  \\ \hline
C2 & $\begin{array}{l} \mbox{Determination of the regularization parameter as the unique fixed point of a} \\ \mbox{function (Theorem~\ref{result_definition_mse_function}) is novel.} \end{array}$  \\  \hline
C3 & $\begin{array}{l} \mbox{The idea of using bounded geometric structure (Theorem~\ref{result_kahm_bounded_function}) to define a measure of the distance}\\ \mbox{from given data samples (Theorem~\ref{result_ratio_distances}) is novel.} \end{array}$    \\ \hline
C4 & $\begin{array}{l} \mbox{Geometrically inspired representations learning at varying abstraction level and corresponding} \\ \mbox{induced measure of the distance from given data samples (Theorem~\ref{result_ratio_distances_cond_deep_KAHM}, Theorem~\ref{result_ratio_distances_wide_cond_deep_KAHM}) is novel.} \end{array}$       \\ \hline
C5 & $\begin{array}{l} \mbox{Quantification of membership inference attack risk as $L2$ distance between density of distance} \\ \mbox{from training data samples and density of distance from test data samples (Eq. (\ref{eq_270220231128})) is novel.} \end{array}$    \\ \hline
C6 & $\begin{array}{l} \mbox{Data fabrication via transforming differentially private data samples to reduce}\\ \mbox{geometric modeling error (Definition~\ref{def_transformation}, Theorem~\ref{result_transformation}, Definition~\ref{def_fabricated_data}) is novel.} \end{array}$    \\ \hline
C7 & $\begin{array}{l} \mbox{The feature of the federated learning that the evaluation of global classifier requires only} \\ \mbox{locally computed KAHM induced distance measures (Fig.~\ref{fig_KAHM_distributed_classifier}) is novel.} \end{array}$ \\ \hline
\hline
\end{tabular}}
\end{table}

The paper is organized into the following sections. The mathematical notation used throughout the paper is provided in Section~\ref{sec_notations}. Section~\ref{sec_kahm} introduces the concept of KAHM. KAHM based wide and deep models are presented in Section~\ref{sec_kahm_compositions} for data representation learning. Differentially private classification application is considered in Section~\ref{sec_privacy} followed by experimentation in Section~\ref{sec_experiments}. Finally, the concluding remarks are presented in Section~\ref{sec_conclusion}.   
\section{Notations}\label{sec_notations} 
The following notations are introduced:
\begin{itemize}
\item Let $N,n,p,M,S,C \in \mathbb{Z}_{+}$ be the positive integers.
\item Let $\mu_{max}(K)$ and $\mu_{min}(K)$ denote the maximum eigenvalue and minimum eigenvalue respectively of a square matrix $K$. 
\item Let $\sigma_{max}(K)$ and $\sigma_{min}(K)$ denote the maximum singular value and minimum singular value of a matrix $K$. 
\item Let $\mathrm{aff}(\mathbf{Y})$ denote the affine hull of a set $\mathbf{Y} \subset  \mathbb{R}^p$. 
\item For a vector $y \in \mathbb{R}^p$, $\| y\|$ denotes the Euclidean norm of $y$. 
\item For a matrix $Y \in \mathbb{R}^{N \times p}$, $\|Y\|_2$ denotes the spectral norm, $\| Y \|_F$ denotes the Frobenius norm, $\|Y\|_1$ denotes the 1-norm, $\|Y\|_{\mathrm{max}}$ denotes the max norm, $(Y)_{i,:}$ denotes the $i-$th row, $(Y)_{:,j}$ denotes the $j-$th column, and $(Y)_{i,j}$ denotes the $(i,j)-$th element of $Y$. 
\item Let $\circ$ denote the Hadamard product.
\item Let $I_N$ denote the identity matrix of the size $N$ and $\mathbf{1}_N$ denotes the $N \times 1$ vector of ones. 
\item Let $\mathbbm{1}_{\mathbf{Y}}$ denote the indicator function of the set $\mathbf{Y}$.
\item Let $\mathcal{X} \subset \mathbb{R}^n$ be a region.
\item $ K \succ 0$ denotes that a symmetric matrix $K$ is positive definite.
\item A Reproducing Kernel Hilbert Space (RKHS), $\mathcal{H}_{k}(\mathcal{X})$, is a Hilbert space of functions $f: \mathcal{X} \rightarrow \mathbb{R}$ on a non-empty set $\mathcal{X}$ with a reproducing kernel $k: \mathcal{X} \times \mathcal{X} \rightarrow \mathbb{R}$ satisfying $\forall x \in \mathcal{X}$ and $\forall f \in \mathcal{H}$,
 \begin{itemize}
 \item $k(\cdot,x) \in \mathcal{H}_k(\mathcal{X})$, 
 \item $\left < f, k(\cdot,x)  \right>_{\mathcal{H}_k(\mathcal{X})} = f(x)$, 
 \end{itemize}  
where $\left < \cdot, \cdot \right>_{\mathcal{H}_k(\mathcal{X})} : \mathcal{H}_k(\mathcal{X}) \times \mathcal{H}_k(\mathcal{X}) \rightarrow \mathbb{R}$ is an inner product on $\mathcal{H}_k(\mathcal{X})$.  
\item Let $\left \| f \right \|_{\mathcal{H}_k(\mathcal{X})} := \sqrt{\left < f, f \right>_{\mathcal{H}_k(\mathcal{X})}}$ denote the norm induced by the inner product on $\mathcal{H}_k(\mathcal{X})$.    
\end{itemize}
\section{Kernel Affine Hull Machines}\label{sec_kahm}
The computation of KAHM requires solving a kernel regularized least squares problem. Therefore the kernel regularized problem is revisited (in Section~\ref{sec_kernel_regularized_least_squares}) with focus on the determination of regularization parameter (in Section~\ref{sec_regularization_parameter}). The obtained solution is applied (in Section~\ref{sec_learning_representation}) to learn data representation in RKHS facilitating the definition of KAHM (in Section~\ref{sec_an_affine_hull_machine}).  
\subsection{Kernel Regularized Least Squares}\label{sec_kernel_regularized_least_squares}
Given a training data set $\left\{ (x^i,y^i) \in \mathcal{X} \times \mathbb{R}^p \; \mid \; i \in \{1,\cdots,N \} \right\}$ such that $\{ x^1,\cdots,x^N\}$ are pairwise distinct points, consider a positive-definite real-valued kernel $k: \mathcal{X} \times \mathcal{X} \rightarrow \mathbb{R}$ on $\mathcal{X}$ with a corresponding RKHS $\mathcal{H}_k(\mathcal{X})$. Assuming that $\mathcal{X} \subset \mathbb{R}^n$, real-valued matrices $X \in \mathbb{R}^{N \times n}$ and $Y \in \mathbb{R}^{N \times p}$ are defined as
 \begin{IEEEeqnarray}{rCl}
X & = & \left[\begin{IEEEeqnarraybox*}[][c]{,c/c/c,} x^1 & \cdots & x^N \end{IEEEeqnarraybox*} \right]^T\\
 Y & = & \left[\begin{IEEEeqnarraybox*}[][c]{,c/c/c,} y^1 & \cdots & y^N\end{IEEEeqnarraybox*} \right]^T.
   \end{IEEEeqnarray} 
Let $(Y)_{:,j}$ denote the $j-$th column of $Y$, i.e.,  
 \begin{IEEEeqnarray}{rCl}    
(Y)_{:,j} & = &    \left[\begin{IEEEeqnarraybox*}[][c]{,c/c/c,} y^1_j & \cdots & y^N_j\end{IEEEeqnarraybox*} \right]^T,
  \end{IEEEeqnarray}        
where $j \in \{1,\cdots,p \}$ and $y^i_j$ is the $j-$th element of $i-$th output sample $y^i$. The solution of the following regularized least squares problem:
   \begin{IEEEeqnarray}{rCl}
 f^*_{k,X,(Y)_{:,j},\lambda} & = & \arg \; \min_{f \in \mathcal{H}_k(\mathcal{X})} \; \left( \sum_{i=1}^N \left |y^i_j - f(x^i) \right |^2 + \lambda \left \| f \right \|^2_{\mathcal{H}_k(\mathcal{X})} \right),\; \lambda \in \mathbb{R}_+,
  \end{IEEEeqnarray} 
using the representer theorem~\cite{10.1007/3-540-44581-1_27}, can be written as 
   \begin{IEEEeqnarray}{rCl}
\label{eq_738824.675718} f^*_{k,X,(Y)_{:,j},\lambda}(x) & = &  \left[\begin{IEEEeqnarraybox*}[][c]{,c/c/c,} k(x,x^1) & \cdots & k(x,x^N)\end{IEEEeqnarraybox*} \right] \left(K_{X} + \lambda I_N \right)^{-1} (Y)_{:,j}
  \end{IEEEeqnarray} 
where $I_N$ is the identity matrix of size $N$ and $K_{X}$ is $N \times N$ kernel matrix whose $(i,j)-$th entry is given as
   \begin{IEEEeqnarray}{rCl}
\label{eq_240220231004}  (K_{X})_{ij} & = & k(x^i, x^j).
  \end{IEEEeqnarray}   
The regularized least squares problem can be solved for each output dimension resulting in a vector-valued function $\mathbf{f}^*_{k,X,Y,\lambda}: \mathcal{X} \rightarrow \mathbb{R}^p$ defined as
\begin{IEEEeqnarray}{rCl}
\label{eq_738824.688114}  \mathbf{f}^*_{k,X,Y,\lambda}(x) & := &  \left[\begin{IEEEeqnarraybox*}[][c]{,c/c/c,} f^*_{k,X,(Y)_{:,1},\lambda}(x) & \cdots & f^*_{k,X,(Y)_{:,p},\lambda}(x) \end{IEEEeqnarraybox*} \right]^T \\
& = & Y^T \left(K_{X} + \lambda I_N \right)^{-1} \left[\begin{IEEEeqnarraybox*}[][c]{,c/c/c,} k(x,x^1) & \cdots & k(x,x^N)\end{IEEEeqnarraybox*} \right]^T.
 \end{IEEEeqnarray}     
\subsection{Determination of Regularization Parameter}\label{sec_regularization_parameter}
To compute the kernel regularized least squares solution (\ref{eq_738824.675718}), a choice for regularization parameter $\lambda \in \mathbb{R}_+$ need to be made. A possible choice could be of setting $\lambda$ larger than the mean squared error on training data. The mean squared error on training data, which obviously depends on the choice of regularization parameter $\lambda$, is given as
    \begin{IEEEeqnarray}{rCl}
e(\lambda) & = & \frac{1}{pN} \sum_{j=1}^p \sum_{i=1}^N |y_j^i - f^*_{k,X,(Y)_{:,j},\lambda}(x^i)  |^2 \\
& = &    \frac{1}{pN} \sum_{j=1}^p \|(Y)_{:,j} - K_{X} \left(K_{X} + \lambda I_N \right)^{-1} (Y)_{:,j} \|^2.
     \end{IEEEeqnarray}   
We choose $\lambda$ to be larger than $e$. That is, there exists a constant $\tau \in \mathbb{R}_{+}$ such that 
    \begin{IEEEeqnarray}{rCl}
\lambda & = &  e(\lambda)  + \tau,\mbox{ i.e.,} \\
\lambda  & = & \frac{1}{pN} \sum_{j=1}^p \|(Y)_{:,j} - K_{X} \left(K_{X} + \lambda I_N \right)^{-1} (Y)_{:,j} \|^2 + \tau. \label{eq_738820.4523}
     \end{IEEEeqnarray}       
Eq. (\ref{eq_738820.4523}) can be solved for $\lambda$ via applying the following result:
\begin{theorem}\label{result_definition_mse_function}
Let $\mathcal{R}_{k,X,Y}: \mathbf{R}_{+} \times \mathbf{R}_{+} \rightarrow \mathbf{R}_{+}$ be a function defined as
   \begin{IEEEeqnarray}{rCl}
\mathcal{R}_{k,X,Y}(e,\tau)& := &   \frac{1}{pN} \sum_{j=1}^p \|(Y)_{:,j}- K_{X} \left(K_{X} + (e+\tau) I_N \right)^{-1} (Y)_{:,j} \|^2.
     \end{IEEEeqnarray} 
We have followings:
\begin{enumerate}
\item For $e, \tau \in \mathbf{R}_{+}$,
   \begin{IEEEeqnarray}{rCl}
\mathcal{R}_{k,X,Y}(e,\tau)& \in &  ( 0,\frac{\|Y \|^2_F}{pN} ). \label{eq_738818.778}
     \end{IEEEeqnarray} 
\item For $e,\tau \in \mathbf{R}_{+}$,
   \begin{IEEEeqnarray}{rCl}
 \frac{\dd \mathcal{R}_{k,X,Y}(e,\tau)}{\dd e} & \in &  (0, \frac{2}{(e+\tau)} \frac{\|Y \|^2_F}{pN}   ). \label{eq_738818.7872}
     \end{IEEEeqnarray}      
 \item For a given $\tau \in \mathbf{R}_{+}$, $\mathcal{R}_{k,X,Y}(e,\tau)$ has at least one fixed point in $(0, \|Y \|^2_F/pN )$.  
\item If we choose 
\begin{IEEEeqnarray}{rCl}
\tau & \geq & \frac{2}{pN}\|Y \|^2_F, \label{eq_738819.4989}
 \end{IEEEeqnarray}
 then the iterations
 \begin{IEEEeqnarray}{rCl}
e|_{it+1} & = & \mathcal{R}_{k,X,Y}(e|_{it},\tau),\; it \in \{0,1,\cdots \} \label{eq_738819.5207} \\
e|_0 & \in & (0,\frac{\|Y \|^2_F}{pN} )  \label{eq_738819.5208}
  \end{IEEEeqnarray}
converge to the unique fixed point of $\mathcal{R}_{k,X,Y}(e,\tau)$.   
 \end{enumerate}
  \begin{proof}
 The proof is provided in Appendix~\ref{appendix1}.
 \end{proof}    
\end{theorem}    
It follows from Theorem~\ref{result_definition_mse_function} that the iterations (\ref{eq_738819.5207})-(\ref{eq_738819.5208}) converge to the unique fixed point of $\mathcal{R}_{k,X,Y}(e,\tau)$ for any $\tau$ satisfying (\ref{eq_738819.4989}). Let $\hat{e}$ be the unique fixed point corresponding to the minimum possible value of $\tau$ satisfying (\ref{eq_738819.4989}) (which is equal to $\frac{2}{pN}\|Y \|^2_F$), i.e.,     
    \begin{IEEEeqnarray}{rClC}
\hat{e} & = & \mathcal{R}_{k,X,Y}(\hat{e},\frac{2}{pN}\|Y \|^2_F).
     \end{IEEEeqnarray} 
Now, the value of regularization parameter $\lambda$ satisfying (\ref{eq_738820.4523}) for $\tau = \frac{2}{pN}\|Y \|^2_F$ is given as
    \begin{IEEEeqnarray}{rClC}
\lambda^* & = &  \hat{e} + \frac{2}{pN}\|Y \|^2_F. \label{eq_738820.7747} 
     \end{IEEEeqnarray}    
\subsection{Learning Representation of Data Points in RKHS}\label{sec_learning_representation}
Given a finite number of pairwise distinct points: $X = \left[\begin{IEEEeqnarraybox*}[][c]{,c/c/c,} x^1 & \cdots & x^N \end{IEEEeqnarraybox*} \right]^T$ with $x^1,\cdots,x^N \in \mathcal{X} \subset \mathbb{R}^n$, a data point $x^i$ can be represented using indicator functions as  
\begin{IEEEeqnarray}{rCl}
  x^i & = &   \mathbbm{1}_{\{x^1\}}(x^i) \: x^1 + \cdots + \mathbbm{1}_{\{x^N\}}(x^i)\: x^N,\; \mbox{for any $i \in \{1,\cdots,N\}$}.
 \end{IEEEeqnarray} 
For a kernel-based representation of data points, the indicator functions $\mathbbm{1}_{\{x^1\}},\cdots,\mathbbm{1}_{\{x^N\}}$ are approximated through functions in RKHS $\mathcal{H}_{k}(\mathcal{X})$. To approximate $\mathbbm{1}_{\{x^i\}}$, a function is fitted on the ordered pairs $\{ \left(x^j, \mathbbm{1}_{\{x^i\}}(x^j)\right) \; \mid \; j \in \{1,\cdots,N \} \}$ via solving the following kernel regularized least squares problem:  
   \begin{IEEEeqnarray}{rCl}
\label{eq_240220231003} h^i_{k,X,\lambda}  & = & \arg \; \min_{f \in \mathcal{H}_k(\mathcal{X})} \; \left( \sum_{j=1}^N \left |\mathbbm{1}_{\{x^i\}}(x^j) - f(x^j) \right |^2 + \lambda \left \| f \right \|^2_{\mathcal{H}_k(\mathcal{X})} \right),\; \lambda \in \mathbb{R}_+.
  \end{IEEEeqnarray} 
Using the representer theorem~\cite{10.1007/3-540-44581-1_27}, the solution of (\ref{eq_240220231003}) is as follows: 
   \begin{IEEEeqnarray}{rCl}
 h^i_{k,X,\lambda}(x)  & = & (I_N)_{i,:} \left(K_{X} + \lambda I_N \right)^{-1} \left[\begin{IEEEeqnarraybox*}[][c]{,c/c/c,} k(x,x^1) & \cdots & k(x,x^N)\end{IEEEeqnarraybox*} \right]^T,
  \end{IEEEeqnarray} 
where $K_{X}$ is kernel matrix defined as in (\ref{eq_240220231004}) and $(I_N)_{i,:}$ denotes the $i-$th row of identity matrix of size $N$. The function $ h^i_{k,X,\lambda}: \mathcal{X} \rightarrow \mathbb{R}$ is a kernel-smoothed approximation of $\mathbbm{1}_{\{x^i\}}:\mathcal{X} \rightarrow \{0,1 \}$, and thus $ h^i_{k,X,\lambda}(x)$ represents the kernel-smoothed ``membership'' of $x$ to the set $\{ x^i\}$. Given a response variable $y^i \in \mathbb{R}^p$ associated to data point $x^i$, a regression model based on the affine combination of response variables can be defined as in the following:
   \begin{IEEEeqnarray}{rCl}         
\label{eq_240220131317}A(x) & = & \frac{ h^1_{k,X,\lambda}(x)}{\sum_{i=1}^N  h^i_{k,X,\lambda}(x)}y^1 + \cdots +  \frac{ h^N_{k,X,\lambda}(x)}{\sum_{i=1}^N  h^i_{k,X,\lambda}(x)}y^N,
  \end{IEEEeqnarray} 
where $ h^i_{k,X,\lambda}(x)/\sum_{i=1}^N  h^i_{k,X,\lambda}(x)$ represents kernel-smoothed relative membership of $x$ to the set $\{x^i\}$. The regression model $A:\mathcal{X} \rightarrow \mathrm{aff}(\{y^1,\cdots,y^N \})$ maps a point $x \in \mathcal{X}$ onto the affine hull of $\{y^1,\cdots,y^N\}$ through an affine combination where the coefficients are computed from the functions in RKHS $\mathcal{H}_{k}(\mathcal{X})$. The model $A$ is referred to as a kernel affine hull machine in this study.   
\subsection{An Affine Hull Machine}\label{sec_an_affine_hull_machine}   
For a finite set $\{y^1,\cdots,y^N\} \subset \mathbb{R}^p$ of $N$ pairwise distinct points, we aim to learn representation of data points in RKHS. For this, we consider a special case of the regression model $A:\mathcal{X} \rightarrow \mathrm{aff}(\{y^1,\cdots,y^N \})$ (defined as in (\ref{eq_240220131317})) for $\mathcal{X}  =  \{ Py \; \mid \; y \in \mathbb{R}^p \}$, where $P \in \mathbb{R}^{n \times p}\: (n \leq p)$ is an encoding matrix such that product $Py$ is a lower-dimensional encoding for $y$. In this case, the indicator function $\mathbbm{1}_{\{Py^i\}}$ is approximated through a function in RKHS fitted on the ordered pairs $\{ \left(Py^j, \mathbbm{1}_{\{Py^i\}}(Py^j)\right) \; \mid \; j \in \{1,\cdots,N \} \}$. This leads to the development of a kernel affine hull machine defined formally in Definition~\ref{def_affine_hull_model}.     
\begin{definition}[Kernel Affine Hull Machine (KAHM)]\label{def_affine_hull_model}
Given a finite number of samples: $Y = \left[\begin{IEEEeqnarraybox*}[][c]{,c/c/c,} y^1 & \cdots & y^N \end{IEEEeqnarraybox*} \right]^T$ with $y^1,\cdots,y^N \in \mathbb{R}^p$ and a subspace dimension $n \leq p$; a kernel affine hull machine $\mathcal{A}_{Y,n}: \mathbb{R}^p \rightarrow \mathrm{aff}(\{y^1,\cdots,y^N \})$ maps an arbitrary point $y \in \mathbb{R}^p$ onto the affine hull of $\{y^1,\cdots,y^N\}$ such that
  \begin{IEEEeqnarray}{rCl}
\label{eq_301220221131}\mathcal{A}_{Y,n}(y) & := & \frac{h_{k_{\theta},YP^T,\lambda^*}^1(Py)}{\sum_{i=1}^Nh_{k_{\theta},YP^T,\lambda^*}^i(Py)}y^1 + \cdots + \frac{h_{k_{\theta},YP^T,\lambda^*}^N(Py)}{\sum_{i=1}^Nh_{k_{\theta},YP^T,\lambda^*}^i(Py)}y^N.
   \end{IEEEeqnarray} 
 Here,
\begin{itemize}
\item $P \in \mathbb{R}^{n \times p}\: (n \leq p)$ is an encoding matrix such that product $Py$ is a lower-dimensional encoding for $y$. For a given subspace dimension $n$, $P$ is defined by setting the $i-$th row of $P$ as equal to transpose of eigenvector corresponding to $i-$th largest eigenvalue of sample covariance matrix of dataset $\{y^1,\cdots,y^N \}$.     
\item $k_{\theta}: \mathcal{X} \times \mathcal{X} \rightarrow \mathbb{R}$ is a positive-definite real-valued kernel on $\mathcal{X}$ with a corresponding reproducing kernel Hilbert space $\mathcal{H}_{k_{\theta}}(\mathcal{X})$ where 
   \begin{IEEEeqnarray}{rCl} 
\mathcal{X} & = & \{ Py \; \mid \; y \in \mathbb{R}^p \}.
   \end{IEEEeqnarray}   
 The kernel function $k_{\theta}$ is chosen of Gaussian type:
   \begin{IEEEeqnarray}{rCl}
\label{eq_260120231329}k_{\theta}(x^i,x^j) & := & \exp\left(-\frac{1}{2n}(x^i-x^j)^T\theta^{-1}(x^i-x^j)\right)    
  \end{IEEEeqnarray} 
where $\theta$ is sample covariance matrix of dataset $\{Py^1,\cdots,Py^N \}$ defined as
   \begin{IEEEeqnarray}{rCl}
   \theta &  = & \frac{1}{N-1}P\left(Y - \mathbf{1}_N \frac{\sum_{i=1}^N(y^i)^T}{N}\right)^T\left(Y - \mathbf{1}_N \frac{\sum_{i=1}^N(y^i)^T}{N}\right)P^T.
  \end{IEEEeqnarray}               
\item The function $h^i_{k_{\theta},YP^T,\lambda}: \mathcal{X} \rightarrow \mathbb{R}$, such that $h^i_{k_{\theta},YP^T,\lambda} \in \mathcal{H}_{k_{\theta}}(\mathcal{X})$, approximates the indicator function $\mathbbm{1}_{\{Py^i\}}: \mathcal{X} \rightarrow \{0,1 \}$ as the solution of following kernel regularized least squares problem:  
  \begin{IEEEeqnarray}{rCl}
h^i_{k_{\theta},YP^T,\lambda} & = & \arg \; \min_{f \in \mathcal{H}_{k_{\theta}}(\mathcal{X})} \; \left( \sum_{j=1}^N \left |\mathbbm{1}_{\{Py^i\}}(Py^j) - f(Py^j) \right |^2 + \lambda \left \| f \right \|^2_{\mathcal{H}_k(\mathcal{X})} \right),\; \lambda \in \mathbb{R}_+. \IEEEeqnarraynumspace
  \end{IEEEeqnarray}
The solution follows as
  \begin{IEEEeqnarray}{rCl}
\label{eq_250220231734} h^i_{k_{\theta},YP^T,\lambda}(\cdot) & = & (I_N)_{i,:} \left(K_{YP^T} + \lambda I_N \right)^{-1}  \left[\begin{IEEEeqnarraybox*}[][c]{,c/c/c,} k_{\theta}(\cdot,Py^1) & \cdots & k_{\theta}(\cdot,Py^N)\end{IEEEeqnarraybox*} \right]^T
  \end{IEEEeqnarray} 
where $(I_N)_{i,:}$ denotes the $i-$th row of identity matrix of size $N$ and $K_{YP^T}$ is $N \times N$ kernel matrix with its $(i,j)-$th element defined as
   \begin{IEEEeqnarray}{rCl}
\label{eq_010120231042}(K_{YP^T})_{ij}& := & k_{\theta}(Py^i,Py^j).
  \end{IEEEeqnarray}
The value $ h^i_{k_{\theta},YP^T,\lambda}(Py)$ represents the kernel-smoothed membership of point $Py$ to the set $\{ Py^i\}$. 
\item The regularization parameter $\lambda^* \in \mathbb{R}_+$ is given as
  \begin{IEEEeqnarray}{rCl}
\label{eq_010120231034}\lambda^* & = &  \hat{e} + \frac{2}{pN}\|Y \|^2_F, 
     \end{IEEEeqnarray}   
where $\hat{e}$ is the unique fixed point of $\mathcal{R}_{k_{\theta},YP^T,Y}$ such that
  \begin{IEEEeqnarray}{rCl}
\label{eq_090120230831}\hat{e} & = & \mathcal{R}_{k_{\theta},YP^T,Y}(\hat{e},\frac{2}{pN}\|Y \|^2_F),
     \end{IEEEeqnarray}   
with $\mathcal{R}_{k_{\theta},YP^T,Y}: \mathbf{R}_{+} \times \mathbf{R}_{+} \rightarrow \mathbf{R}_{+}$ defined as
   \begin{IEEEeqnarray}{rCl}
\label{eq_090120230832}\mathcal{R}_{k_{\theta},YP^T,Y}(e,\tau)& := &   \frac{1}{pN} \sum_{j=1}^p \|(Y)_{:,j} - K_{YP^T} \left(K_{YP^T} + (e+\tau) I_N \right)^{-1} (Y)_{:,j}\|^2.
     \end{IEEEeqnarray} 
The following iterations
 \begin{IEEEeqnarray}{rCl}
e|_{it+1} & = & \mathcal{R}_{k_{\theta},YP^T,Y}(e|_{it},\frac{2}{pN}\|Y \|^2_F),\; it \in \{0,1,\cdots \}  \\
e|_0 & \in & (0,\frac{1}{pN} \|Y \|^2_F)  
  \end{IEEEeqnarray}
converge to $\hat{e}$.     
\item The image of $\mathcal{A}_{Y,n}$ defines a region in the affine hull of $\{y^1,\cdots,y^N\}$. That is,
 \begin{IEEEeqnarray}{rCCCl}
\label{eq_738825.6428} \mathcal{A}_{Y,n}[\mathbb{R}^p]& := & \{ \mathcal{A}_{Y,n}(y) \; \mid \; y \in \mathbb{R}^p  \}  & \subset & \mathrm{aff}(\{y^1,\cdots,y^N \}).  
  \end{IEEEeqnarray}   
Fig.~\ref{fig_KAHM_examples} provides examples of two dimensional datasets and KAHM images.    
\begin{figure}
\centerline{\subfigure[first example]{\includegraphics[width=0.33\textwidth]{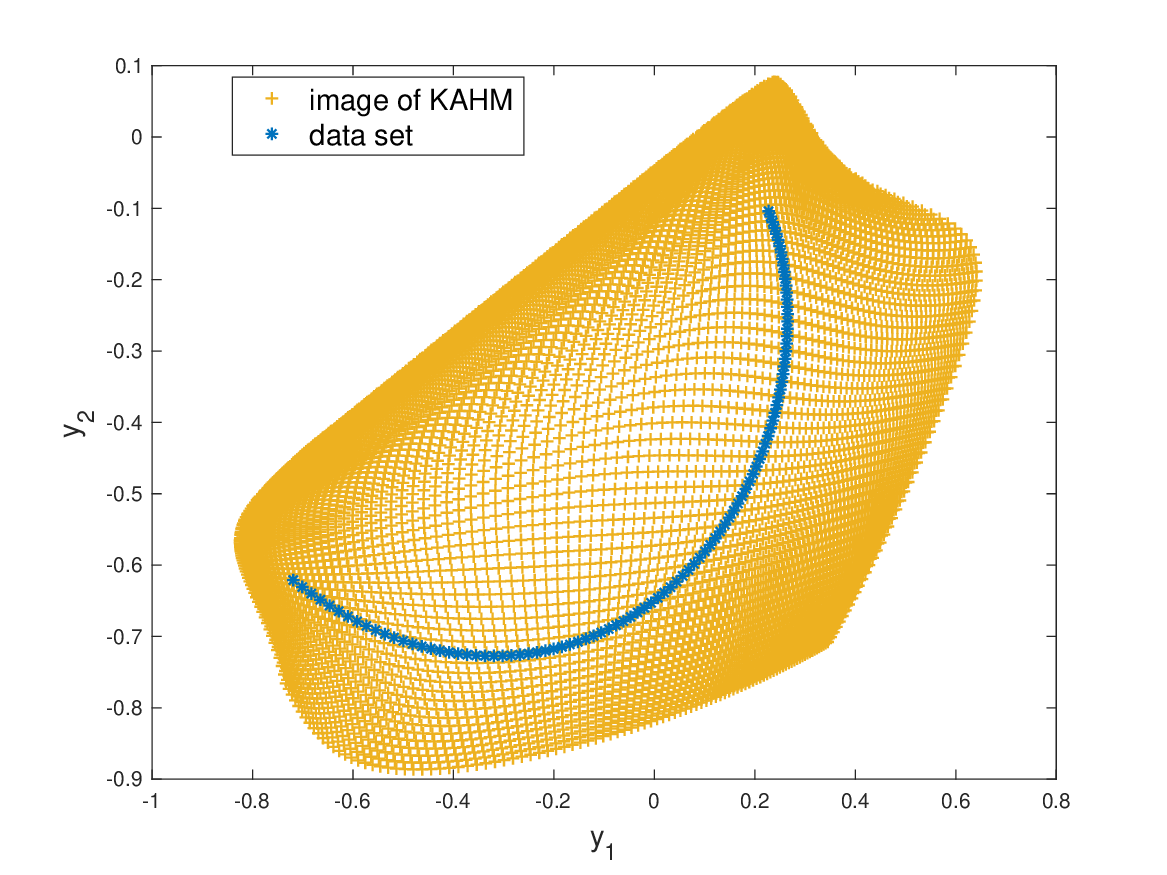}\label{fig_KAHM_example_1}} \hfil 
\subfigure[second example]{\includegraphics[width=0.33\textwidth]{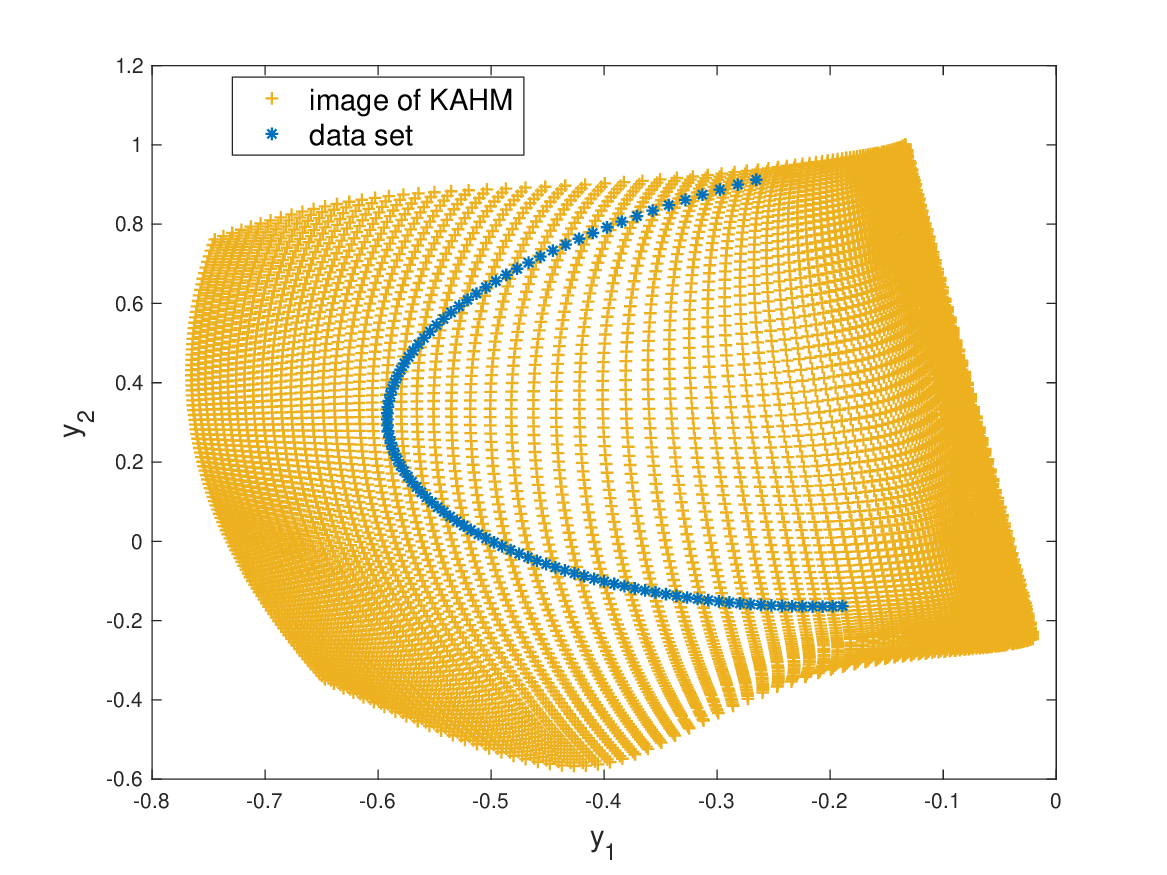}\label{fig_KAHM_example_2}} \hfil \subfigure[third example]{\includegraphics[width=0.33\textwidth]{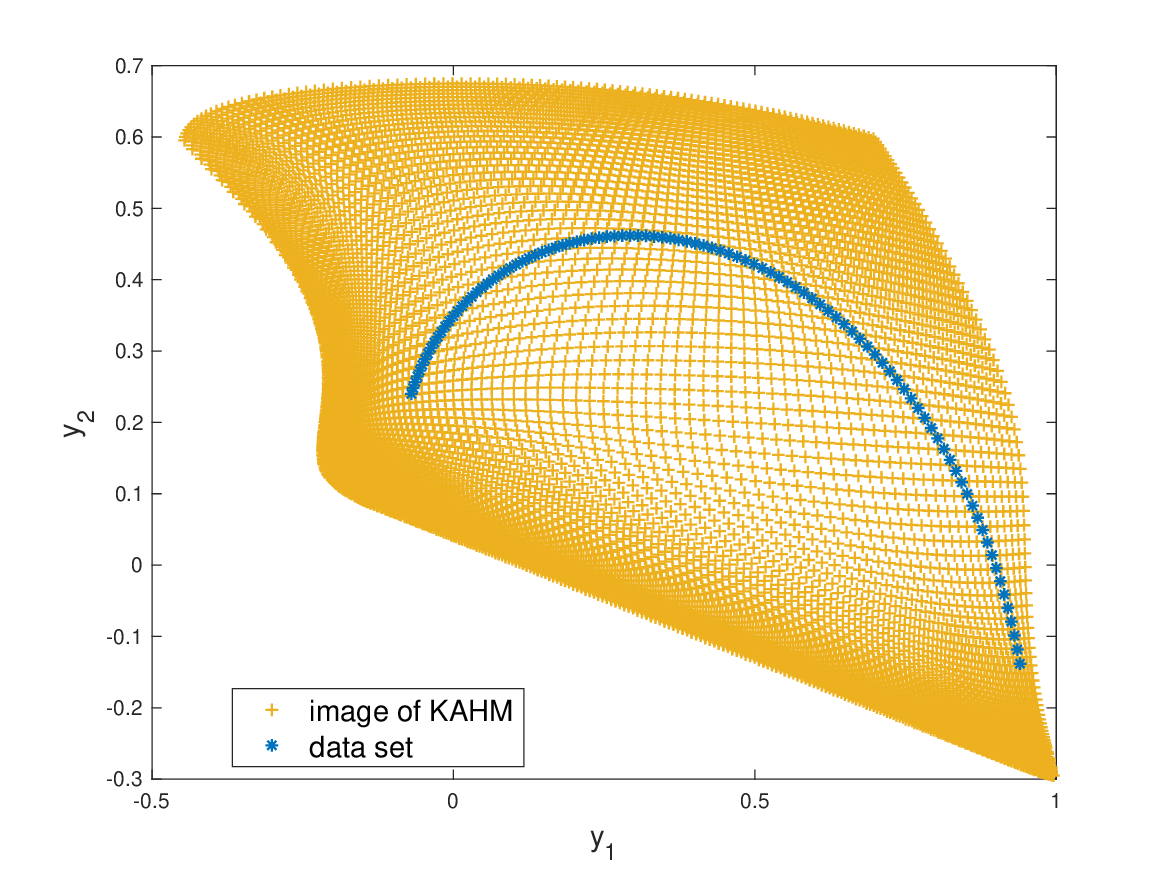}\label{fig_KAHM_example_3}}  }
\caption{A few examples of two dimensional data sets and KAHM images.}
\label{fig_KAHM_examples}
\end{figure}
\end{itemize}
 \end{definition}
 \begin{remark}[Computational complexity]\label{remark_computational_cost}
The computational complexity of the KAHM is asymptotically dominated by the computation of inverse of the $N \times N$ dimensional matrix $\left(K_{YP^T} + \lambda I_N \right)$. Therefore, computational complexity of the KAHM is given as $\mathcal{O}(N^3)$.      
\end{remark}
KAHM is a bounded function as stated in Theorem~\ref{result_kahm_bounded_function} in the following.
\begin{theorem}\label{result_kahm_bounded_function}
The KAHM $\mathcal{A}_{Y,n}$,  associated to $Y = \left[\begin{IEEEeqnarraybox*}[][c]{,c/c/c,} y^1 & \cdots & y^N \end{IEEEeqnarraybox*} \right]^T$ with $y^1,\cdots,y^N \in \mathbb{R}^p$,  is a bounded function on $\mathbb{R}^p$ such that for any $y \in \mathbb{R}^p$, 
\begin{IEEEeqnarray}{rCCCl}
\label{eq_100120231400} \| \mathcal{A}_{Y,n}(y)\| & < & \left\|Y \right \|_2 \frac{\lambda^* + \mu_{max}(K_{YP^T} )}{\lambda^* + \mu_{min}(K_{YP^T} )}& < &  \left\|Y \right \|_2\left(1 + \frac{pN^2}{2\|Y\|_F^2} \right)
\end{IEEEeqnarray}  
where $\lambda^*  \in \mathbb{R}_+$ is defined as in (\ref{eq_010120231034}) and $K_{YP^T}$ is defined as in (\ref{eq_010120231042}). Thus, the image of $\mathcal{A}_{Y,n}$ is bounded such that 
 \begin{IEEEeqnarray}{rCl}
 \mathcal{A}_{Y,n}[\mathbb{R}^p]& \subset &\left\{ y \in \mathbb{R}^p \; \mid \;  \| y\| < \left\|Y \right \|_2 \frac{\lambda^* + \mu_{max}(K_{YP^T} )}{\lambda^* + \mu_{min}(K_{YP^T} )} \right \}.
   \end{IEEEeqnarray} 
\begin{proof}
The proof is provided in Appendix~\ref{appendix1_5}.
\end{proof} 
\end{theorem}   
A distance function can be associated to KAHM as in Definition~\ref{def_distance_function_KAHM}:
\begin{definition}[A Distance Function Induced by KAHM]\label{def_distance_function_KAHM}
Given a KAHM $\mathcal{A}_{Y,n}$, the distance of an arbitrary point $y\in \mathbb{R}^p$ from its image under $\mathcal{A}_{Y,n}$ is given as
 \begin{IEEEeqnarray}{rCl}
 \Gamma_{\mathcal{A}_{Y,n}}(y) & := & \left \| y - \mathcal{A}_{Y,n}(y) \right \|.
   \end{IEEEeqnarray}    
\end{definition}
A significant property of the distance function is that its value at a point can not be arbitrary large provided that the point is \emph{sufficiently} close to the samples represented by KAHM. This property is being stated by Theorem~\ref{result_ratio_distances} in the following.   
\begin{theorem}\label{result_ratio_distances}
The ratio of the distance of a point $y\in \mathbb{R}^p$ from its image under $\mathcal{A}_{Y,n}$ to the distance of $y$ from $\{y^1,\cdots,y^N\}$ evaluated as $\left \|\left[\begin{IEEEeqnarraybox*}[][c]{,c/c/c,} y - y^1 & \cdots & y - y^N \end{IEEEeqnarraybox*} \right] \right \|_2$ remains upper bounded as
 \begin{IEEEeqnarray}{rCCCl}
\label{eq_100120231432} \frac{ \Gamma_{\mathcal{A}_{Y,n}}(y) }{\left \|\left[\begin{IEEEeqnarraybox*}[][c]{,c/c/c,} y - y^1 & \cdots & y - y^N \end{IEEEeqnarraybox*} \right] \right \|_2} & < & \frac{\lambda^* + \mu_{max}(K_{YP^T} )}{\lambda^* + \mu_{min}(K_{YP^T} )}  & < &  1 + \frac{p N^2}{2 \|Y\|_F^2}
   \end{IEEEeqnarray} 
where $\lambda^*  \in \mathbb{R}_+$ is defined as in (\ref{eq_010120231034}), $K_{YP^T}$ is defined as in (\ref{eq_010120231042}), and $Y = \left[\begin{IEEEeqnarraybox*}[][c]{,c/c/c,} y^1 & \cdots & y^N \end{IEEEeqnarraybox*} \right]^T$. 
\begin{proof}
The proof is provided in Appendix~\ref{appendix2}.
\end{proof}    
   \end{theorem}
Theorem~\ref{result_ratio_distances} signifies that if a point $y$ is close to points $\{y^1,\cdots,y^N\}$, then the value $\Gamma_{\mathcal{A}_{Y,n}}(y)$ can not be large. Thus, a large value of the distance function at a point $y$ indicates that $y$ is at far distance from $\{y^1,\cdots,y^N\}$.        
\section{KAHM for Data Representation Learning}\label{sec_kahm_compositions}
For data representation learning, KAHM based models are introduced (in Section~\ref{sec_220320231642}) and applied to the classification problem (in Section~\ref{sec_220320231647}). To evaluate the risk of membership inference attack through KAHM based classifier, a membership-inference score is defined (in Section~\ref{subsec_mis}).    
\subsection{Wide and Conditionally Deep KAHMs}\label{sec_220320231642}
\begin{definition}[Conditionally Deep Kernel Affine Hull Machine]\label{def_DKAHM}
Given a finite number of samples: $Y = \left[\begin{IEEEeqnarraybox*}[][c]{,c/c/c,} y^1 & \cdots & y^N \end{IEEEeqnarraybox*} \right]^T$ with $y^1,\cdots,y^N \in \mathbb{R}^p$, a subspace dimension $n \leq p$, and number of layers $L \leq n$; a conditionally deep kernel affine hull machine $\mathcal{D}_{Y,n,L}:\mathbb{R}^p \rightarrow \mathrm{aff}(\{y^1,\cdots,y^N\})$ maps an arbitrary point $y \in \mathbb{R}^p$ onto the affine hull of $\{y^1,\cdots,y^N\}$ through a nested composition of kernel affine hull machines (as illustrated in Fig.~\ref{fig_deep_KAHM}) such that  
\begin{IEEEeqnarray}{rCl}
\label{eq_220120231153}\mathcal{D}_{Y,n,L}(y) & = &  \mathcal{M}_{Y,n,\hat{l}(y)}(y),\\
\label{eq_220120231154}\mathcal{M}_{Y,n,l}(y) & = & \left(\mathcal{A}_{ Y,n-l+1} \circ \cdots \circ \mathcal{A}_{ Y,n-1} \circ \mathcal{A}_{ Y,n}\right)(y),\\ 
\label{eq_220120231155}\hat{l}(y) & = & \arg \; \min_{l \in \{1,2,\cdots,L \}} \;\left \| y - \mathcal{M}_{Y,n,l}(y) \right \|,
  \end{IEEEeqnarray}
where $\mathcal{A}_{Y,\cdot}$ is a KAHM (Definition~\ref{def_affine_hull_model}) and $\mathcal{M}_{Y,n,l}(y)$ is the image of $y$ onto the affine hull of $\{y^1,\cdots,y^N\}$ by the $l-$th layer and the output $\mathcal{D}_{Y,n,L}(y)$ is equal to the image of $y$ onto the affine hull of $\{y^1,\cdots,y^N\}$ by $\hat{l}-$th layer (which is the layer resulting in minimum Euclidean distance between input vector $y$ and its image onto the affine hull of $\{y^1,\cdots,y^N\}$). The deep KAHM discovers layers of increasingly abstract data representation with lowest-level data features being modeled by first layer and the highest-level by end layer. Fig.~\ref{fig_cond_deep_KAHM_example} illustrates through an example the data representation learning at varying abstraction levels across different layers such that $\mathcal{M}_{Y,n,1}(y)$ is least abstract representation and $\mathcal{M}_{Y,n,L}(y)$ is most abstract representation of the input vector $y$.
\begin{figure}
\centering
\scalebox{0.75}{
\begin{tikzpicture}
 \draw[fill=gray!5] (0.5,-4.5) rectangle (12.5,1); 
  \draw[arrows=->](0,0)--(1.5,0) node[below,pos=0]{$y$};
   \draw (1.5,-0.5) rectangle (3,0.5) node[pos=0.5]{$\mathcal{A}_{Y,n}$};  
     \draw[arrows=->](3,0)--(4,0);
 \draw[dashed](4,0)--(6.5,0);
       \draw[arrows=->](6.5,0)--(7,0);
 \draw (7,-0.5) rectangle (8.75,0.5) node[pos=0.5]{$\mathcal{A}_{Y,n-L+1}$};  
 \draw[arrows=->](8.75,0)--(10.5,0) node[above, pos = 0.5]{$\mathcal{M}_{Y,n,L}(y)$};  
 \draw (10.5,-4.1) rectangle (12,0.5) node[pos=0.5]{$\begin{array}{c}\mbox{output} \\ \mbox{layer} \end{array}$};      
 \draw[dashed](10,-0.5)--(10,-2);
   \draw[](3.5,0)--(3.5,-3);
     \draw[arrows=->](3.5,-3)--(10.5,-3) node[above, pos = 0.875]{$\mathcal{M}_{Y,n,1}(y)$}; 
 \draw[](1,0)--(1,-3.75);    
 \draw[arrows=->](1,-3.75)--(10.5,-3.75) node[above, pos = 0.935]{$y$}; 
   \draw[arrows=->](12,-1.8)--(13.5,-1.8) node[below, pos = 1]{$\mathcal{D}_{Y,n,L}(y)$}; 
 \draw (5.5,-7.5) rectangle (7.5,-6.5) node[pos=0.5]{$\mathcal{D}_{Y,n,L}$};  
    \draw[arrows=->](4,-7)--(5.5,-7) node[left,pos=0]{$y$};
  \draw[arrows=->](7.5,-7)--(8.5,-7) node[right, pos = 1]{$\mathcal{D}_{Y,n,L}(y)$}; 
   \path[fill=gray!10](0.5,-4.5)--(5.5,-6.5)--(7.5,-6.5)--(12.5,-4.5)--cycle;
\end{tikzpicture}
}
\caption{The structure of conditionally deep $L-$layered kernel affine hull machine.}
\label{fig_deep_KAHM}
\end{figure}
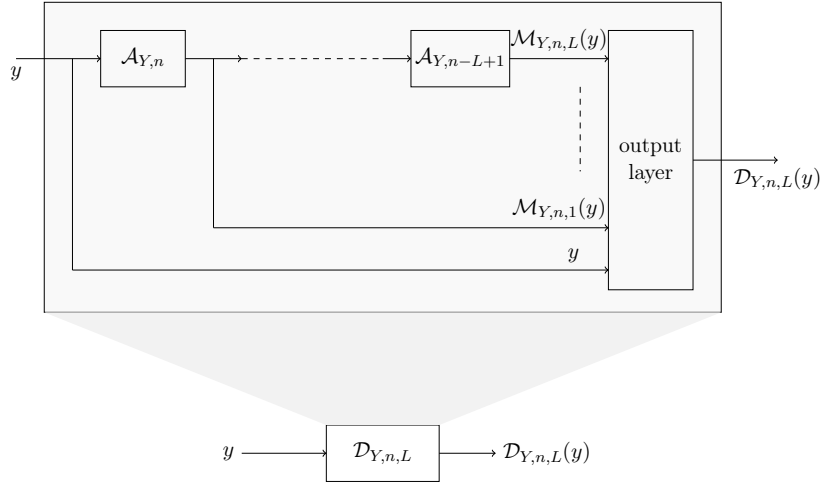
\begin{figure}
\centering
\includegraphics[width=\textwidth]{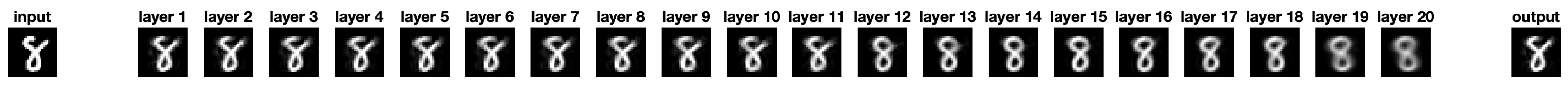}
\caption{A dataset $Y$ consisting of 1000 randomly chosen samples of digit 8 from MNIST dataset was considered. Corresponding to an input sample $y$ (displayed at extreme left of the figure), the outputs of different layers (i.e. $\mathcal{M}_{Y,n,1}(y), \cdots,\mathcal{M}_{Y,n,20}(y)$) have been displayed. The output of conditionally deep KAHM (i.e. $\mathcal{D}_{Y,n,20}(y)$) has been displayed at extreme right of the figure.}
\label{fig_cond_deep_KAHM_example}
\end{figure}
\end{definition}
\begin{definition}[A Distance Function Induced by Conditionally Deep KAHM]\label{def_distance_function_cond_deep_KAHM}
Given a conditionally deep KAHM $\mathcal{D}_{Y,n,L}$, the distance of an arbitrary point $y\in \mathbb{R}^p$ from its image under $\mathcal{D}_{Y,n,L}$ is given as
 \begin{IEEEeqnarray}{rCl}
 \Gamma_{\mathcal{D}_{Y,n,L}}(y) & := & \left \| y - \mathcal{D}_{Y,n,L}(y) \right \|.
   \end{IEEEeqnarray}    
\end{definition}
\begin{theorem}\label{result_ratio_distances_cond_deep_KAHM}
The ratio of the distance of a point $y\in \mathbb{R}^p$ from its image under $\mathcal{D}_{Y,n,L}$ to the distance of $y$ from $\{y^1,\cdots,y^N\}$ evaluated as $\left \|\left[\begin{IEEEeqnarraybox*}[][c]{,c/c/c,} y - y^1 & \cdots & y - y^N \end{IEEEeqnarraybox*} \right] \right \|_2$ remains upper bounded as
 \begin{IEEEeqnarray}{rCCCl}
\label{eq_220120231130} \frac{ \Gamma_{\mathcal{D}_{Y,n,L}}(y) }{\left \|\left[\begin{IEEEeqnarraybox*}[][c]{,c/c/c,} y - y^1 & \cdots & y - y^N \end{IEEEeqnarraybox*} \right] \right \|_2} & \leq & \frac{ \Gamma_{\mathcal{A}_{Y,n}}(y) }{\left \|\left[\begin{IEEEeqnarraybox*}[][c]{,c/c/c,} y - y^1 & \cdots & y - y^N \end{IEEEeqnarraybox*} \right] \right \|_2}  & < &  1 + \frac{p N^2}{2 \|Y\|_F^2}
   \end{IEEEeqnarray} 
where $Y = \left[\begin{IEEEeqnarraybox*}[][c]{,c/c/c,} y^1 & \cdots & y^N \end{IEEEeqnarraybox*} \right]^T$.
\begin{proof}
The proof is provided in Appendix~\ref{appendix3}.
\end{proof}
   \end{theorem}
For big datasets, the total data can be partitioned into subsets and corresponding to each data-subset a separate KAHM can be built to avoid computational challenges associated to big datasets. This motivates to introduce a wide condition deep KAHM in the following:
\begin{definition}[Wide Conditionally Deep Kernel Affine Hull Machine]\label{def_WDKAHM}
Given a big but finite number of samples: $Y = \left[\begin{IEEEeqnarraybox*}[][c]{,c/c/c,} y^1 & \cdots & y^N \end{IEEEeqnarraybox*} \right]^T$ with $y^1,\cdots,y^N \in \mathbb{R}^p$, a subspace dimension $n \leq p$, number of layers $L \leq n$, and number of branches $S \leq N$; a wide conditionally deep kernel affine hull machine $\mathcal{W}_{Y,n,L,S}:\mathbb{R}^p \rightarrow \mathrm{aff}(\{y^1,\cdots,y^N\})$ maps an arbitrary point $y \in \mathbb{R}^p$ onto the affine hull of $\{y^1,\cdots,y^N\}$ through a parallel composition of conditionally deep $L-$layered kernel affine hull machines (as illustrated in Fig.~\ref{fig_wide_deep_KAHM}) such that     
\begin{IEEEeqnarray}{rCl}
\mathcal{W}_{Y,n,L,S}(y) & = & \mathcal{D}_{Y_{\hat{s}(y)},n,L}(y),\\
\hat{s}(y) & = & \arg \; \min_{s \in \{1,2,\cdots,S \}} \;\left \| y - \mathcal{D}_{Y_s,n,L}(y) \right \|,\\
\label{eq_220120231804}Y_s & = & \left[\begin{IEEEeqnarraybox*}[][c]{,c/c/c,} y^{1,1} & \cdots & y^{N_s,s} \end{IEEEeqnarraybox*} \right]^T\\
\left\{\{y^{1,1},\cdots,y^{N_1,1}\},\cdots, \{y^{1,S},\cdots,y^{N_S,S}\}\right\} & = & \mathrm{clustering}(\{y^1,\cdots,y^N\}, S),
  \end{IEEEeqnarray}  
where $\mathcal{D}_{Y_{\cdot},n,L}$ is the conditionally deep KAHM (Definition~\ref{def_DKAHM}) and $ \mathrm{clustering}(\{y^1,\cdots,y^N\}, S)$ represents $k-$means clustering into $S$ subsets, where $S$ can be chosen e.g. as equal to rounding of $N/1000$ towards nearest integer i.e.
\begin{IEEEeqnarray}{rCl}
\label{eq_190220231832}S & = & \lceil N/1000 \rceil.
  \end{IEEEeqnarray} 
Each of $S$ data clusters leads to a separate conditionally deep KAHM and the output $\mathcal{W}_{Y,n,L,S}(y)$ is equal to the image of $y$ onto the affine hull of $\{y^1,\cdots,y^N\}$ by $\hat{s}-$th conditionally deep KAHM (which is the KAHM resulting in minimum Euclidean distance between input vector $y$ and its image onto the affine hull of $\{y^1,\cdots,y^N\}$).    
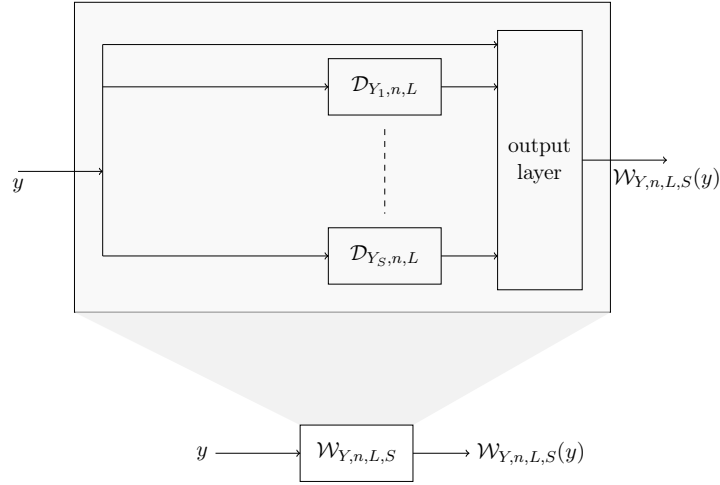
\begin{figure}
\centering
\scalebox{0.75}{
\begin{tikzpicture}
  \draw[fill=gray!5] (1,-4.5) rectangle (10.5,1); 
    \draw[arrows=->](3.5,-7)--(5,-7) node[left,pos=0]{$y$};
  \draw[arrows=->](7,-7)--(8,-7) node[right, pos = 1]{$\mathcal{W}_{Y,n,L,S}(y)$}; 
   \path[fill=gray!10](1,-4.5)--(5,-6.5)--(7,-6.5)--(10.5,-4.5)--cycle;
    \draw (5,-7.5) rectangle (7,-6.5) node[pos=0.5]{$\mathcal{W}_{Y,n,L,S}$};  
      \draw[arrows=->](0,-2)--(1.5,-2) node[below,pos=0]{$y$};
    \draw[](1.5,-2)--(1.5,-3.5); 
    \draw[arrows=->](1.5,-3.5)--(5.5,-3.5);
      \draw (5.5,-4) rectangle (7.5,-3) node[pos=0.5]{$\mathcal{D}_{Y_S,n,L}$};  
  \draw[](1.5,-2)--(1.5,0.25);     
   \draw[arrows=->](1.5,-0.5)--(5.5,-0.5);
         \draw (5.5,-1) rectangle (7.5,0) node[pos=0.5]{$\mathcal{D}_{Y_1,n,L}$};  
\draw[arrows=->](7.5,-0.5)--(8.5,-0.5);
\draw[arrows=->](7.5,-3.5)--(8.5,-3.5);
 \draw[dashed](6.5,-1.25)--(6.5,-2.75);
  \draw (8.5,-4.1) rectangle (10,0.5) node[pos=0.5]{$\begin{array}{c}\mbox{output} \\ \mbox{layer} \end{array}$};     
    \draw[arrows=->](1.5,0.25)--(8.5,0.25); 
   \draw[arrows=->](10,-1.8)--(11.5,-1.8) node[below, pos = 1]{$\mathcal{W}_{Y,n,L,S}(y)$}; 
\end{tikzpicture}}
\caption{The structure of $S-$branches wide conditionally deep $L-$layered KAHM.}
\label{fig_wide_deep_KAHM}
\end{figure}
\end{definition}
\begin{definition}[A Distance Function Induced by Wide Conditionally Deep KAHM]\label{def_distance_function_wide_cond_deep_KAHM}
Given a wide conditionally deep KAHM $\mathcal{W}_{Y,n,L,S}$, the distance of an arbitrary point $y\in \mathbb{R}^p$ from its image under $\mathcal{W}_{Y,n,L,S}$ is given as
 \begin{IEEEeqnarray}{rCl}
 \Gamma_{\mathcal{W}_{Y,n,L,S}}(y) & := & \left \| y - \mathcal{W}_{Y,n,L,S}(y) \right \|.
   \end{IEEEeqnarray}    
\end{definition}
\begin{theorem}\label{result_ratio_distances_wide_cond_deep_KAHM}
The ratio of the distance of a point $y\in \mathbb{R}^p$ from its image under $\mathcal{W}_{Y,n,L,S}$ to the distance of $y$ from $\{y^1,\cdots,y^N\}$ evaluated as $\left \|\left[\begin{IEEEeqnarraybox*}[][c]{,c/c/c,} y - y^1 & \cdots & y - y^N \end{IEEEeqnarraybox*} \right] \right \|_F$ remains upper bounded as  
\begin{IEEEeqnarray}{rCl}
\label{eq_220120231835} \frac{ \Gamma_{\mathcal{W}_{Y,n,L,S}}(y) }{\left \|\left[\begin{IEEEeqnarraybox*}[][c]{,c/c/c,} y - y^1 & \cdots & y - y^N \end{IEEEeqnarraybox*} \right] \right \|_F} & < & \min_{s \in \{1,2,\cdots,S \}} \;  \left( 1 + \frac{p N_s^2}{2 \|Y_s\|_F^2} \right)
\end{IEEEeqnarray}
where $Y_s$ is given as in (\ref{eq_220120231804}).  
\begin{proof}
The proof is provided in Appendix~\ref{appendix4}.
\end{proof}
\end{theorem}
\subsection{Classification Applications}\label{sec_220320231647}
The KAHM induced distance function, with a property as stated in Theorem~\ref{result_ratio_distances_wide_cond_deep_KAHM}, can be leveraged to define a classifier. The significance of inequality (\ref{eq_220120231835}) is that if a data point $y \in \mathbb{R}^p$ is close to samples $\{y^1,\cdots,y^N \}$, then the value $ \Gamma_{\mathcal{W}_{Y,n,L,S}}(y)$ remains small. This allows to define a classifier, as in Definition~\ref{def_KAHM_classifier}, via modeling each class's region through a separate wide conditionally deep KAHM and assigning to a point the label of the class with the minimum distance function value.
 \begin{definition}[KAHM Based Classifier]\label{def_KAHM_classifier}
Given a multi-class labelled dataset $\{ \left(Y_i,\mathrm{cl}^i \right) \; \mid \; Y_i = \left[\begin{IEEEeqnarraybox*}[][c]{,c/c/c,} y^{1,i} & \cdots & y^{N_i,i} \end{IEEEeqnarraybox*} \right]^T,\;y^{\cdot,i} \in \mathbb{R}^p,\;\mathrm{cl}^i \in \{1,2,\cdots,C \},\; i \in \{1,2,\cdots,C \} \}$, a KAHM based classifier $\mathcal{C}:\mathbb{R}^p \rightarrow \{1,2,\cdots,C \}$ is defined as
\begin{IEEEeqnarray}{rCl}
\label{eq_738832.478752}\mathcal{C}(y;\mathcal{W}_{Y_1,n,L,S_1},\cdots,\mathcal{W}_{Y_C,n,L,S_C}) & = &  \arg \; \min_{c \in \{1,2,\cdots,C \}} \;  \Gamma_{\mathcal{W}_{Y_c,n,L,S_c}}(y),
  \end{IEEEeqnarray}   
where $\mathcal{W}_{Y_c,n,L,S_c}$ is the wide conditionally deep KAHM (Definition~\ref{def_WDKAHM}) modeling the $c-$th class labelled data points and $\Gamma_{\mathcal{W}_{Y_c,n,L,S_c}}(\cdot)$ is the distance function (Definition~\ref{def_distance_function_wide_cond_deep_KAHM}) induced by $\mathcal{W}_{Y_c,n,L,S_c}$. The classifier assigns to an arbitrary point $y \in \mathbb{R}^p$ the label of the class which has the minimum distance between $y$ and $y$'s image onto the affine hull of samples of that class. 
\begin{figure}
\centerline{\subfigure[$Y_1$ and $\Gamma_{\mathcal{W}_{Y_1,2,5,1}}(\cdot)$]{\includegraphics[width=0.25\textwidth]{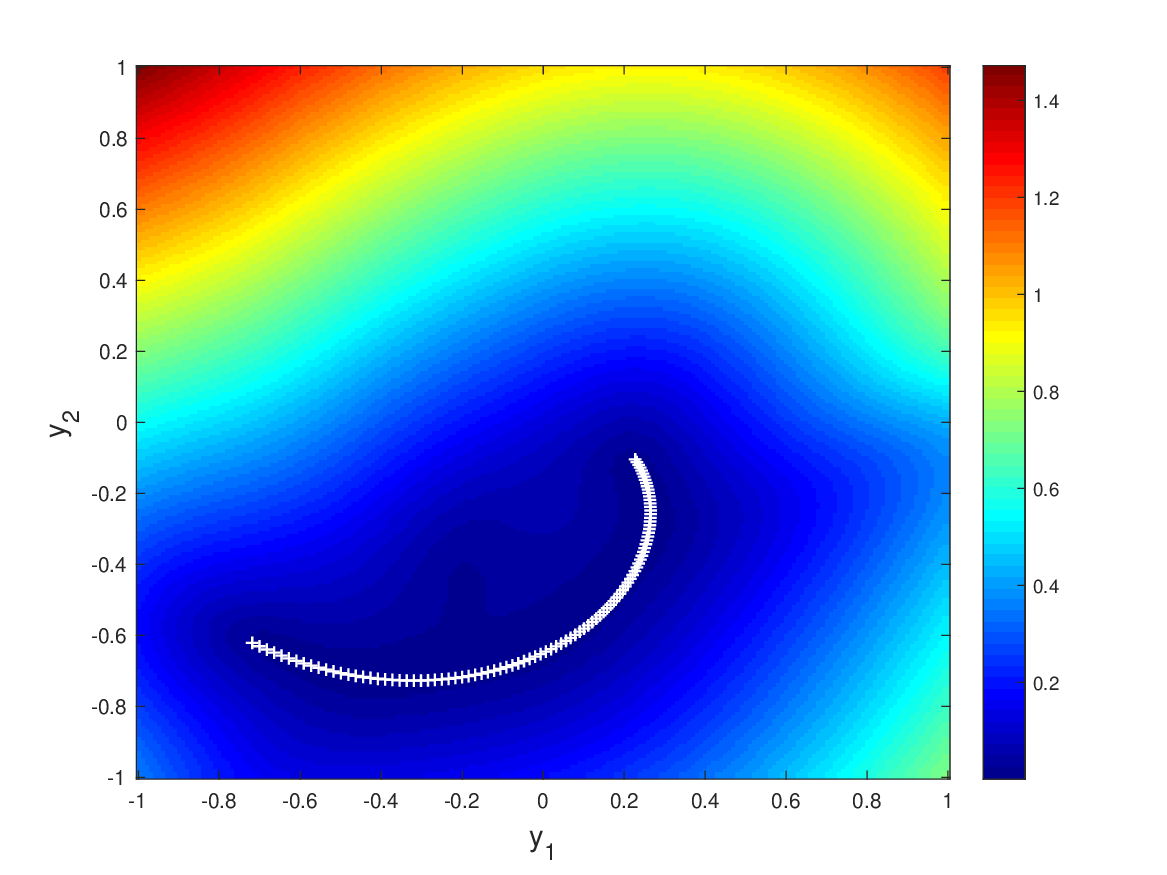}\label{fig_demo_classifier_1}} \hfil 
\subfigure[$Y_2$ and $\Gamma_{\mathcal{W}_{Y_2,2,5,1}}(\cdot)$]{\includegraphics[width=0.25\textwidth]{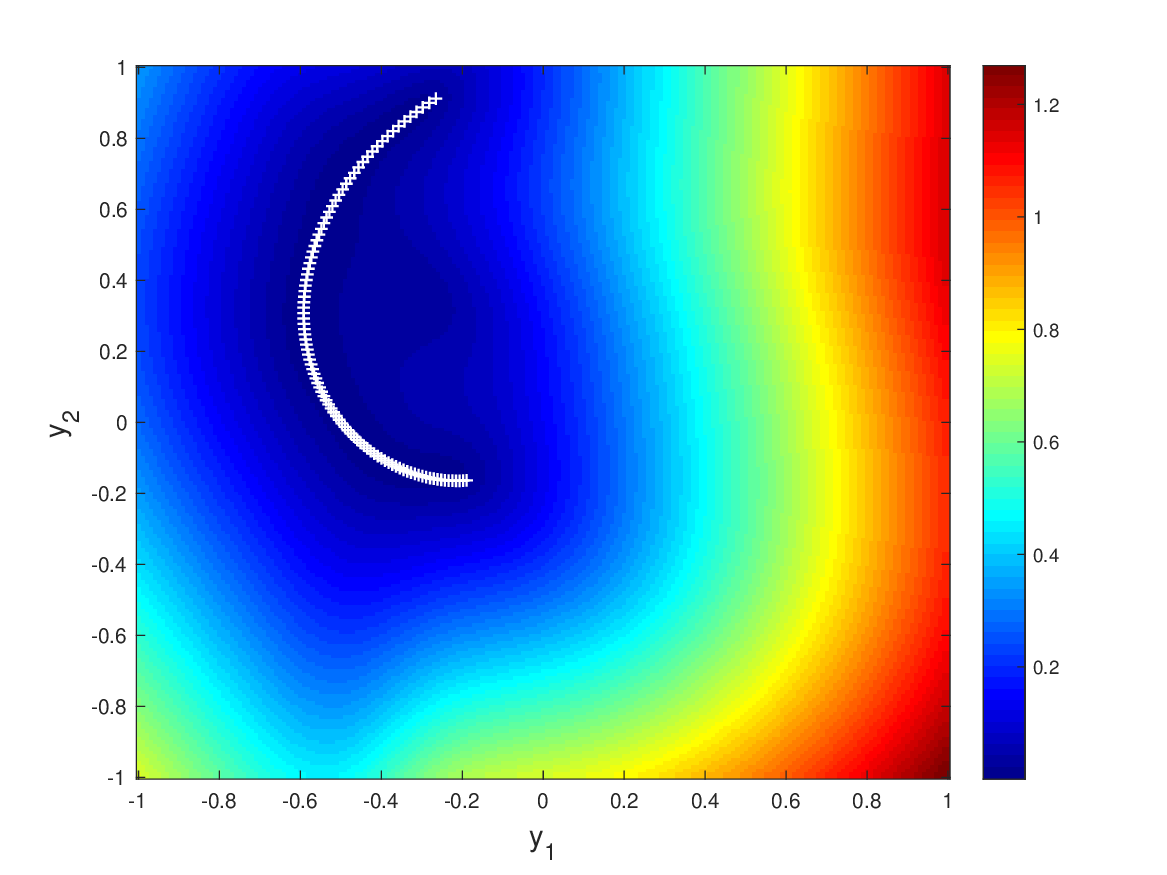}\label{fig_demo_classifier_2}} \hfil 
\subfigure[$Y_3$ and $\Gamma_{\mathcal{W}_{Y_3,2,5,1}}(\cdot)$]{\includegraphics[width=0.25\textwidth]{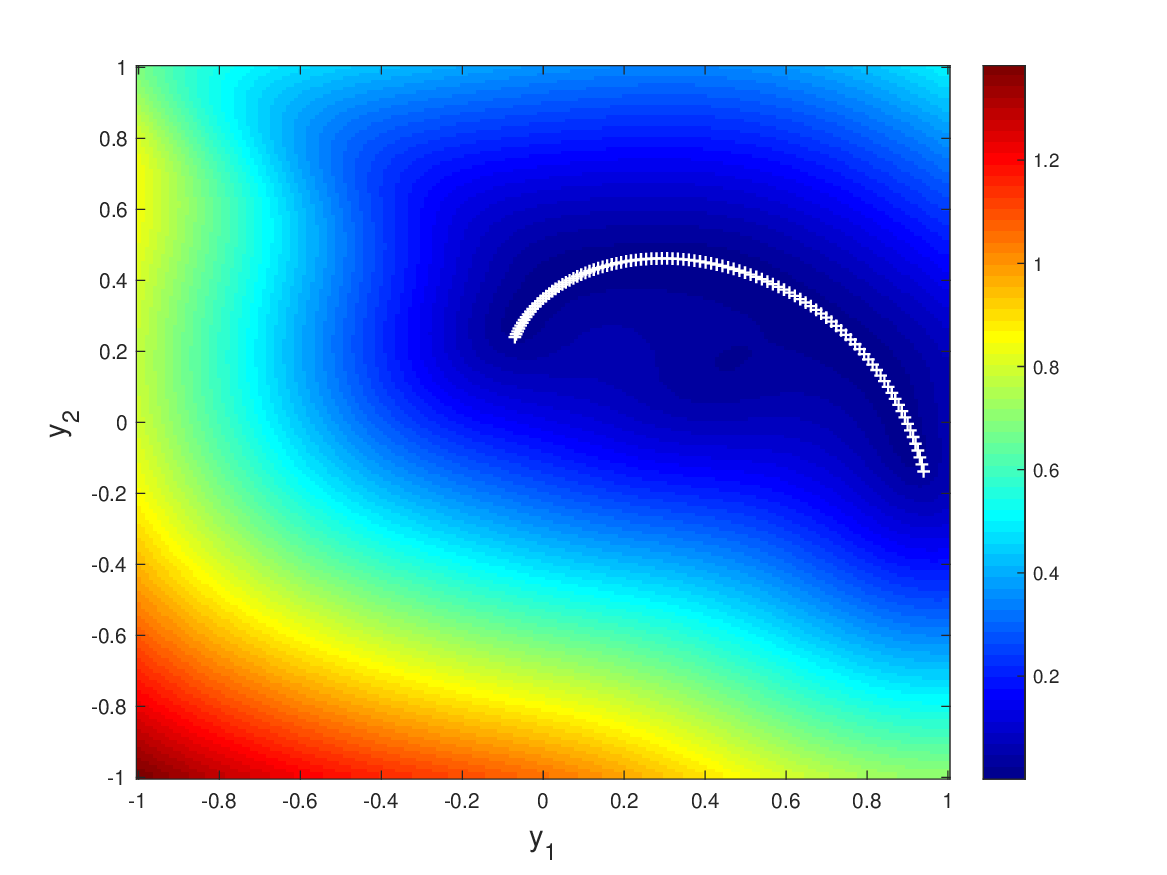}\label{fig_demo_classifier_3}}  \hfil
\subfigure[decision boundaries determined by the classifier~(\ref{eq_738832.478752})]{\includegraphics[width=0.25\textwidth]{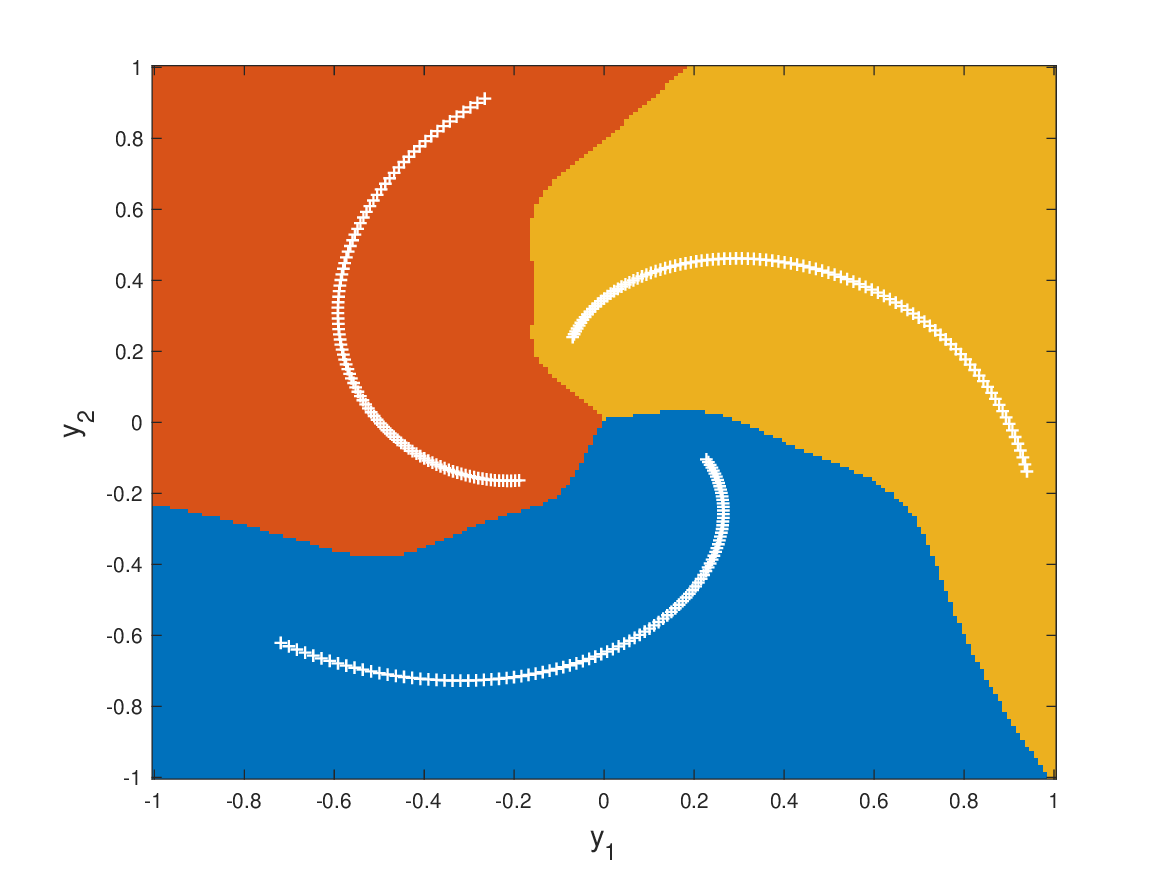}\label{fig_demo_classifier_4}}}
\caption{An example of KAHM based classifier built with a 2-dimensional dataset with 3 classes. The data samples have been displayed using `+' marker and the distance function $ \Gamma_{\mathcal{W}_{Y,n,L,S}}(\cdot)$ has been displayed as color plot.}
\label{fig_classifier_example}
\end{figure}
Fig.~\ref{fig_classifier_example} shows an example of a KAHM based classifier built with a 2-dimensional dataset with 3 classes. 
\end{definition}
The distance function can be further used to define a class-matching score as in Definition~\ref{def_matching_score}.     
\begin{definition}[Class-Matching Score]\label{def_matching_score}
Given the set $\{ \mathcal{W}_{Y_c,n,L,S_c}\}_{c=1}^C$ (where $\mathcal{W}_{Y_c,n,L,S_c}$ is the wide conditionally deep KAHM (Definition~\ref{def_WDKAHM}) modeling the $c-$th class labelled data points), the matching-score of a point $y \in \mathbb{R}^p$ to $c-$th class is defined as
\begin{IEEEeqnarray}{rCl}
ms(y;\mathcal{W}_{Y_1,n,L,S_1},\cdots,\mathcal{W}_{Y_C,n,L,S_C})& = & \exp\left(-\frac{|\Gamma_{\mathcal{W}_{Y_c,n,L,S_c}}(y) |^2}{\sum_{c=1}^C|\Gamma_{\mathcal{W}_{Y_c,n,L,S_c}}(y) |^2}\right)
   \end{IEEEeqnarray}  
where $\Gamma_{\mathcal{W}_{Y_c,n,L,S_c}}(\cdot)$ is the distance function (Definition~\ref{def_distance_function_wide_cond_deep_KAHM}) induced by $\mathcal{W}_{Y_c,n,L,S_c}$.    
\end{definition}
\subsection{Membership-Inference Score For KAHM Based Classifier}\label{subsec_mis}
The KAHM based classifier (Definition~\ref{def_KAHM_classifier}) is built using the training dataset,
\begin{IEEEeqnarray}{rCl}
\mathbf{D}_{trn} & = & \{y^{j,i} \; \mid \; j \in \{1,2,\cdots,N_i \},\; i \in \{1,2,\cdots,C \} \}.
  \end{IEEEeqnarray}  
As observed in~(\ref{eq_738832.478752}), the classifier assigns a label to a vector $y$ based on distance function values: $\{\Gamma_{\mathcal{W}_{Y_c,\cdot,\cdot,\cdot}}(y)\}_{c = 1}^C$. It is obvious that a point either belonging to or lying close to points represented by matrix $Y_c$ (i.e. $\{y^{1,c},  \cdots , y^{N_c,c} \}$) will have the value of distance function $\Gamma_{\mathcal{W}_{Y_c,\cdot,\cdot,\cdot}}$ smaller than the value corresponding to a point lying away from points $\{y^{1,c},  \cdots , y^{N_c,c} \}$. So the distance function $\Gamma_{\mathcal{W}_{Y_c,\cdot,\cdot,\cdot}}$ carries an information about the membership of a point to the set of points represented by $Y_c$. Similarly, the value $\min_{c \in \{1,\cdots,C \}} \;  \Gamma_{\mathcal{W}_{Y_c,\cdot,\cdot,\cdot}}(y)$ carries an information about the membership of $y$ to the training dataset $\mathbf{D}_{trn}$. To evaluate the potential of function $\min_{c \in \{1,\cdots,C \}} \;  \Gamma_{\mathcal{W}_{Y_c,\cdot,\cdot,\cdot}}$ in inferring the membership of a data point to training dataset $\mathbf{D}_{trn}$, a score, referred to as ``membership-inference score'', is defined. For this we define, for a given small positive number $\mathrm{o} \in \mathbb{R}_{\geq 0}$, sets $\mathbf{T}_{\mathrm{o}}, \mathbf{T}_{\mathrm{o}}^{\prime}  \subset \mathbb{R}_{\geq 0}$ as         
\begin{IEEEeqnarray}{rCl}
\mathbf{T}_{\mathrm{o}} & = & \{ y \in \mathbb{R}^p
 \; \mid \;  \min_{y^{\prime} \in \mathbf{D}_{trn}}\;\| y - y^{\prime} \| \leq \mathrm{o}  \},\; \mathrm{o} \in \mathbb{R}_{\geq 0}, \\
  \mathbf{T}_{\mathrm{o}}^{\prime} & = & \{ y \in \mathbb{R}^p
 \; \mid \;  \min_{y^{\prime} \in \mathbf{D}_{trn}}\;\| y - y^{\prime} \| > \mathrm{o}  \},\; \mathrm{o} \in \mathbb{R}_{\geq 0}.
 \end{IEEEeqnarray} 
Further define two non-negative functions, $\mathbf{r}_{\mathrm{o}}:\mathbf{T}_{\mathrm{o}} \rightarrow \mathbb{R}_{\geq 0}$ and $\mathbf{r}_{\mathrm{o}}^{\prime}:\mathbf{T}_{\mathrm{o}}^{\prime} \rightarrow \mathbb{R}_{\geq 0}$, as
\begin{IEEEeqnarray}{rCl}
\mathbf{r}_{\mathrm{o}}(y) & = & \min_{c \in \{1,2,\cdots,C \}} \;  \Gamma_{\mathcal{W}_{Y_c,n,L,S_c}}(y),\; y \in \mathbf{T}_{\mathrm{o}},\\
 \mathbf{r}_{\mathrm{o}}^{\prime}(y^{\prime}) & = & \min_{c \in \{1,2,\cdots,C \}} \;  \Gamma_{\mathcal{W}_{Y_c,n,L,S_c}}(y^{\prime}),\; y^{\prime} \in \mathbf{T}_{\mathrm{o}}^{\prime}.
 \end{IEEEeqnarray} 
Let $f_{\mathbf{r}_{\mathrm{o}}}$ and $f_{ \mathbf{r}_{\mathrm{o}}^{\prime}}$ denote the densities of probability distributions on $\mathbf{r}_{\mathrm{o}}$ and $ \mathbf{r}_{\mathrm{o}}^{\prime}$ respectively. It is obvious that $f_{\mathbf{r}_{\mathrm{o}}}$ characterizes the distribution of values taken by the function $\min_{c \in \{1,2,\cdots,C \}} \;  \Gamma_{\mathcal{W}_{Y_c,\cdot,\cdot,\cdot}}$ over data points lying within the distance of $\mathrm{o}$ from any data point included in the set $\mathbf{D}_{trn}$. Similarly, $f_{ \mathbf{r}_{\mathrm{o}}^{\prime}}$ characterizes the distribution of values taken by the function $\min_{c \in \{1,2,\cdots,C \}} \;  \Gamma_{\mathcal{W}_{Y_c,\cdot,\cdot,\cdot}}$ over data points lying away from the training dataset. Thus, the difference between $f_{\mathbf{r}_{\mathrm{o}}}$ and $f_{ \mathbf{r}_{\mathrm{o}}^{\prime}}$ is a measure of the potential of function $\min_{c \in \{1,\cdots,C \}} \;  \Gamma_{\mathcal{W}_{Y_c,\cdot,\cdot,\cdot}}$ in inferring the membership of a datapoint to training dataset $\mathbf{D}_{trn}$. Hence, the membership-inference score is defined as the $L2$ distance between $f_{\mathbf{r}_{\mathrm{o}}}$ and $f_{ \mathbf{r}_{\mathrm{o}}^{\prime}}$:
\begin{IEEEeqnarray}{rCl}
\label{eq_270220231128}mis & := & \int (f_{\mathbf{r}_{\mathrm{o}}}(r) - f_{ \mathbf{r}_{\mathrm{o}}^{\prime}}(r))^2 \dd r   .
 \end{IEEEeqnarray}   
Taking $\mathrm{o} = 0$, training dataset $\mathbf{D}_{trn}$ can be used to generate samples from $f_{\mathbf{r}_{0}}$, i.e., $\{ \mathbf{r}_{0}(y) \; \mid \; y \in \mathbf{D}_{trn} \}$ is the set of samples generated from $f_{\mathbf{r}_{0}}$. Similarly, test dataset (which is typically used to evaluate the classifier's performance), $\mathbf{D}_{tst}$, can be used to generate samples from $f_{\mathbf{r}_{0}}^{\prime}$, i.e., $\{ \mathbf{r}_{0}^{\prime}(y^{\prime}) \; \mid \; y^{\prime} \in \mathbf{D}_{tst} \}$ is the set of samples generated from $f_{\mathbf{r}_{0}}^{\prime}$. Now, the membership-inference score can be computed by approximating the $L2$ distance between $f_{\mathbf{r}_{0}}$ and $f_{ \mathbf{r}_{0}^{\prime}}$ from the samples $\{ \mathbf{r}_{0}(y) \; \mid \; y \in \mathbf{D}_{trn} \}$ and $\{ \mathbf{r}_{0}^{\prime}(y^{\prime}) \; \mid \; y^{\prime} \in \mathbf{D}_{tst} \}$ using a density-difference estimation method~\shortcite{10.1162/NECO_a_00492}.     
\section{Privacy-Preserving Learning}\label{sec_privacy}
Assuming that training dataset is private, KAHM based classification problem is considered under differential privacy framework. For this, an optimal differentially private noise adding mechanism is reviewed (in Section~\ref{sec_220320231639}) and a novel differentially private data fabrication method is developed for classification applications (in Section~\ref{sec_220320231757}). The application of KAHM to differentially private federated learning is considered (in Section~\ref{sec_federated_learning}). 
\subsection{An Optimal $(\epsilon,\delta)-$Differentially Private Noise Adding Mechanism}\label{sec_220320231639}
A given computational algorithm, operating on a data matrix $Y \in \mathbb{R}^{N \times p}$, can be represented by a mapping, $alg: \mathbb{R}^{N \times p} \rightarrow Range(alg)$. The privacy of data matrix $Y$ can be preserved via adding a suitable random noise to $Y$ before the application of algorithm $alg$ on the data matrix. This will result in a private version of algorithm $alg$ which is formally defined by Definition~\ref{definition_private_algorithm}.
\begin{definition}[A Private Algorithm on a Data Matrix]\label{definition_private_algorithm}
Given a computational algorithm $alg: \mathbb{R}^{N \times p} \rightarrow Range(alg)$, a private version of $alg$, $alg^+ : \mathbb{R}^{N \times p} \rightarrow Range(alg^+)$, is defined as
\begin{IEEEeqnarray}{rCl}
alg^+\left(Y\right) & := & alg\left(Y^+ \right),\\
Y^+ & = & Y + \mathrm{V},\;  \mathrm{V} \in \mathbb{R}^{N \times p},
\end{IEEEeqnarray}  
where $\mathrm{V}$ is a random noise matrix with $f_{\mathrm{v}_j^i}(v)$ being the probability density function of its $(i,j)-$th element $\mathrm{v}_j^i$; $\mathrm{v}_j^i$ and $\mathrm{v}_j^{i^{\prime}}$ are independent from each other for $i \neq i^{\prime}$; and $alg: \mathbb{R}^{N \times p} \rightarrow Range(alg)$ is a given mapping representing a computational algorithm. The range of $alg^+$ is as
 \begin{IEEEeqnarray}{rCl}
 Range(alg^+) & = & \left \{ alg\left(Y + \mathrm{V}\right)\; | \;  Y \in \mathbb{R}^{N \times p}, \mathrm{V} \in \mathbb{R}^{N \times p} \right \}. 
\end{IEEEeqnarray}  
\end{definition}  
We consider a threat scenario that an adversary seeks to gain an information about the data matrix $Y$ from an analysis of the change in output of algorithm $alg$ as a result of a change in data matrix. In particularly, we seek to attain differential privacy for algorithm $alg^+$ against the perturbation in an element of $Y$, say $(i_0,j_0)-$th element, such that magnitude of the perturbation is upper bounded by a scalar $d$.
\begin{definition}[$d-$Adjacency for Data Matrices]\label{def_adjacency_matrices}
Two matrices $Y,Y^{\prime} \in \mathbb{R}^{N \times p}$ are $d-$adjacent if for a given $d  \in \mathbb{R}_{+}$, there exist $i_0 \in \{1,2,\cdots,N\}$ and $j_0 \in \{1,2,\cdots,p\}$ such that $\forall i \in \{1,2,\cdots,N\},\; j \in \{1,2,\cdots,p\}$,
 \begin{IEEEeqnarray*}{rCl}
\left | (Y)_{i,j} - (Y^{\prime})_{i,j} \right | & \leq & \left\{ \begin{array}{cc}
d, & \mbox{if $i = i_0,j = j_0$} \\
0, & \mbox{otherwise}
  \end{array} \right.
\end{IEEEeqnarray*}    
where $(Y)_{i,j}$ and $(Y^{\prime})_{i,j} $ denote the $(i,j)-$th element of $Y$ and $Y^{\prime}$ respectively. Thus, $Y$ and $Y^{\prime}$ differ by only one element and the magnitude of the difference is upper bounded by $d$. 
\end{definition} 
\begin{definition}[$(\epsilon,\delta)-$Differential Privacy for $alg^+$~\cite{Kumar/IWCFS2019}]\label{def_differential_privacy}
The algorithm $alg^+\left(Y\right)$ is $(\epsilon,\delta)-$differentially private if
 \begin{IEEEeqnarray}{rCl}
\label{eq_differential_privacy}  Pr\{ alg^+\left(Y\right) \in \mathcal{O} \} & \leq & \exp(\epsilon) Pr\{ alg^+\left(Y^{\prime}\right)) \in \mathcal{O} \} + \delta
\end{IEEEeqnarray}     
for any measurable set $\mathcal{O} \subseteq  Range(alg^+) $ and for $d-$adjacent matrices pair $(Y,Y^{\prime})$.     
\end{definition} 
Definition~\ref{def_differential_privacy} implies that changing the value of an element in the matrix $Y$ by an amount upper bounded by $d$ can change the distribution of output of the algorithm $alg^+$ only by a factor of $\exp(\epsilon)$ with probability at least $1-\delta$. Thus, the lower value of $\epsilon$ and $\delta$ lead to a higher amount of privacy. 
\begin{result}[An Optimal $(\epsilon,\delta)-$Differentially Private Noise~\cite{Kumar/IWCFS2019}]\label{result_optimal_noise_epsilon_delta_privacy}
The probability density function of noise, that minimizes the expected noise magnitude together with satisfying the sufficient conditions for $(\epsilon,\delta)-$differential privacy for $alg^+$, is given as
\begin{IEEEeqnarray}{rCl}
\label{eq_optimal_density_epsilon_delta_privacy} f_{\mathrm{v}_j^i}^*(v;\epsilon,\delta,d) &  = & \left \{\begin{array}{cl}  {\delta}\: Dirac\delta(v), & v = 0 \\
 (1- \delta)\frac{\displaystyle \epsilon}{\displaystyle  2 d} \exp(-\frac{\displaystyle  \epsilon}{\displaystyle   d} |v|), & v \in   \mathbb{R} \setminus \{0\}
\end{array} \right.
\end{IEEEeqnarray}  
where $Dirac\delta(v)$ is Dirac delta function satisfying $\int_{-\infty}^{\infty}Dirac\delta(v)\: \dd v = 1$. 
\end{result}
\begin{remark}[Generating Random Samples from $f_{\mathrm{v}_j^i}^*$] \label{remark_sampling}
The method of {\it inverse transform sampling} can be used to generate random samples from cumulative distribution function. The cumulative distribution function of $  f_{\mathrm{v}_j^i}^*$ is given as
 \begin{IEEEeqnarray}{rCl}
F_{\mathrm{v}_j^i}(v;\epsilon,\delta,d) & = & \left \{\begin{array}{ll} \frac{ 1-\delta}{  2} \exp(\frac{\epsilon }{d} v), & v < 0 \\
\frac{ 1 + \delta}{  2}, &  v = 0  \\
1 - \frac{ 1 - \delta}{  2}\exp(-\frac{\epsilon }{d} v), & v > 0
\end{array} \right. 
\end{IEEEeqnarray}      
The inverse cumulative distribution function is given as
 \begin{IEEEeqnarray}{rCl}
\label{eq_inverse_cdf}F_{\mathrm{v}_j^i}^{-1}(t^i_j;\epsilon,\delta,d) & = & \left \{\begin{array}{ll} \frac{d}{\epsilon} \log(\frac{2t^i_j}{1 - \delta}), & t^i_j <  \frac{1- \delta}{2} \\
0, &  t^i_j \in [ \frac{1- \delta}{2}, \frac{1+ \delta}{2}]  \\
 -\frac{d}{\epsilon} \log(\frac{2(1-t^i_j)}{1-\delta}), & t^i_j > \frac{1+\delta}{2}
\end{array} \right.,\; t^i_j \in (0,1).
\end{IEEEeqnarray}   
Thus, via generating random samples from the uniform distribution on $(0,1)$ and using~(\ref{eq_inverse_cdf}), the noise additive mechanism can be implemented.   
\end{remark} 
\begin{algorithm}
\caption{Differentially private approximation of a data matrix~\cite{kumar2023differentially}}
\label{algorithm_differential_private_approximation}
\begin{algorithmic}[1]
\REQUIRE  Data matrix $Y \in \mathbb{R}^{N \times p}$; differential privacy parameters: $d  \in \mathbb{R}_{+}$,  $\epsilon  \in \mathbb{R}_{+}$, $\delta \in (0,1)$.
\STATE Compute $\forall \; i \in \{1,2,\cdots,N\},\; j \in \{1,2,\cdots,p\}$, 
\begin{IEEEeqnarray}{rCl}
(Y^+_{\epsilon})_{i,j}  & = & (Y)_{i,j}+ F_{\mathrm{v}_j^{i}}^{-1}(t^{i}_j;\epsilon,\delta,d),\; t^{i}_j \in (0,1),
\end{IEEEeqnarray}     
where $t^{i}_j $ is chosen from the uniform distribution on $(0,1)$ and $ F_{\mathrm{v}_j^{i}}^{-1}$ is given by (\ref{eq_inverse_cdf}).    
 \RETURN $Y^+_{\epsilon}$ (where the subscript $\epsilon$ indicates the given privacy-loss bound $\epsilon$). 
\end{algorithmic}
\end{algorithm} 
For a given value of $(\epsilon,\delta,d)$, Algorithm~\ref{algorithm_differential_private_approximation} is stated for the differentially private approximation of a data matrix.   
\subsection{Differentially Private Data Fabrication and Classification}\label{sec_220320231757}
A computational algorithm can be made to ensure differential privacy (i.e. inequality (\ref{eq_differential_privacy})) via applying the algorithm on the data matrix returned by Algorithm~\ref{algorithm_differential_private_approximation}. Hence, a KAHM based differentially private classifier can be built as in Definition~\ref{def_diff_priv_KAHM_classifier_ref}.  
\begin{definition}[A KAHM Based Differentially Private Classifier]\label{def_diff_priv_KAHM_classifier_ref}
Given a multi-class labelled differentially private dataset $\{ (Y_{\epsilon, i}^+,\mathrm{cl}^i ) \; \mid \; Y_{\epsilon,i}^+ \in \mathbb{R}^{N_i \times p},\;\mathrm{cl}^i \in \{1,2,\cdots,C \} \}$, a KAHM based differentially private classifier $\mathcal{C}:\mathbb{R}^p \rightarrow \{1,2,\cdots,C \}$ is defined as
\begin{IEEEeqnarray}{rCl}
\label{eq_130220231237}\mathcal{C}(y;\mathcal{W}_{Y_{\epsilon,1}^+,n,L,S_1},\cdots,\mathcal{W}_{Y_{\epsilon,C}^+,n,L,S_C}) & = &  \arg \; \min_{c \in \{1,2,\cdots,C \}} \;  \Gamma_{\mathcal{W}_{Y_{\epsilon,c}^+,n,L,S_c}}(y),
  \end{IEEEeqnarray}   
where $\mathcal{W}_{Y_{\epsilon,c}^+,\cdot,\cdot,\cdot}$ is the wide conditionally deep KAHM (Definition~\ref{def_WDKAHM}) modeling the $c-$th class labelled data points and $\Gamma_{\mathcal{W}_{Y_{\epsilon,c}^+,\cdot,\cdot,\cdot}}(\cdot)$ is the distance function (Definition~\ref{def_distance_function_wide_cond_deep_KAHM}) induced by $\mathcal{W}_{Y_{\epsilon,c}^+,\cdot,\cdot,\cdot}$, and $Y_{\epsilon,c}^+$ is differentially private data matrix obtained by Algorithm~\ref{algorithm_differential_private_approximation}. The classifier assigns to an arbitrary point $y \in \mathbb{R}^p$ the label of the class which has the minimum distance between $y$ and $y$'s image onto the affine hull of differentially private samples of that class. 
\end{definition}
Since a differentially private algorithm operates on noise added data, the algorithm's performance is adversely affected. An obvious effect of adding noise to data matrix is an increase in the modeling error of data samples by a KAHM. Typically, we have
\begin{IEEEeqnarray}{rCl}
\sum_{i=1}^N\| y^{+i}_{\epsilon} -  \mathcal{A}_{Y^+_{\epsilon},n}\left(y^{+i}_{\epsilon}\right)\| & > & \sum_{i=1}^N\| y^i -  \mathcal{A}_{Y,n}\left(y^i\right)  \|,
 \end{IEEEeqnarray}  
where $y^{+i}_{\epsilon} = ((Y^+_{\epsilon})_{i,:})^T$. Thus, an approach to alleviate the effect of added noise on the performance of a KAHM based algorithm is of processing the noise added data matrix through a data smoother such that the smoothed data matrix leads to a KAHM with modeling error not larger than the modeling error on original data samples. One such smoother is defined as in Definition~\ref{def_transformation}.   
\begin{definition}[A Smoother for Differentially Private Data]\label{def_transformation}
Given a differentially private matrix $Y^+_{\epsilon}  \in \mathbb{R}^{N \times p}$, a subspace dimension $n \leq p$, and a positive integer $M \in \mathbb{Z}_+$; $Y^+_{\epsilon}$ is transformed into $\hat{Y}_{M-1} \in \mathbb{R}^{N \times p}$ through following recursions run from $m=0$ to $m = M-1$:   
\begin{IEEEeqnarray}{rCl}
\label{eq_010120231251}\hat{y}^{i,0} & = & ((Y^+_{\epsilon})_{i,:})^T,\; \forall i \in \{1,2,\cdots,N \},\\
\label{eq_010120231252}\hat{y}^{i,m+1} & = &  \left(\sum_{j=1}^Nh_{k_{\theta_m},\hat{Y}_mP^T_m,\lambda^*_m}^j(P_m\hat{y}^{i,m}) \right) \times \mathcal{A}_{\hat{Y}_{m},n}\left(\hat{y}^{i,m} \right),\\
\label{eq_010120231253}  \hat{Y}_m & = & \left[\begin{IEEEeqnarraybox*}[][c]{,c/c/c,}  \hat{y}^{1,m}  & \cdots & \hat{y}^{N,m}  \end{IEEEeqnarraybox*} \right]^T,
     \end{IEEEeqnarray}   
where
\begin{itemize}
\item $P_m$ is defined by setting the $i-$th row of $P_m$ as equal to transpose of eigenvector corresponding to $i-$th largest eigenvalue of sample covariance matrix of the dataset $\{\hat{y}^{1,m},\cdots,\hat{y}^{N,m} \}$.
\item $\theta_m$ is sample covariance matrix of dataset $\{P_m\hat{y}^{1,m},\cdots,P_m\hat{y}^{N,m} \}$, i.e., 
   \begin{IEEEeqnarray}{rCl}
   \theta_m &  = & \frac{1}{N-1}_mP_m\left(\hat{Y}_m - \mathbf{1}_N \frac{\sum_{i=1}^N(\hat{y}^{i,m})^T}{N}\right)^T\left(\hat{Y}_m - \mathbf{1}_N \frac{\sum_{i=1}^N(\hat{y}^{i,m})^T}{N}\right)P_m^T.
  \end{IEEEeqnarray}
  \item $\lambda^*_m \in \mathbb{R}_+$ is given as
  \begin{IEEEeqnarray}{rCl}
\lambda^*_m & = &  \hat{e}_m + \frac{2}{pN}\|\hat{Y}_m \|^2_F, 
     \end{IEEEeqnarray}   
where $\hat{e}_m$ is the unique fixed point of the function $\mathcal{R}_{k_{\theta_m},\hat{Y}_mP_m^T,\hat{Y}_m}$ (which is defined as in (\ref{eq_090120230832})).
\item $k_{\theta_m}(\cdot,\cdot)$ and $h_{k_{\theta_m},\hat{Y}_mP^T_m,\lambda^*_m}^i(\cdot)$ are defined as in (\ref{eq_260120231329}) and (\ref{eq_250220231734}) respectively.  
\end{itemize}
The transformation of $Y^+_{\epsilon}$ into $\hat{Y}_{M-1} $ is represented as
  \begin{IEEEeqnarray}{rCl}
\hat{Y}_{M-1} & = &   \mathrm{T}_{n,M}(Y^+_{\epsilon}).
     \end{IEEEeqnarray}         
\end{definition}  
The transformation of $Y^+_{\epsilon}$ into $\hat{Y}_{M-1}$ has been defined in a particular way to ensure a property related to the error in KAHM modeling of data samples. This property is sated in Theorem~\ref{result_transformation}.         
\begin{theorem}\label{result_transformation}
The error in KAHM modeling of smoothed data matrix $\hat{Y}_{M-1}  =    \mathrm{T}_{n,M}(Y^+_{\epsilon})$ converges asymptotically with an increasing value of $M$ to zero, i.e.,
\begin{IEEEeqnarray}{rCl}
 \lim_{M \to \infty} \sum_{i=1}^N \| \hat{y}^{i,M-1} - \mathcal{A}_{\hat{Y}_{M-1},n}(\hat{y}^{i,M-1}) \|  & = & 0.
 \end{IEEEeqnarray}    
where $\hat{Y}_{M-1} = \left[\begin{IEEEeqnarraybox*}[][c]{,c/c/c,}  \hat{y}^{1,M-1}  & \cdots & \hat{y}^{N,M-1}  \end{IEEEeqnarraybox*} \right]^T$ is computed using recursions (\ref{eq_010120231251}-\ref{eq_010120231253}) from $m=0$ to $m = M-1$.
\begin{proof} 
The proof is provided in Appendix~\ref{appendix5}.
\end{proof}
\end{theorem}     
It follows from Theorem~\ref{result_transformation} that the KAHM modeling error of smoothed data samples can be reduced to an arbitrary low value by choosing a sufficiently large value of $M$. Thus, it is possible to find the smallest number, say $\tilde{M} \in \mathbb{Z}_+$, ensuring that   
\begin{IEEEeqnarray}{rCl}
\sum_{i=1}^N \| \hat{y}^{i,\tilde{M}-1} - \mathcal{A}_{\hat{Y}_{\tilde{M}-1},n}(\hat{y}^{i,\tilde{M}-1}) \| & \leq & r,
 \end{IEEEeqnarray} 
 where $r$ is the error in KAHM modeling of original data matrix $Y$ defined as  
 \begin{IEEEeqnarray}{rCl}
 r & = & \sum_{i=1}^N \|y^i - \mathcal{A}_{Y,n}(y^i) \|.
 \end{IEEEeqnarray}       
That is, error in KAHM modeling of smoothed data matrix $\hat{Y}_{\tilde{M}-1}$ is lower than the error in KAHM modeling of original data matrix $Y$, which suggests that applying a KAHM based computational algorithm on $\hat{Y}_{\tilde{M}-1}$ (instead of $Y^+_{\epsilon}$) may alleviate the effect of added noise on the performance of a KAHM based computational algorithm. This motivates to use the KAHM associated to $\hat{Y}_{\tilde{M}-1}$, i.e. $\mathcal{A}_{\hat{Y}_{\tilde{M}-1},n}$, for fabricating data samples meant for building KAHM based models.      
\begin{definition}[Differentially Private Fabricated Data]\label{def_fabricated_data}
Given a differentially private matrix $Y^+_{\epsilon}  \in \mathbb{R}^{N \times p}$ ensuring the privacy-loss bound $\epsilon \in \mathbb{R}_+$, a subspace dimension $n \leq p$, and error in KAHM modeling of original data matrix $Y$ evaluated as $r = \sum_{i=1}^N  \|y^i - \mathcal{A}_{Y,n}(y^i) \|$; a differentially private fabricated data matrix $\tilde{Y} \in \mathbb{R}^{N \times p}$ is defined as
\begin{IEEEeqnarray}{rCl}
\tilde{Y} &  = & \left[\begin{IEEEeqnarraybox*}[][c]{,c/c/c,} \mathcal{A}_{\hat{Y}_{\tilde{M}-1},n}( \hat{y}^{1,\tilde{M}-1})  & \cdots & \mathcal{A}_{\hat{Y}_{\tilde{M}-1},n}( \hat{y}^{N,\tilde{M}-1})   \end{IEEEeqnarraybox*} \right]^T,\\
\hat{y}^{i,\tilde{M}-1} & = & ( (\hat{Y}_{\tilde{M}-1})_{i,:})^T,\\
\hat{Y}_{\tilde{M}-1} & = & \mathrm{T}_{n,\tilde{M}}(Y^+_{\epsilon}),
 \end{IEEEeqnarray}
where smoother $\mathrm{T}_{n,\tilde{M}}$ (Definition~\ref{def_transformation}) computes $ \hat{Y}_{\tilde{M}-1}$ through recursions (\ref{eq_010120231251}-\ref{eq_010120231253}) from $m= 0$ to $m = \tilde{M}-1$, and $\tilde{M}$ is defined as 
\begin{IEEEeqnarray}{rCl}
\tilde{M} &=& \min\left\{ m \in \mathbb{Z}_{+}\; \mid \;  \sum_{i=1}^N \| \hat{y}^{i,m-1} - \mathcal{A}_{\hat{Y}_{m-1},n}(\hat{y}^{i,m-1}) \| \leq  r   \right\}. \IEEEeqnarraynumspace
 \end{IEEEeqnarray}  
 The computing of $\tilde{Y}$ is represented as
\begin{IEEEeqnarray}{rCl}
\tilde{Y} & = & \mathcal{F}_{n}(Y^+_{\epsilon};r).
 \end{IEEEeqnarray}  
The fabricated data matrix $\tilde{Y}$ is computed from $Y^+_{\epsilon}$ (which is a differentially private approximation of $Y$) and not from original data matrix $Y$, and thus $\tilde{Y}$ remains differentially private.  
\end{definition}
\begin{figure}
\centering
\includegraphics[width=0.5\textwidth]{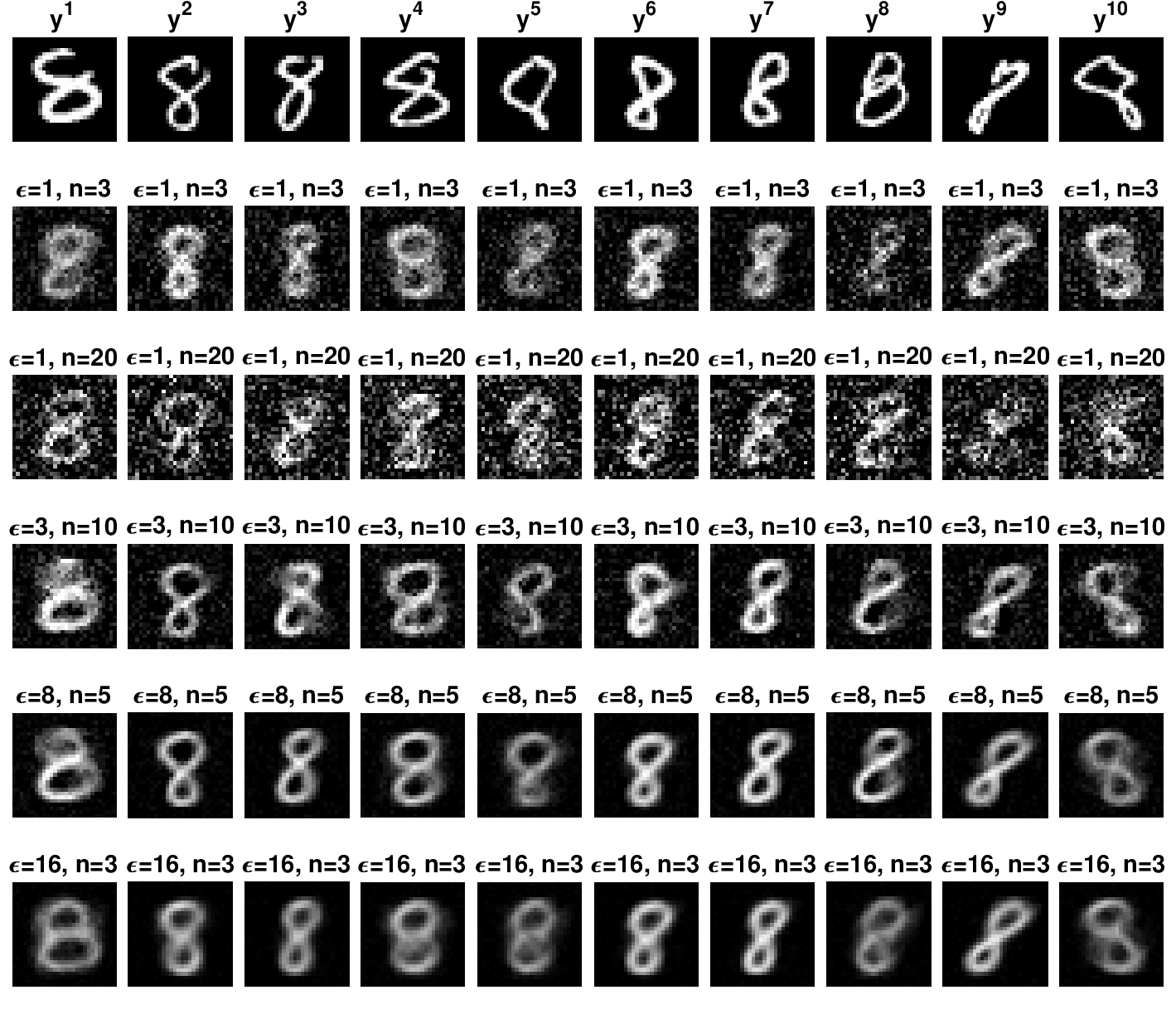}
\caption{A dataset $Y$ consisting of 1000 randomly chosen samples of digit 8 from MNIST dataset was considered. For 10 randomly selected samples from $Y$ (displayed at top row of the figure), the corresponding samples from differentially private fabricated data $\tilde{Y} = \mathcal{F}_{n}(Y^+_{\epsilon};\sum_{i=1}^{1000} \|y^i - \mathcal{A}_{Y,n}(y^i) \|)$ have been displayed for different values of privacy-loss bound $\epsilon$ and subspace dimension $n$.}
\label{fig_demo_dp_fabricated_data}
\end{figure} 
Fig.~\ref{fig_demo_dp_fabricated_data} displays a few examples of fabricated data samples corresponding to different choices for privacy-loss bound $\epsilon$ and subspace dimension $n$. As expected and also observed in Fig.~\ref{fig_demo_dp_fabricated_data}, more and more features of original data samples get masked in the fabricated data with a decrease in $\epsilon$ and/or $n$.   
\begin{remark}[Big Data Fabrication]\label{rem_big_data_fabrication}
For the big datasets with large $N$, the data can be divided into subsets via e.g. k-means clustering and fabricated data matrix is computed from each subset independently. That is, $\tilde{Y}$ is fabricated as follows:   
\begin{IEEEeqnarray}{rCl}
\tilde{Y} & = &  \left[\begin{IEEEeqnarraybox*}[][c]{,c/c/c,} \left(\mathcal{F}_{n}(Y_1^+;r_1)\right)^T  & \cdots & \left(\mathcal{F}_{n}(Y_S^+;r_S)\right)^T   \end{IEEEeqnarraybox*} \right]^T,\\
Y_s^+ & \leftarrow & Algorithm~\ref{algorithm_differential_private_approximation}(Y_s,d,\epsilon,\delta),\\
r_s & = & \sum_{i=1}^{N_s}  \|y^{i,s} - \mathcal{A}_{Y_s,n}(y^{i,s}) \|,\\
Y_s & = & \left[\begin{IEEEeqnarraybox*}[][c]{,c/c/c,} y^{1,1} & \cdots & y^{N_s,s} \end{IEEEeqnarraybox*} \right]^T,\\
\left\{y^{1,s},\cdots,y^{N_s,s}\right\}_{s=1}^S & = & \mathrm{clustering}(\{y^1,\cdots,y^N\}, S),\\
S & = & \lceil N/1000 \rceil,
  \end{IEEEeqnarray}  
where $ \mathrm{clustering}(\{y^1,\cdots,y^N\}, S)$ represents $k-$means clustering into $S$ subsets
\end{remark}
As the fabricated data remain differentially private, a KAHM based classifier can be built using fabricated data to ensure differential privacy in the sense of inequality~(\ref{eq_differential_privacy}).
\begin{definition}[A Differentially Private Classifier Based on Fabricated Data]\label{def_diff_priv_KAHM_classifier}
Given a multi-class labelled differentially private fabricated dataset $\{ (\tilde{Y}_i,\mathrm{cl}^i ) \; \mid \; \tilde{Y}_i \in \mathbb{R}^{N_i \times p},\;\mathrm{cl}^i \in \{1,2,\cdots,C \} \}$, a classifier $\mathcal{C}:\mathbb{R}^p \rightarrow \{1,2,\cdots,C \}$ is defined as
\begin{IEEEeqnarray}{rCl}
\label{eq_030220231011}\mathcal{C}(y;\mathcal{W}_{\tilde{Y}_1,n,L,S_1},\cdots,\mathcal{W}_{\tilde{Y}_C,n,L,S_C}) & = &  \arg \; \min_{c \in \{1,2,\cdots,C \}} \;  \Gamma_{\mathcal{W}_{\tilde{Y}_c,n,L,S_c}}(y),
  \end{IEEEeqnarray}   
where $\mathcal{W}_{\tilde{Y}_c,\cdot,\cdot,\cdot}$ is the wide conditionally deep KAHM (Definition~\ref{def_WDKAHM}) modeling the $c-$th class labelled data points and $\Gamma_{\mathcal{W}_{\tilde{Y}_c,\cdot,\cdot,\cdot}}(\cdot)$ is the distance function (Definition~\ref{def_distance_function_wide_cond_deep_KAHM}) induced by $\mathcal{W}_{\tilde{Y}_c,\cdot,\cdot,\cdot}$, and $\tilde{Y}_c$ is differentially private fabricated data matrix (Definition~\ref{def_fabricated_data}). The classifier assigns to an arbitrary point $y \in \mathbb{R}^p$ the label of the class which has the minimum distance between $y$ and $y$'s image onto the affine hull of differentially private fabricated samples of that class. 
\end{definition}
\subsection{Application to Differentially Private Federated Learning}\label{sec_federated_learning}
\begin{figure}
\centerline{\subfigure[images of three different KAHMs built independently using three different datasets]{\includegraphics[width=0.4\textwidth]{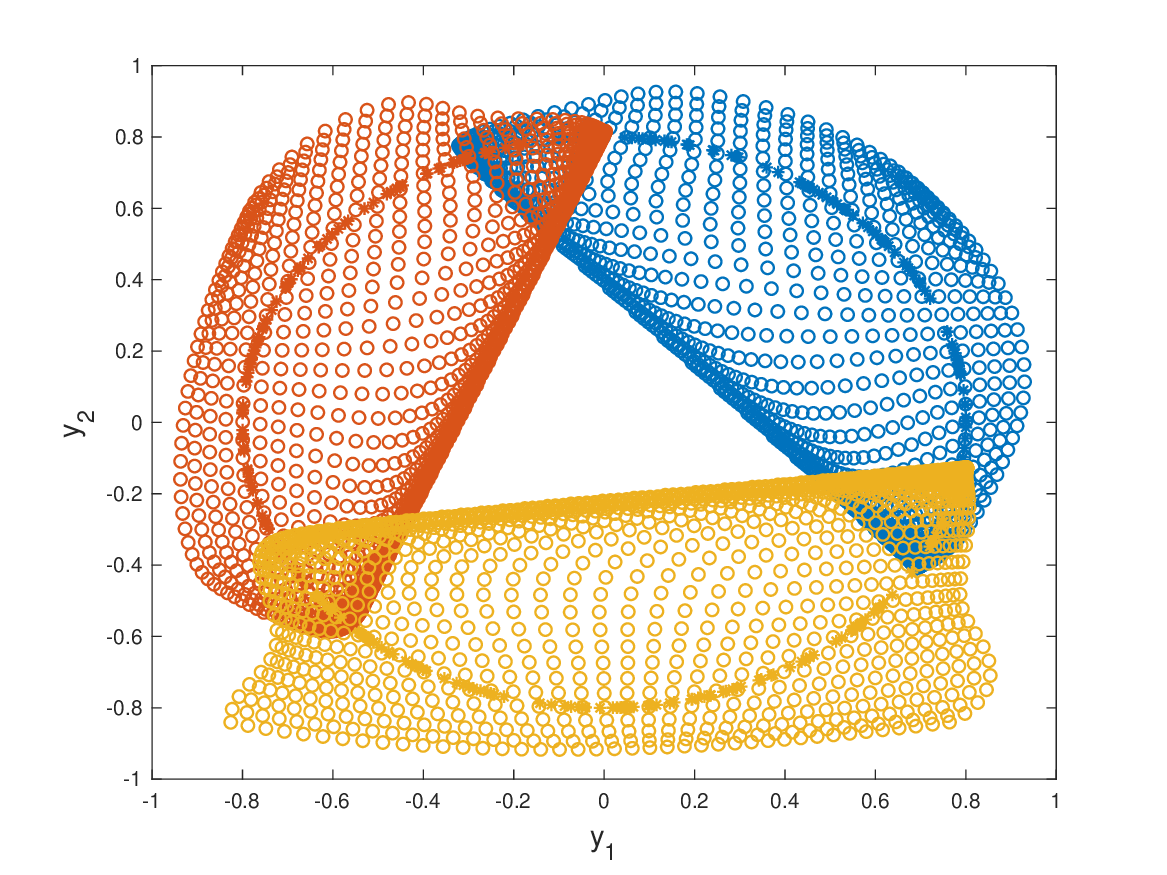}\label{fig_demo_wide_1}} \hfil 
\subfigure[image of the global KAHM combining together independently built KAHMs]{\includegraphics[width=0.4\textwidth]{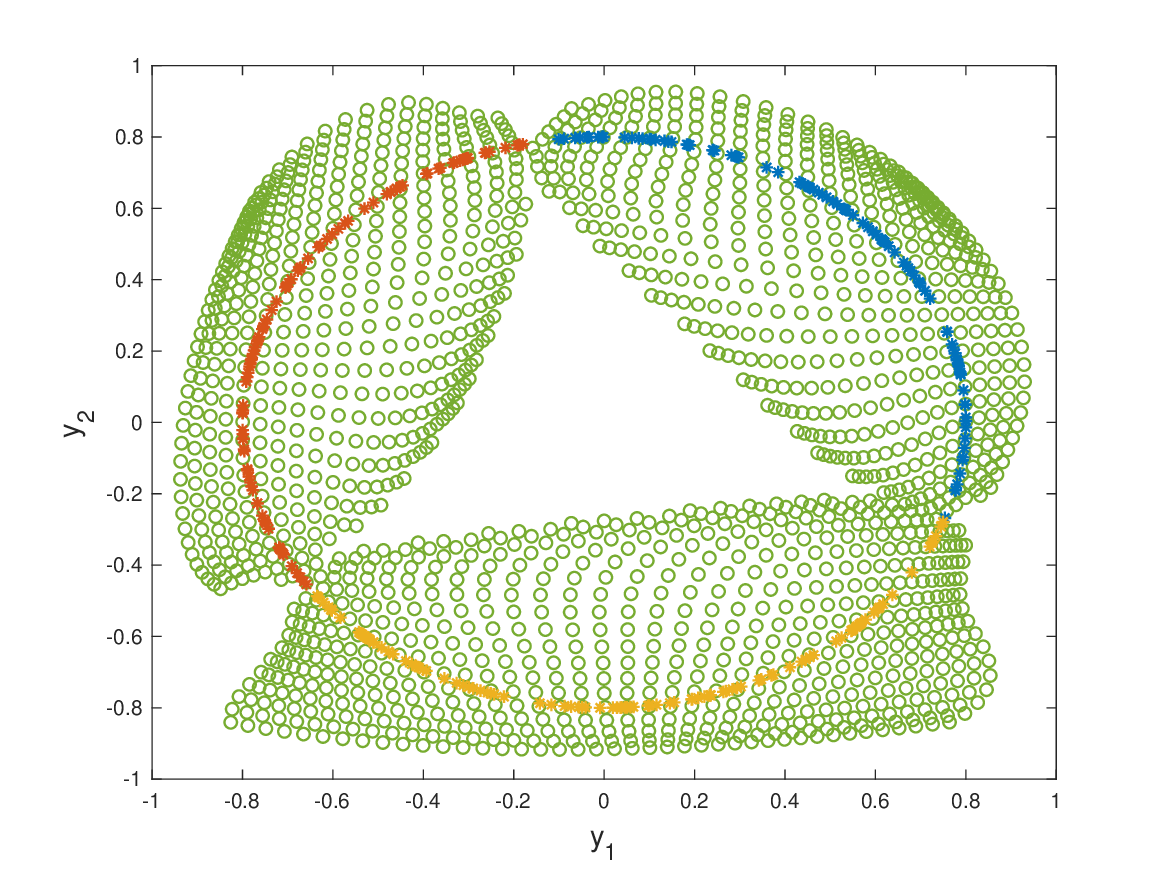}\label{fig_demo_wide_2}}}
\caption{An example of combining together local KAHMs to build a global KAHM.}
\label{fig_demo_combination}
\end{figure}
\begin{figure}  
\centering
\scalebox{0.85}{\begin{tikzpicture}[scale=0.9]
\path[fill=green!10](-3,-8.25)--(3,-8.25)--(3,1.5)--(-3,1.5)--cycle;
\draw[green,line width = 0.25mm](-3,-8.25)--(3,-8.25)--(3,1.5)--(-3,1.5)--cycle;
\draw (0,-7.75) node[]{\bfseries $\begin{array}{c} \mbox{\scriptsize Party 1} \\ \mbox{\scriptsize \faMale\:\faFemale} \end{array}$};
\draw (0,-6.25) node[rounded corners,draw=red,fill=red!5](n1){\footnotesize $\begin{array}{c}\mbox{local training data} \\ \mbox{to be protected} \\ \mbox{\small $\{ Y^1_1,\cdots, Y^1_C \}$} \end{array}$};
\draw (0,-4) node[rounded corners,draw](nadd1){\footnotesize $\begin{array}{c}\mbox{differentially private} \\ \mbox{fabricated data} \\ \mbox{\small $\{ \tilde{Y}^1_1,\cdots, \tilde{Y}^1_C \}$}\end{array}$};
\draw[-latex,line width=0.2mm] (n1) to [out=90,in=-90] (nadd1);  
\draw (0,-1.9) node[rounded corners,draw](n2){ \footnotesize $\begin{array}{c}\mbox{local models} \\ \mbox{\small  $ \{ \mathcal{W}_{\tilde{Y}_1^1,n,L,S_1^1},\cdots, \mathcal{W}_{\tilde{Y}_C^1,n,L,S_C^1} \}$} \end{array}$};
\draw[-latex,line width=0.2mm] (nadd1) to [out=90,in=-90] (n2);  
\draw (0,0.05) node[rounded corners,draw](n4){ \footnotesize $\begin{array}{c}\mbox{inference} \\ \mbox{\small $\{  \Gamma_{\mathcal{W}_{\tilde{Y}_c^1,n,L,S_c^1}}(y) \}_{c=1}^C$} \end{array}$};
\draw[-latex,line width=0.2mm] (n2) to [out=90,in=-90] (n4);   
\path[fill=yellow!10](9,-8.25)--(15,-8.25)--(15,1.5)--(9,1.5)--cycle;
\draw[yellow,line width = 0.25mm](9,-8.25)--(15,-8.25)--(15,1.5)--(9,1.5)--cycle;
\draw (11.75,-7.75) node[]{\bfseries $\begin{array}{c} \mbox{\scriptsize Party Q} \\ \mbox{\scriptsize \faMale\:\faFemale} \end{array}$};
\draw (12,0.05) node[rounded corners,draw](n11){ \footnotesize $\begin{array}{c}\mbox{inference} \\ \mbox{\small $\{  \Gamma_{\mathcal{W}_{\tilde{Y}_c^Q,n,L,S_c^Q}}(y) \}_{c=1}^C$} \end{array}$};
 \draw (12,-1.9) node[rounded corners,draw](n14){ \footnotesize $\begin{array}{c}\mbox{local models} \\ \mbox{\small $ \{ \mathcal{W}_{\tilde{Y}_1^Q,n,L,S_1^Q},\cdots, \mathcal{W}_{\tilde{Y}_C^Q,n,L,S_C^Q} \}$} \end{array}$};
  \draw[-latex,line width=0.2mm] (n14) to [out=90,in=-90] (n11); 
\draw (12,-4) node[rounded corners,draw](nadd2){\footnotesize $\begin{array}{c}\mbox{differentially private} \\ \mbox{fabricated data} \\ \mbox{\small $\{ \tilde{Y}^Q_1,\cdots, \tilde{Y}^Q_C \}$}\end{array}$};
\draw[-latex,line width=0.2mm] (nadd2) to [out=90,in=-90] (n14);  
\draw (12,-6.25) node[rounded corners,draw=red,fill= red!5](n15){\footnotesize $\begin{array}{c}\mbox{local training data} \\ \mbox{to be protected} \\ \mbox{\small $\{ Y^Q_1,\cdots, Y^Q_C \}$} \end{array}$};
\draw[-latex,line width=0.2mm] (n15) to [out=90,in=-90] (nadd2);  
\path[fill=blue!10](4,-8.25)--(8,-8.25)--(8,1.5)--(4,1.5)--cycle;   
\draw[blue!40,line width = 0.25mm](4,-8.25)--(8,-8.25)--(8,1.5)--(4,1.5)--cycle; 
\draw (6,-7.75) node[]{\bfseries $\begin{array}{c} \mbox{\scriptsize User} \\ \mbox{\scriptsize \faUser} \end{array}$};
  \draw (6,-3.5) node[](n20){ \footnotesize $\begin{array}{c}\mbox{input} \\ \mbox{\small $y$} \end{array}$};
\draw[thick,line width=0.2mm](n20) -- (6,-2.5);
\draw[thick,line width=0.2mm](3.5,-2.5) -- (8.5,-2.5);
\draw[-latex,thick,line width=0.2mm] (3.5,-2.5) to [out=90,in=0] (n4);   
 \draw[-latex,thick,line width=0.2mm] (8.5,-2.5) to [out=90,in=180] (n11); 
  \node[cloud,
    draw = gray,
    fill = cyan!5,
    minimum width = 11cm,
    minimum height = 3.5cm,
    cloud puffs = 18] (c) at (6,4) {};
\draw (6,3.85) node[rounded corners,draw, fill=gray!5](n8){ \footnotesize $\begin{array}{c} \mbox{\small $ \arg \; \min_{c \in \{1,2,\cdots,C \}}\; \left( \min_{q \in \{1,2,\cdots,Q \}}\;\Gamma_{ \mathcal{W}_{\tilde{Y}^q_c,n,L,S^q_c}}(y) \right)$}  \end{array}$};
\draw[-latex,line width=0.2mm,cyan] (n4) to [out=90,in=180] (n8);  
\draw (6,4.8) node[]{\bfseries {\scriptsize $\begin{array}{c} \mbox{global aggregator} \end{array}$  }};
\draw[-latex,line width=0.2mm,cyan] (n11) to [out=90,in=0] (n8);  
\draw (6,0.5) node[rounded corners,draw, fill = gray!5](n17){ \footnotesize $\begin{array}{c}\mbox{output} \\ \mbox{\small $\mathcal{GC}(y) $} \end{array}$};
\draw[-latex,line width=0.2mm,cyan] (n8) to [out=-90,in=90] (n17);  
\end{tikzpicture}}
\caption{The structural representation of the KAHM based federated learning scheme. The limitation of passing users' inputs to the parties can be addressed as suggested in Remark~\ref{rem_input_passing_to_clients}.}
\label{fig_KAHM_distributed_classifier}
\end{figure}
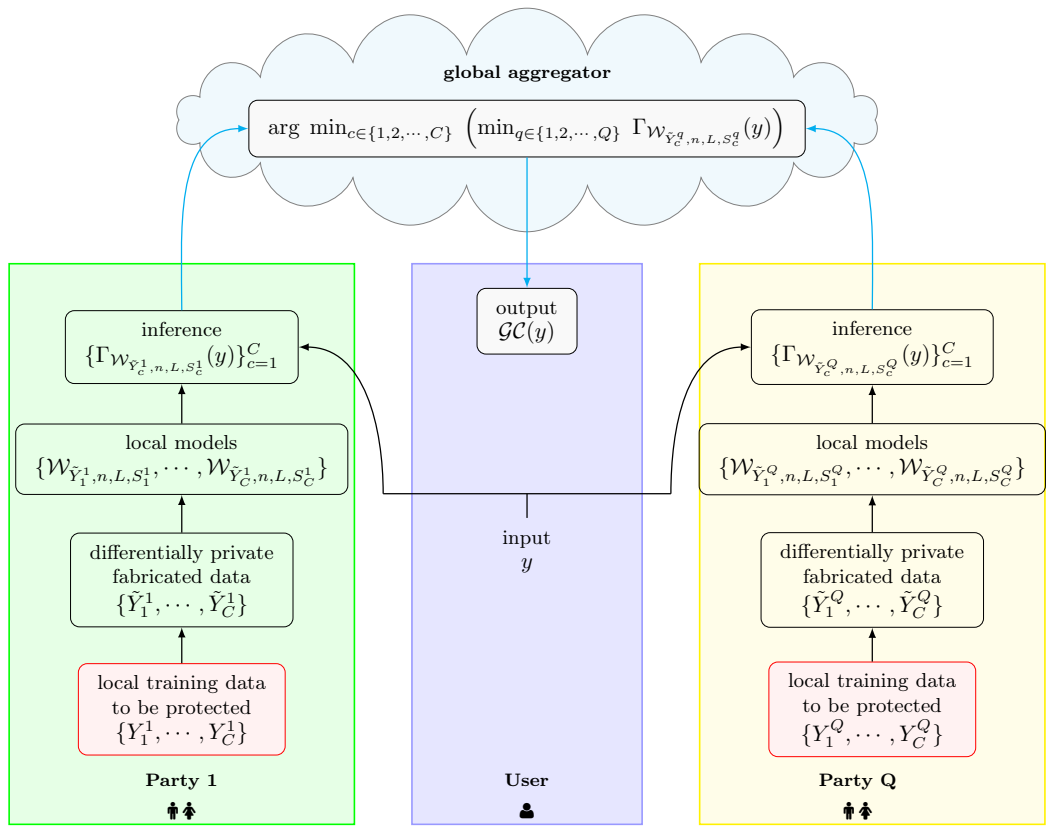
The multi-class classification problem is considered under the scenario of data being distributed amongst different parties. For the case of data being privately owned by local parties, our federated learning approach is of combining together the local privacy-preserving KAHM based classifiers using the distance functions induced by local KAHMs. For this, a combination of different KAHMs is considered using the distance measure. Given $Q$ different wide conditionally deep KAHMs $\mathcal{W}_{Y^1,n,L,S^1},\cdots,\mathcal{W}_{Y^Q,n,L,S^Q}$ built independently using datasets $Y^1,\cdots,Y^Q$ respectively, a possible way to combine together the KAHMs is as follows:            
\begin{IEEEeqnarray}{rCl}
\label{eq_290520231906}\mathcal{GW}(y;\{ \mathcal{W}_{Y^q,n,L,S^q}\}_{q=1}^Q) & = & \mathcal{W}_{Y^{\hat{q}(y)},n,L,S^{\hat{q}(y)}}(y),\\
\label{eq_290520231907}\hat{q}(y) & = & \arg \; \min_{q \in \{1,2,\cdots,Q \}}\;\Gamma_{ \mathcal{W}_{Y^q,n,L,S^q}}(y),
  \end{IEEEeqnarray}   
where $\mathcal{GW}$ is the global KAHM (that combines together the individual KAHMs) and $\Gamma_{\mathcal{W}_{Y^q,n,L,S^q}}$ is the distance function (Definition~\ref{def_distance_function_wide_cond_deep_KAHM}) induced by $\mathcal{W}_{Y^q,n,L,S^q}$. For an input $y \in \mathbb{R}^p$, the global KAHM output is equal to the output of $\hat{q}-$th KAHM (which is the KAHM resulting in minimum Euclidean distance between input vector $y$ and its image onto the affine hull of data samples). A 2-dimensional data example where three different KAHMs are combined to build a global KAHM is provided in Fig.~\ref{fig_demo_combination}. Fig.~\ref{fig_demo_combination} shows the images of individual KAHMs (in Fig.~\ref{fig_demo_wide_1}) and the image of global KAHM (in Fig.~\ref{fig_demo_wide_2}).       

The local KAHMs modeling a specific class can be combined together to build a global KAHM (that models the region (in data space) of that class) and a global classifier can be built from all class-specific global KAHMs. Mathematically, the global classifier, $\mathcal{GC}:\mathbb{R}^p \rightarrow \{1,2,\cdots,C \}$, is defined as
\begin{IEEEeqnarray}{rCl}
\label{eq_300320231210}\mathcal{GC}(y) & = &  \arg \; \min_{c \in \{1,2,\cdots,C \}}\; \|y - \mathcal{GW}(y;\{ \mathcal{W}_{\tilde{Y}^q_c,n,L,S^q_c}\}_{q=1}^Q)  \|,
  \end{IEEEeqnarray}            
where $\tilde{Y}_c^q$ represents the $c-$th class labelled differentially private data samples fabricated locally by the $q-$th party and $\mathcal{GW}$ is the global KAHM (\ref{eq_290520231906}). The global classifier (\ref{eq_300320231210}) assigns to an arbitrary point $y \in \mathbb{R}^p$ the label of the class which has the minimum distance between $y$ and $y$'s image onto the affine hull of differentially private fabricated samples of that class. (\ref{eq_300320231210}) can be alternatively expressed as
\begin{IEEEeqnarray}{rCl}
\label{eq_300320231343}\mathcal{GC}(y) & = &  \arg \; \min_{c \in \{1,2,\cdots,C \}}\; \left( \min_{q \in \{1,2,\cdots,Q \}}\;\Gamma_{ \mathcal{W}_{\tilde{Y}^q_c,n,L,S^q_c}}(y) \right).
  \end{IEEEeqnarray} 
An important feature of the global classifier evaluation using (\ref{eq_300320231343}) is that the evaluation doesn't require individual KAHMs (that are owned by different parties) but only the distance measures. This allows to design a KAHM based differentially private federated learning scheme as illustrated in Fig.~\ref{fig_KAHM_distributed_classifier}.     
\begin{remark}[Local Training Data with Missing Classes]\label{rem_missing_classes}
If the $q-$th party has zero $c-$th labelled data samples, the global classifier (\ref{eq_300320231343}) is evaluated taking $\Gamma_{ \mathcal{W}_{\tilde{Y}^q_c,n,L,S^q_c}}(y) = \infty$. 
\end{remark}
\begin{remark}[Addressing the Limitation of Passing Users' Inputs to the Clients]\label{rem_input_passing_to_clients}
A limitation of the federated learning scheme, as sketched in Fig.~\ref{fig_KAHM_distributed_classifier}, is that a user's input query is passed to all of the parties, causing an increased communication cost and concerns regarding the privacy of user's input. This limitation can be easily addressed by transferring all of the local models $\{\{\mathcal{W}_{\tilde{Y}^q_c,n,L,S^q_c}\}_{c=1}^C\}_{q=1}^Q$ to the cloud.    
\end{remark}

\section{Experiments}\label{sec_experiments}
The aim of the experiments is to 1) investigate the performance of KAHM based classifier (in Section~\ref{sec_grocery_dataset}); 2) evaluate the proposed privacy-preserving learning method in-terms of both accuracy and risk of membership inference attack (in Section~\ref{sec_membership_inference_attack_experiments}); 3) investigate the performance of the proposed differentially private federated learning scheme (in Section~\ref{sec_federated_learning_experiments}); and 4) study the computational time of KAHM in relation to increasing data dimension, subspace dimension, and data sample size (in Section~\ref{sec_computational_time_experiments}).    
\subsection{KAHM Based Classification of High-Dimensional Feature Vectors}\label{sec_grocery_dataset}
The ``Freiburg Groceries Dataset''~\shortcite{DBLP:journals/corr/JundAEB16} is considered to evaluate the performance of KAHM based classifier (Definition~\ref{def_KAHM_classifier}). This dataset has around 5000 labeled images of grocery products commonly sold in Germany. The images have been divided into 25 different categories of grocery products. Following the previous studies~\cite{8888203,10.1007/978-3-030-87101-7_14} on this dataset, image features were extracted from ``AlexNet'' and ``VGG-16'' networks (which are pre-trained Convolutional Neural Networks). The activations of the fully connected layer ``fc6'' in AlexNet constitute a $4096-$dimensional feature vector. Also, the activations of the fully connected layer ``fc6'' in VGG-16 constitute another $4096-$dimensional feature vector. The features extracted by both networks were joined together to form a $8192-$dimensional vector. The feature vectors were scaled along each dimension to take values between -1 and 1.   
\begin{table}[h]
\renewcommand{\arraystretch}{1.3}
\caption{Experiments on 5 different train-test splits of Freiburg groceries data: I}
\label{table_results_grocery_images} \centering
\begin{tabular}{c|*{6}{c}}
\hline
\bfseries \multirow{2}{*}{\bfseries methods} & \multicolumn{6}{c}{\bfseries accuracy (in \%) on test images}  \\
\cline{2-7} 
 &   1 &   2 &   3 &   4 &   5 &  mean \\ 
\hline \hline 
KAHM Classifier (Definition~\ref{def_KAHM_classifier}) & \textbf{89.29} & \textbf{87.16} & \textbf{87.00} & \textbf{86.73} & \textbf{87.09} & \textbf{87.46}  \\ \hline 
$\begin{array}{c} \mbox{membership-mappings} \\ \mbox{\cite{10.1007/978-3-030-87101-7_14}} \end{array}$ & 87.82 & \underline{87.06}  & \underline{85.88}  & \underline{85.63} & \underline{86.19} & \underline{86.52}  \\ \hline
$\begin{array}{c} \mbox{nonparametric fuzzy} \\ \mbox{image mapping}\\ \mbox{\cite{8888203}} \end{array}$ & \underline{88.21} & 86.64  & 85.36  & 85.13 & 85.79 &   86.23 \\ \hline
$\begin{array}{c} \mbox{Gaussian fuzzy-mapping} \\ \mbox{\cite{KUMAR20211}} \end{array}$ & 83.50  & 81.52  & 79.73  & 79.60  & 80.48  & 80.97  \\ \hline
SVM~\cite{10.1007/978-3-030-87101-7_14} & 77.90   & 79.54    & 77.17    & 76.98     & 76.98 & 77.71  \\ \hline
$1$-NN~\cite{10.1007/978-3-030-87101-7_14} & 78.00   & 77.97    & 77.38    & 76.58    & 76.28 & 77.24   \\ \hline
$\begin{array}{c} \mbox{Back-propagation} \\ \mbox{training of a deep network} \\ \mbox{\cite{10.1007/978-3-030-87101-7_14}} \end{array}$ & 75.25 & 77.24 & 72.67  & 73.37 & 71.57 & 74.02 \\ \hline
$2$-NN~\cite{10.1007/978-3-030-87101-7_14} & 73.48  & 73.38  & 70.11  & 70.05  & 70.57 & 71.52  \\ \hline
$4$-NN~\cite{10.1007/978-3-030-87101-7_14} & 72.50 & 73.39  & 68.89  & 71.16  & 70.87 & 71.36  \\ \hline
$\begin{array}{c} \mbox{Random Forest}\\ \mbox{\cite{10.1007/978-3-030-87101-7_14}} \end{array}$ &  63.17 & 62.63 & 59.47 & 59.50 &  59.76 & 60.90\\ \hline
$\begin{array}{c} \mbox{Naive Bayes} \\ \mbox{\cite{10.1007/978-3-030-87101-7_14}} \end{array}$ & 56.78  & 56.78      & 53.74   & 55.08     & 56.26 & 55.73  \\ \hline
$\begin{array}{c} \mbox{Ensemble Learning} \\ \mbox{\cite{10.1007/978-3-030-87101-7_14}} \end{array}$ & 38.31  & 39.35 & 38.89  &  37.69    & 38.34 & 38.51  \\ \hline
$\begin{array}{c} \mbox{Decision Tree} \\ \mbox{\cite{10.1007/978-3-030-87101-7_14}} \end{array}$ & 31.34   & 30.59    & 32.14    & 31.06    & 30.73 & 31.17  \\ 
\hline \hline
\end{tabular}
\end{table}
 \begin{table}[h]
\renewcommand{\arraystretch}{1.3}
\caption{Experiments on 5 different train-test splits of Freiburg groceries data: II}
\label{table_roc_grocery_images} \centering
\begin{tabular}{c|*{5}{c}}
\hline
\bfseries \multirow{2}{*}{\bfseries methods} & \multicolumn{5}{c}{\bfseries area under ROC curve (averaged per class)}  \\
\cline{2-6} 
 &  1 &  2 &   3 &   4 &   5 \\ 
\hline \hline 
KAHM (Definition~\ref{def_matching_score}) & \textbf{0.9901}  &  \textbf{0.9925}  &  \textbf{0.9970}  &  \textbf{0.9969}  &  \textbf{0.9945}  \\ \hline 
$\begin{array}{c} \mbox{nonparametric fuzzy} \\ \mbox{image mapping}\\ \mbox{\cite{8888203}} \end{array}$ & \underline{0.9818}  &  \underline{0.9775}   &  \underline{0.9754} & 0.9767 & \underline{0.9761} \\ \hline
$\begin{array}{c} \mbox{deep fuzzy nonparametric} \\ \mbox{model \cite{ZHANG2022128}} \end{array}$ & 0.9612 & 0.9574 & 0.9601 & 0.9582 & 0.9531 \\ \hline
$\begin{array}{c} \mbox{SVM}\\ \mbox{\cite{8888203}} \end{array}$ & 0.9806    & 0.9766     & 0.9711     & \underline{0.9777} & 0.9760   \\ \hline
$\begin{array}{c} \mbox{Random Forest}\\ \mbox{\cite{8888203}} \end{array}$ & 0.9489  & 0.9510   & 0.9372   & 0.9437   & 0.9466 \\ \hline
$\begin{array}{c} \mbox{$4$-NN}\\ \mbox{\cite{8888203}} \end{array}$ & 0.9425  & 0.9336   & 0.9325   & 0.9378   & 0.9280   \\ \hline
$\begin{array}{c} \mbox{$2$-NN}\\ \mbox{\cite{8888203}} \end{array}$ & 0.9219    & 0.9125   & 0.9118   & 0.9117   & 0.9048   \\ \hline
$\begin{array}{c} \mbox{Naive Bayes}\\ \mbox{\cite{8888203}} \end{array}$ & 0.8999  & 0.9100   & 0.8866   & 0.9013   & 0.8908  \\ \hline
$\begin{array}{c} \mbox{$1$-NN}\\ \mbox{\cite{8888203}} \end{array}$ & 0.8881    & 0.8802     & 0.8803     & 0.8837    & 0.8752    \\ \hline
$\begin{array}{c} \mbox{Ensemble Learning}\\ \mbox{\cite{8888203}} \end{array}$ & 0.8856  & 0.8896   & 0.8813   & 0.8818   & 0.8776 \\ \hline
$\begin{array}{c} \mbox{Decision Tree}\\ \mbox{\cite{8888203}} \end{array}$ & 0.6591  & 0.6473   & 0.6528   & 0.6539   & 0.6443   \\ 
\hline \hline
\end{tabular}
\end{table}

The authors of~\cite{DBLP:journals/corr/JundAEB16} provide five different train-test splits of images to evaluate the classification performance. For each of the five train-test data splits, training feature vectors of each class are modeled through a separate wide conditionally deep KAHM taking subspace dimension $n = 20$, number of layers $L = 5$, and number of branches $S$ as given in (\ref{eq_190220231832}). The performance of the proposed KAHM based classifier is compared in Table~\ref{table_results_grocery_images} with previous studies on this dataset. A related application is of detecting the presence of an individual grocery category in an image based on the value of KAHM based class-matching score (i.e. Definition~\ref{def_matching_score}). To study the application potential of proposed class-matching score, the receiver operating characteristic (ROC) curves are plotted for test images taking a particular image category as positive class. Table~\ref{table_roc_grocery_images} reports the performances of different methods evaluated in-term of area under ROC curve. The best performance of the KAHM based classifier on each of the five train-test data splits is observed in Table~\ref{table_results_grocery_images} and Table~\ref{table_roc_grocery_images}. The proposed KAHM based classifier is more competitive than the previously studied methods on this dataset.    

\subsection{KAHM Based Differentially Private Classification with Fabricated Data}\label{sec_membership_inference_attack_experiments}
The proposed KAHM based approach to privacy-preserving classification is studied on different datasets where the performance of the classifier is evaluated in-terms of both accuracy on test data and the value of membership-inference score. The membership-inference score, $mis$ (\ref{eq_270220231128}), is computed using a density-difference estimation method~\cite{10.1162/NECO_a_00492}.   
\subsubsection{MNIST Dataset}
A handwritten digits recognition problem is considered with the widely used MNIST dataset. The dataset contains $28 \times 28$ sized images divided into training set of 60000 images and testing set of 10000 images. The images' pixel values are divided by 255 to normalize the values in the range from $0$ to $1$. The $28 \times 28$ normalized values of each image are flattened to an equivalent $784-$dimensional data point.  
\begin{table}[h]
\renewcommand{\arraystretch}{1.3}
\caption{Results of privacy-preserving learning experiments on MNIST dataset}
\label{table_results_mnist} \centering
\begin{tabular}{c|c|c|c|c}
\hline
$(\epsilon,n)$ & $\begin{array}{c}\mbox{accuracy by}\\ \mbox{classifier}\\ \mbox{(Def. \ref{def_diff_priv_KAHM_classifier})} \end{array}$ & $\begin{array}{c}\mbox{accuracy by}\\ \mbox{classifier}\\ \mbox{(Def. \ref{def_diff_priv_KAHM_classifier_ref})} \end{array}$ & $\begin{array}{c}\mbox{$mis$ by}\\ \mbox{classifier}\\ \mbox{(Def. \ref{def_diff_priv_KAHM_classifier})} \end{array}$  & $\begin{array}{c}\mbox{$mis$ by}\\ \mbox{classifier}\\ \mbox{(Def. \ref{def_diff_priv_KAHM_classifier_ref})} \end{array}$ \\  \hline \hline
$(1,20)$ & 0.9491 & 0.9453 & 0.00000 & 0.00017 \\ \hline
$(1.5,20)$ & 0.9644 & 0.9645 & 0.00025 & 0.00035 \\ \hline
$(2, 20)$ & 0.9711 & 0.9709 & 0.00074 & 0.00076 \\ \hline
$(3, 20)$ & 0.9779 & 0.9776 & 0.00299 & 0.00589 \\ \hline
$(4, 20)$ & 0.9802 & 0.9796 & 0.00687 & 0.01605 \\ \hline
$(5,20)$ & 0.9820 & 0.9805 & 0.01275 & 0.03361 \\ \hline
$(8,20)$ & 0.9833 & 0.9845 & 0.02968 & 0.11685 \\ \hline
$(16, 20)$ & 0.9854 & 0.9858 & 0.05459 & 0.30601 \\ \hline
$(32, 20)$ & 0.9851 & 0.9858 & 0.06171 & 0.35388 \\ \hline
$(32,5)$ & 0.9680 & 0.9676 & 0.00503 & 0.01225 \\ \hline
$(32,10)$ & 0.9799 & 0.9810 & 0.01794 & 0.08117 \\ \hline
$(32,15)$ & 0.9851 & 0.9861 & 0.03501 & 0.20903 \\ \hline
$(32,20)$ & 0.9851 & 0.9854 & 0.06145 & 0.35451 \\ \hline
$(32,25)$ & 0.9845 & 0.9854 & 0.09111 & 0.49186 \\ \hline
 & \textbf{0.9772} (mean) & \textbf{0.9771} (mean) & \textbf{0.02715} (mean) & \textbf{0.14160} (mean) \\
\hline \hline
\end{tabular}
\end{table}

\begin{figure}
\centerline{\subfigure[comparison of accuracies]{\includegraphics[width=0.45\textwidth]{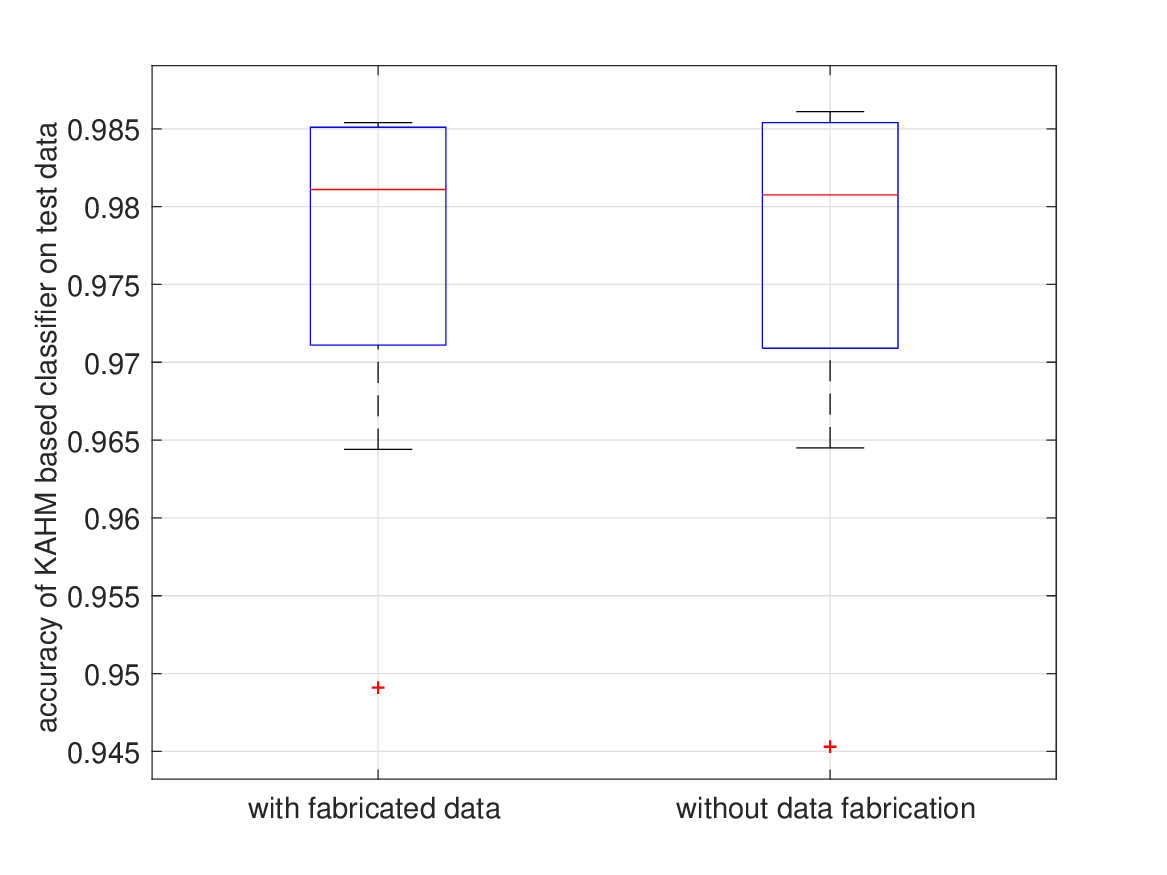}\label{results_experiments_mnist_1}}  \hfil \subfigure[comparison of $mis$ values]{\includegraphics[width=0.45\textwidth]{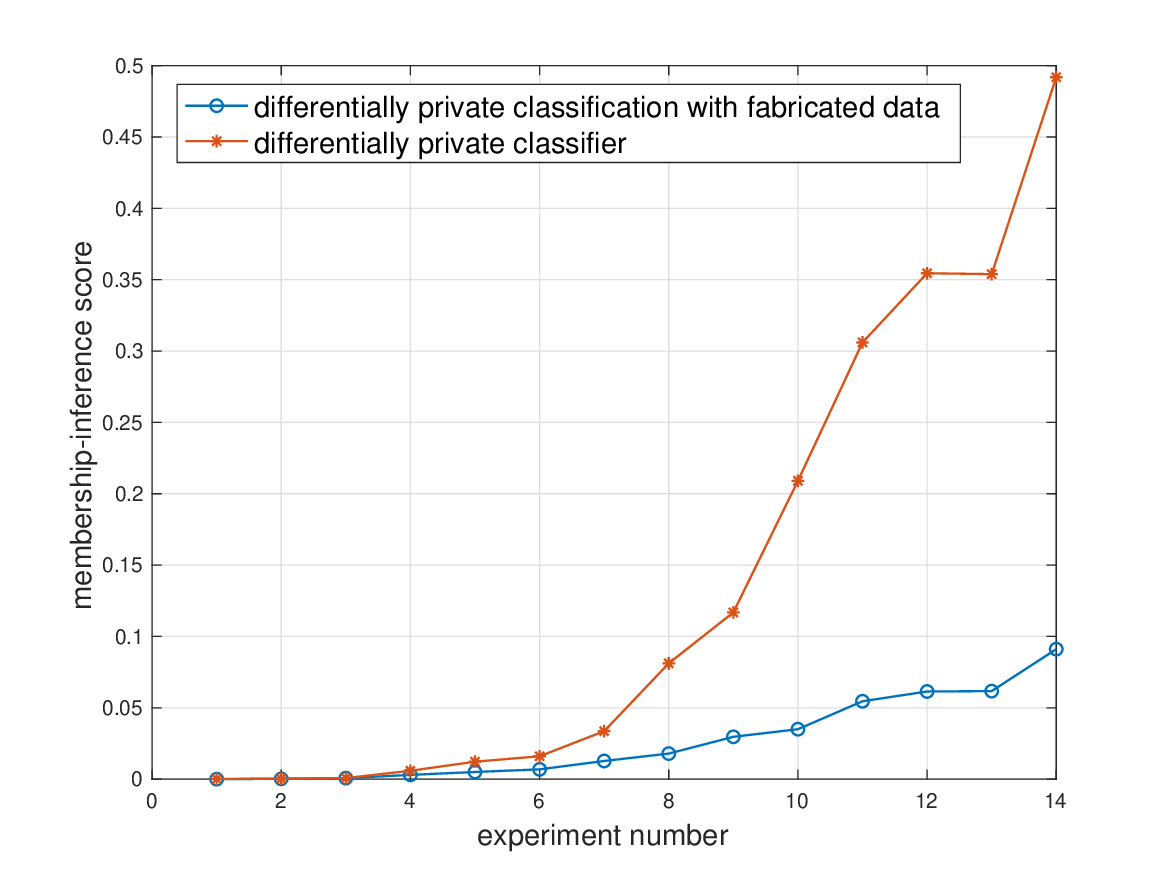}\label{results_experiments_mnist_2}}  }
\caption{The effect of using fabricated data in classification for MNIST.}
\label{fig_results_mnist}
\end{figure}

\begin{figure}
\centerline{\subfigure[histograms of distances of training and test data points from the affine hull of training samples]{\includegraphics[width=0.45\textwidth]{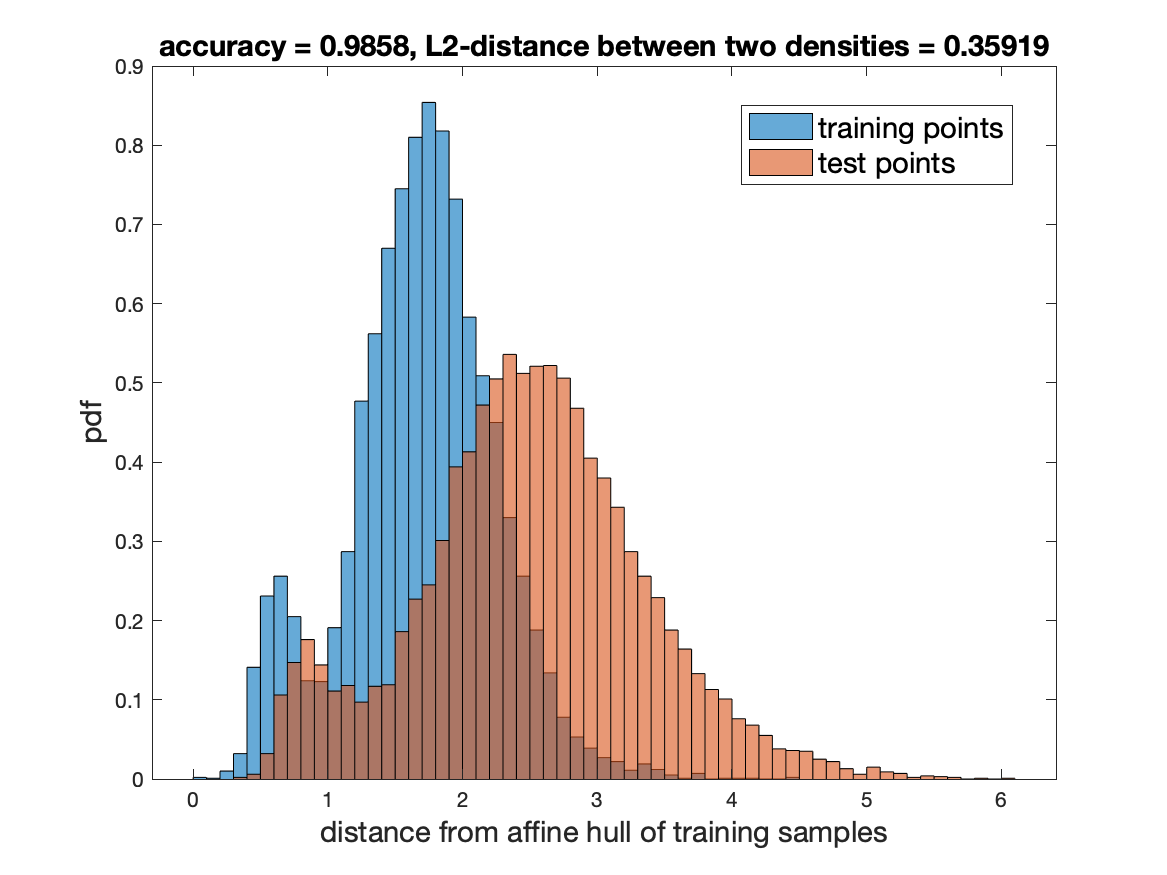}\label{results_experiments_mnist_3}}  \hfil \subfigure[histograms of distances of training and test data points from the affine hull of fabricated samples]{\includegraphics[width=0.45\textwidth]{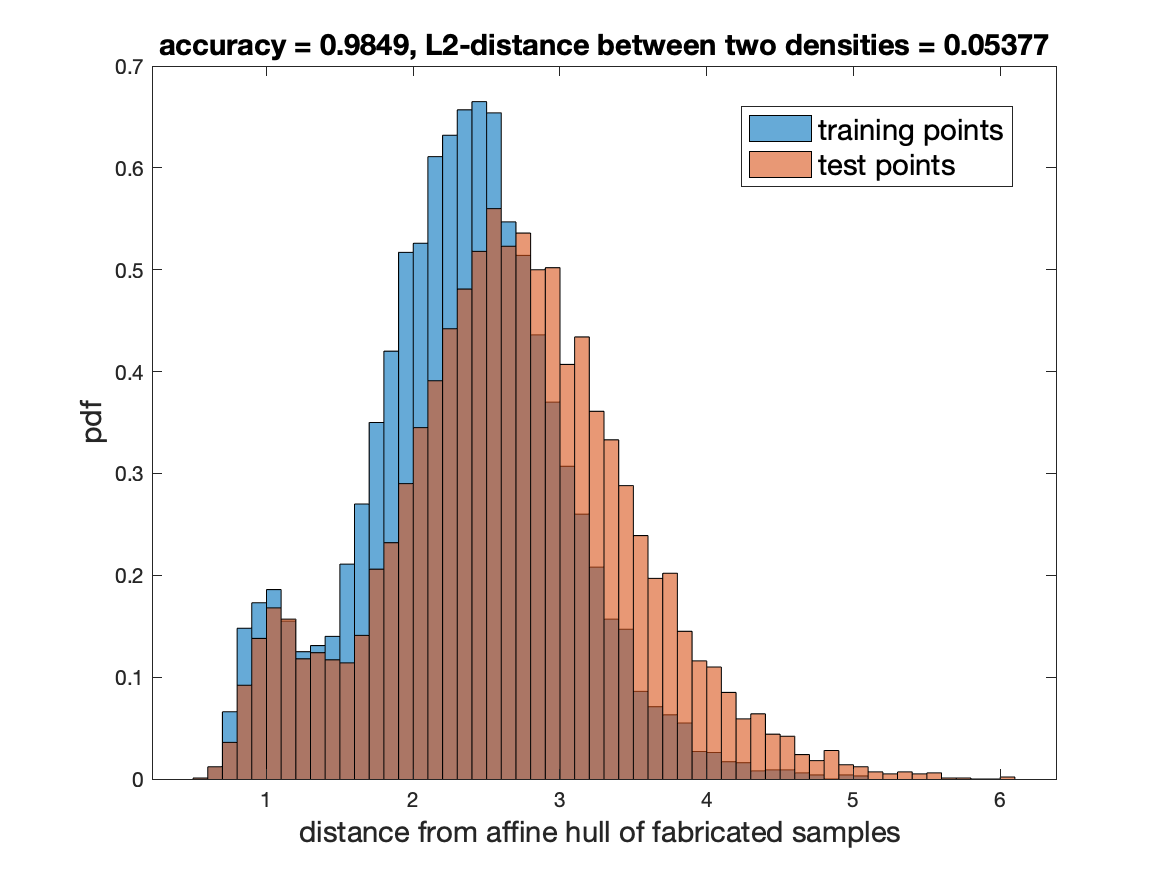}\label{results_experiments_mnist_4}}  }
\caption{Illustration of the reduction in $mis$ value through data fabrication for MNIST.}
\label{fig_results_msi_mnist}
\end{figure}

The performances of both differentially private classifier (Definition~\ref{def_diff_priv_KAHM_classifier_ref}) and differentially private classifier based on fabricated data (Definition~\ref{def_diff_priv_KAHM_classifier}) are evaluated for different values of privacy-loss bound $\epsilon$ and subspace dimension $n$ while keeping the number of layers $L = 5$ and number of branches $S$ as given in (\ref{eq_190220231832}). Table~\ref{table_results_mnist} reports the obtained results. For a visualization of the results, Fig.~\ref{fig_results_mnist} compares the accuracy and $mis$ values obtained by the two methods. It is observed in Fig.~\ref{fig_results_mnist} that while both classifiers (with and without using fabricated data) achieve nearly the same level of accuracy (as observed in Fig.~\ref{results_experiments_mnist_1}), the membership-inference score is considerably lower in the case of fabricated data (as observed in Fig.~\ref{results_experiments_mnist_2}). The use of fabricated data reduces greatly the averaged $mis$ from 0.14160 to 0.02715 with the marginal change in averaged accuracy from 0.9771 to 0.9772. As an example, Fig.~\ref{fig_results_msi_mnist} shows the histograms of distances of training and test points from the affine hull of training samples (in Fig.~\ref{results_experiments_mnist_3}) and from the affine hull of fabricated samples (in Fig.~\ref{results_experiments_mnist_4}). The use of fabricated data in this example reduces the $mis$ from 0.35919 to 0.05377 with loss of accuracy from 0.9858 to 0.9849.    
\subsubsection{Freiburg Groceries Dataset}
The Freiburg groceries dataset is revisited to study the KAHM based differentially private classifiers. The $8192-$dimensional feature vectors, extracted as stated in Section~\ref{sec_grocery_dataset}, are considered to study both differentially private classifier (Definition~\ref{def_diff_priv_KAHM_classifier_ref}) and differentially private classifier based on fabricated data (Definition~\ref{def_diff_priv_KAHM_classifier}) for different values of privacy-loss bound $\epsilon$ and subspace dimension $n$ while keeping the number of layers $L = 5$ and number of branches $S$ as given in (\ref{eq_190220231832}).               
\begin{table}[h]
\renewcommand{\arraystretch}{1.3}
\caption{Results of privacy-preserving learning experiments on Freiburg groceries dataset}
\label{table_results_grocery_privacy} \centering
\begin{tabular}{c|c|c|c|c}
\hline
$(\epsilon,n)$ & $\begin{array}{c}\mbox{accuracy by}\\ \mbox{classifier}\\ \mbox{(Def. \ref{def_diff_priv_KAHM_classifier})} \end{array}$ & $\begin{array}{c}\mbox{accuracy by}\\ \mbox{classifier}\\ \mbox{(Def. \ref{def_diff_priv_KAHM_classifier_ref})} \end{array}$ & $\begin{array}{c}\mbox{$mis$ by}\\ \mbox{classifier}\\ \mbox{(Def. \ref{def_diff_priv_KAHM_classifier})} \end{array}$  & $\begin{array}{c}\mbox{$mis$ by}\\ \mbox{classifier}\\ \mbox{(Def. \ref{def_diff_priv_KAHM_classifier_ref})} \end{array}$ \\  \hline \hline
$(5,20)$ & 0.8016 & 0.8153 & 0.02600 & 0.04207 \\ \hline
$(8, 20)$ & 0.8556 & 0.8733 & 0.12512 & 0.20650 \\ \hline
$(16, 20)$ & 0.8792 & 0.8919 & 0.27377 & 0.51909 \\ \hline
$(32, 20)$ & 0.8811 & 0.8919 & 0.25753 & 0.40348 \\ \hline
$(32, 5)$ & 0.7996 & 0.8212 & 0.04019 & 0.08048 \\ \hline
$(32, 10)$ & 0.8595 & 0.8694 & 0.12220 & 0.24263 \\ \hline
$(32, 15)$ & 0.8752 & 0.8841 & 0.19986 & 0.34380 \\ \hline
$(32, 20)$ & 0.8811 & 0.8919 & 0.25753 & 0.40348 \\ \hline
$(32, 25)$ & 0.8782 & 0.8929 & 0.29724 & 0.45099 \\ \hline
& \textbf{0.8568} (mean) & \textbf{0.8702} (mean) & \textbf{0.17772} (mean) & \textbf{0.29917} (mean) \\
\hline \hline
\end{tabular}
\end{table}
\begin{figure}
\centerline{\subfigure[Freiburg groceries dataset]{\includegraphics[width=0.45\textwidth]{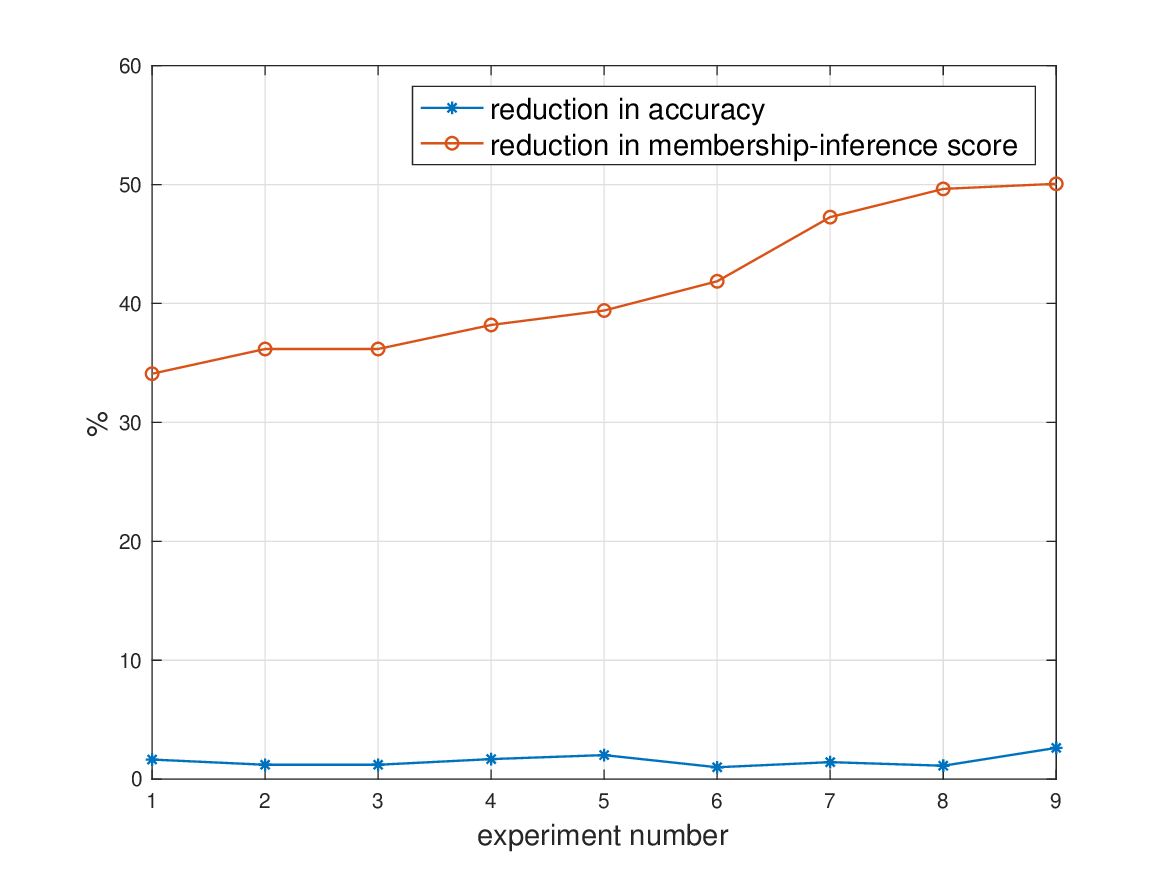}\label{results_experiments_grocery_1}} \hfil \subfigure[heart rate variability dataset]{\includegraphics[width=0.45\textwidth]{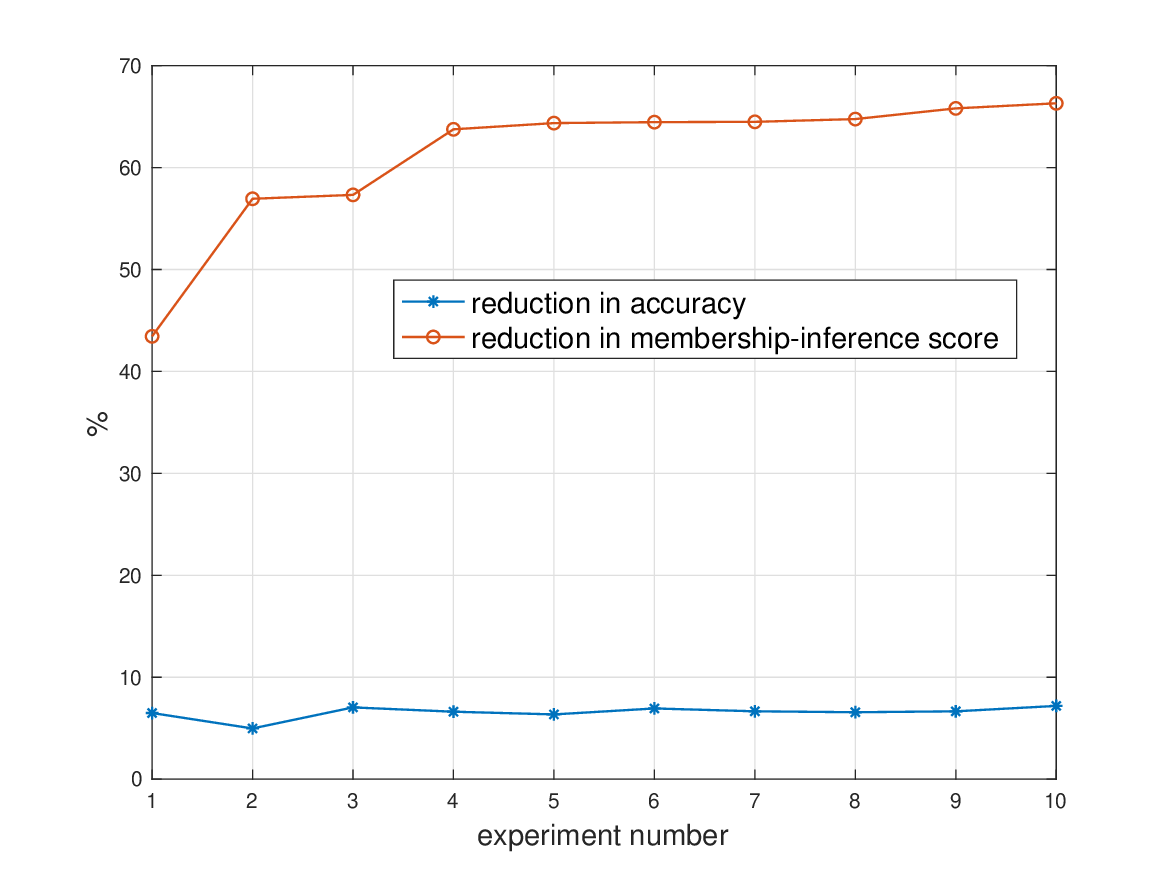}\label{results_experiments_stress_1}}}
\caption{The plots of \% reduction in accuracy and $mis$ values as a result of the use of fabricated data.}
\end{figure}

\begin{figure}
\centerline{\subfigure[histograms of distances of training and test data points from the affine hull of training samples]{\includegraphics[width=0.45\textwidth]{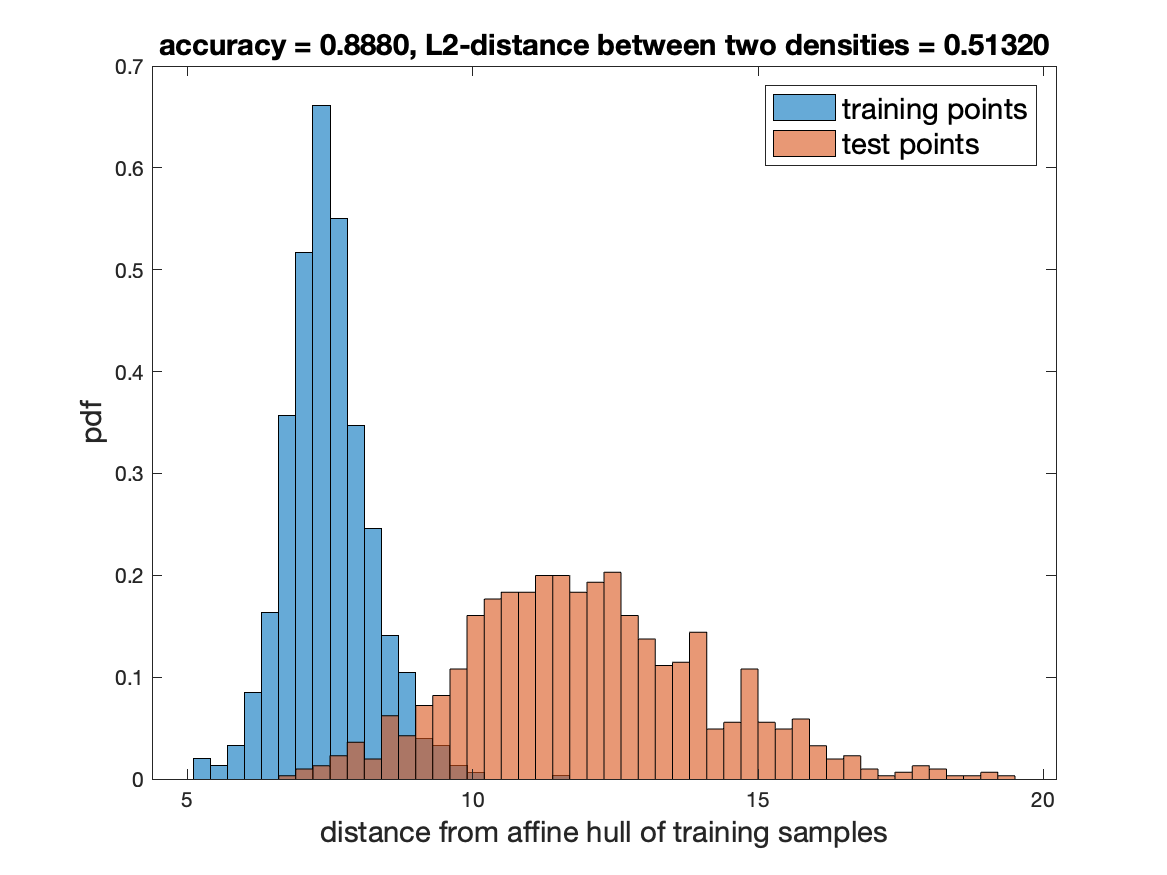}\label{results_experiments_grocery_2}}  \hfil \subfigure[histograms of distances of training and test data points from the affine hull of fabricated samples]{\includegraphics[width=0.45\textwidth]{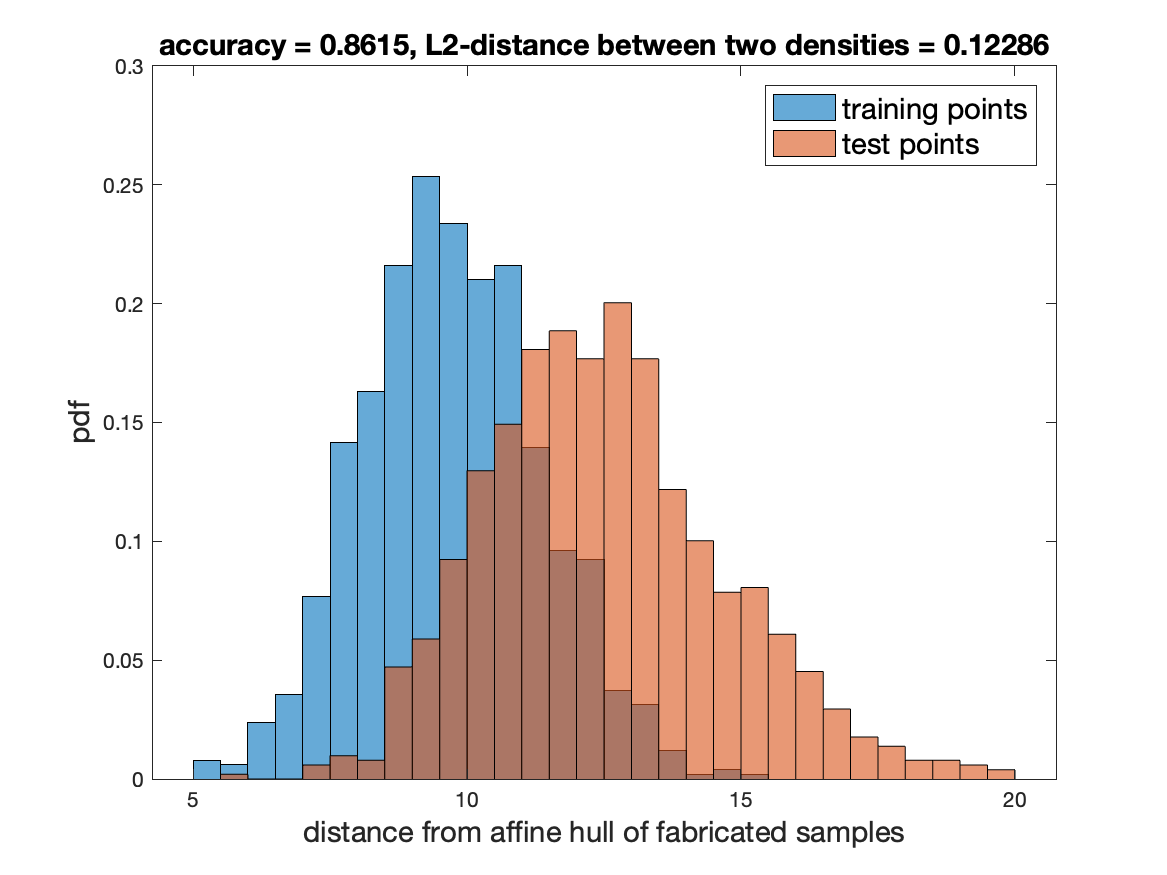}\label{results_experiments_grocery_3}}  }
\caption{Illustration of the reduction in $mis$ value through data fabrication on Freiburg groceries dataset.}
\label{fig_results_msi_grocery}
\end{figure}

The experimental results are reported in Table~\ref{table_results_grocery_privacy}. The averaged $msi$ decreases from 0.29917 to 0.17772 together with the loss of averaged accuracy from 0.8702 to 0.8568. Fig.~\ref{results_experiments_grocery_1} illustrates the results via plotting the \% reduction in both accuracy and $mis$ values due to the use of fabricated data. It is observed from Fig.~\ref{results_experiments_grocery_1} that the use of fabricated data leads to a considerable reduction in $mis$ value with relatively much smaller loss of accuracy. This is demonstrated through an example in Fig.~\ref{fig_results_msi_grocery} where the histograms of distances of training and test points from the affine hull of training samples (in Fig.~\ref{results_experiments_grocery_2}) and from the affine hull of fabricated samples (in Fig.~\ref{results_experiments_grocery_3}) are plotted. As the result of using fabricated data in this example, the $msi$ reduces from 0.51320 to 0.12286 together with relatively smaller loss of accuracy from 0.8880 to 0.8615.    
\subsubsection{A Real Biomedical Dataset}
A dataset related to the mental stress detection problem~\cite{9216097,10012502} is considered to evaluate the proposed differentially private classifier based on fabricated data. This dataset consists of heart rate interval measurements of different subjects together with a stress-score on a scale from 0 to 100. The aim is to detect stress on an individual based on the analysis of recorded sequence of R-R intervals, $\{RR^i\}_i$. The R-R data vector at $i-$th time-index, $y^i$, is defined as $y^i  =  \left[\begin{IEEEeqnarraybox*}[][c]{,c/c/c/c,} RR^i & RR^{i-1} & \cdots & RR^{i-d}\end{IEEEeqnarraybox*} \right]^T$. That is, the current interval and history of previous $d$ intervals constitute the data vector. Assuming an average heartbeat of 72 beats per minute, $d$ is chosen as equal to $72 \times 3 = 216$ so that R-R data vector consists of on an average 3-minutes long R-R intervals sequence. Following~\cite{9216097,10012502}, a dataset, say $\{y^i\}_{i}$, is built via 1) preprocessing the R-R interval sequence $\{RR^i\}_i$ with an impulse rejection filter for artifacts detection, and 2) excluding the R-R data vectors containing artifacts from the dataset. A class-label of either ``\emph{no-stress}'' or ``\emph{under-stress}'' is assigned to each data sample $y^i$ based on the stress-score. For each subject, 50\% of the data samples serve as training data while remaining as test data. Both differentially private classifier (Definition~\ref{def_diff_priv_KAHM_classifier_ref}) and differentially private classifier based on fabricated data (Definition~\ref{def_diff_priv_KAHM_classifier}) are considered for the stress detection problem for different values of privacy-loss bound $\epsilon$ and subspace dimension $n$ while keeping the number of layers $L = 5$ and number of branches $S$ as given in (\ref{eq_190220231832}). 
\begin{table}[h]
\renewcommand{\arraystretch}{1.3}
\caption{Results of privacy-preserving learning experiments on a biomedical dataset}
\label{table_results_stress} \centering
\begin{tabular}{c|c|c|c|c}
\hline
$(\epsilon,n)$ & $\begin{array}{c}\mbox{accuracy by}\\ \mbox{classifier}\\ \mbox{(Def. \ref{def_diff_priv_KAHM_classifier})} \end{array}$ & $\begin{array}{c}\mbox{accuracy by}\\ \mbox{classifier}\\ \mbox{(Def. \ref{def_diff_priv_KAHM_classifier_ref})} \end{array}$ & $\begin{array}{c}\mbox{$mis$ by}\\ \mbox{classifier}\\ \mbox{(Def. \ref{def_diff_priv_KAHM_classifier})} \end{array}$  & $\begin{array}{c}\mbox{$mis$ by}\\ \mbox{classifier}\\ \mbox{(Def. \ref{def_diff_priv_KAHM_classifier_ref})} \end{array}$ \\  \hline \hline
$(8, 5)$ & 0.8485 & 0.9074 & 0.02773 & 0.04903 \\ \hline
$(16,5)$ & 0.8817 & 0.9484 & 0.06739 & 0.15790 \\ \hline
$(24, 5)$ & 0.8925 & 0.9557 & 0.07625 & 0.21036 \\ \hline
$(32, 5)$ & 0.8964 & 0.9572 & 0.08310 & 0.23319 \\ \hline
$(32, 1)$ & 0.8608 & 0.9058 & 0.00973 & 0.02259 \\ \hline
$(32, 3)$ & 0.8777 & 0.9457 & 0.04045 & 0.12006 \\ \hline
$(32, 5)$ & 0.8931 & 0.9568 & 0.07603 & 0.22235 \\ \hline
$(32, 7)$ & 0.8960 & 0.9599 & 0.11834 & 0.33322 \\ \hline
$(32, 10)$ & 0.8938 & 0.9566 & 0.17230 & 0.48887 \\ \hline
$(32,15)$ & 0.8795 & 0.9451 & 0.24173 & 0.68009 \\ \hline
& \textbf{0.8820} (mean) & \textbf{0.9439} (mean) & \textbf{0.09131} (mean) & \textbf{0.25177} (mean) \\
\hline \hline
\end{tabular}
\end{table}

Table~\ref{table_results_stress} reports the experimental results. The experimental results have been visualized in Fig.~\ref{results_experiments_stress_1} via plotting the \% reduction in both accuracy and $mis$ values due to the use of fabricated data. It is observed from Fig.~\ref{results_experiments_stress_1} that the use of fabricated data reduces considerably the $mis$ value with relatively much smaller loss of accuracy. As a result of using fabricated data, the averaged $mis$ decreases by 63.7328\% (from 0.25177 to 0.09131 in absolute terms) together with averaged accuracy loss of 6.5579\% (from 0.9439 to 0.8820 in absolute terms).     
\subsection{Federated Learning}\label{sec_federated_learning_experiments}
MNIST dataset (containing the samples of 10 classes) is reconsidered under the following federated learning scenarios: \underline{Scenario 1:} The training data are distributed among 10 parties such that all samples of a class are possessed by only a single party. That is, a party has all the samples of a class. \underline{Scenario 2:} The training data are distributed among 20 parties such that samples of a class are shared equally between two parties. That is, a party has 50\% samples of a class. \underline{Scenario 3:} The training data are distributed randomly independent of the classes among $Q$ number of parties where $Q \in \{2, 5, 10, 20, 50, 100\}$. For all of the considered scenarios, the local classifiers are built for privacy-loss bound $\epsilon \in \{1,1,5,2,3,4,5,8,16\}$, subspace dimension $n = 20$, number of layers $L = 5$, and number of branches $S$ as given in (\ref{eq_190220231832}). The performance of the global classifier (\ref{eq_300320231343}) is evaluated on test data. As a reference, the performance in the case of non-federated learning (i.e. in the case of centralized data) is also evaluated using classifier (\ref{eq_030220231011}).
\begin{table}
\renewcommand{\arraystretch}{1.3}
\caption{Results of federated learning experiments under Scenario 1 on MNIST dataset}
\label{table_results_federated_1} \centering
\begin{tabular}{c|c|c|c}
\hline
$\begin{array}{c}\mbox{privacy-loss}\\ \mbox{bound $\epsilon$} \end{array}$ & $\begin{array}{c}\mbox{accuracy by}\\ \mbox{classifier (\ref{eq_300320231343})} \\ \mbox{(distributed data)}  \end{array}$  & $\begin{array}{c}\mbox{accuracy by}\\ \mbox{classifier (\ref{eq_030220231011})} \\ \mbox{(centralized data)}  \end{array}$ & $\begin{array}{c}\mbox{change in accuracy}\\ \mbox{due to data} \\ \mbox{being distributed} \end{array}$\\ \hline \hline   
1 & 0.9460 & 0.9478 & -0.0018 \\ \hline
1.5 & 0.9633 & 0.9629 & 0.0004 \\ \hline
2 & 0.9710 & 0.9707 & 0.0003 \\ \hline
3 & 0.9783 & 0.9779 & 0.0004 \\ \hline
4 & 0.9790 & 0.9798 & -0.0008 \\ \hline
5 & 0.9813 & 0.9819 &-0.0006 \\ \hline
8 & 0.9829 & 0.9837 & -0.0008 \\ \hline
16 & 0.9847 & 0.9851 & -0.0004 \\ \hline
& \textbf{0.9733} (mean) & \textbf{0.9737} (mean) & \textbf{-0.0004} (mean) \\  
\hline \hline
\end{tabular}
\end{table}
\begin{table}
\renewcommand{\arraystretch}{1.3}
\caption{Results of federated learning experiments under Scenario 2 on MNIST dataset}
\label{table_results_federated_2} \centering
\begin{tabular}{c|c|c|c}
\hline
$\begin{array}{c}\mbox{privacy-loss}\\ \mbox{bound $\epsilon$} \end{array}$ & $\begin{array}{c}\mbox{accuracy by}\\ \mbox{classifier (\ref{eq_300320231343})} \\ \mbox{(distributed data)}  \end{array}$  & $\begin{array}{c}\mbox{accuracy by}\\ \mbox{classifier (\ref{eq_030220231011})} \\ \mbox{(centralized data)}  \end{array}$ & $\begin{array}{c}\mbox{change in accuracy}\\ \mbox{due to data} \\ \mbox{being distributed} \end{array}$\\ \hline \hline   
1 & 0.9384  & 0.9478 & -0.0094  \\ \hline
1.5 & 0.9603  & 0.9629 & -0.0026  \\ \hline
2 & 0.9670  & 0.9707 & -0.0037  \\ \hline
3 & 0.9733  & 0.9779 & -0.0046 \\ \hline
4 & 0.9758 & 0.9798 & -0.0040  \\ \hline
5 & 0.9778 & 0.9819 & -0.0041  \\ \hline
8 & 0.9803 & 0.9837 & -0.0034  \\ \hline
16 & 0.9813  & 0.9851 & -0.0038  \\ \hline
& \textbf{0.9693} (mean) & \textbf{0.9737} (mean)  & \textbf{-0.0044} (mean)  \\
\hline \hline
\end{tabular}
\end{table}
\begin{table}
\renewcommand{\arraystretch}{1.3}
\caption{Results of 10 independent federated learning experiments under Scenario 3 on MNIST dataset for privacy-loss bound $\epsilon = 16$}
\label{table_results_federated_3} \centering
\begin{tabular}{c|c|c|c}
\hline
$\begin{array}{c}\mbox{number of}\\ \mbox{parties $Q$} \end{array}$ & $\begin{array}{c}\mbox{mean accuracy by}\\ \mbox{classifier (\ref{eq_300320231343})} \\ \mbox{(distributed data)}  \end{array}$ & $\begin{array}{c}\mbox{accuracy by}\\ \mbox{classifier (\ref{eq_030220231011})} \\ \mbox{(centralized data)}  \end{array}$  & $\begin{array}{c}\mbox{change in mean accuracy}\\ \mbox{due to data} \\ \mbox{being distributed} \end{array}$\\ \hline \hline   
2 & 0.9817 &  0.9847  & -0.0030   \\ \hline
5 & 0.9770   & 0.9847 & -0.0077   \\ \hline
10 & 0.9752  & 0.9847 & -0.0095   \\ \hline
20 & 0.9734   & 0.9847 & -0.0113  \\ \hline
50 & 0.9728   & 0.9847 & -0.0119   \\ \hline
100 & 0.9717  & 0.9847  & -0.0130   \\ \hline
& \textbf{0.9753} (mean) & \textbf{0.9847} (mean) & \textbf{-0.0094} (mean)  \\
\hline \hline
\end{tabular}
\end{table}

\begin{figure}
\centerline{\subfigure[Scenario 1]{\includegraphics[width=0.33\textwidth]{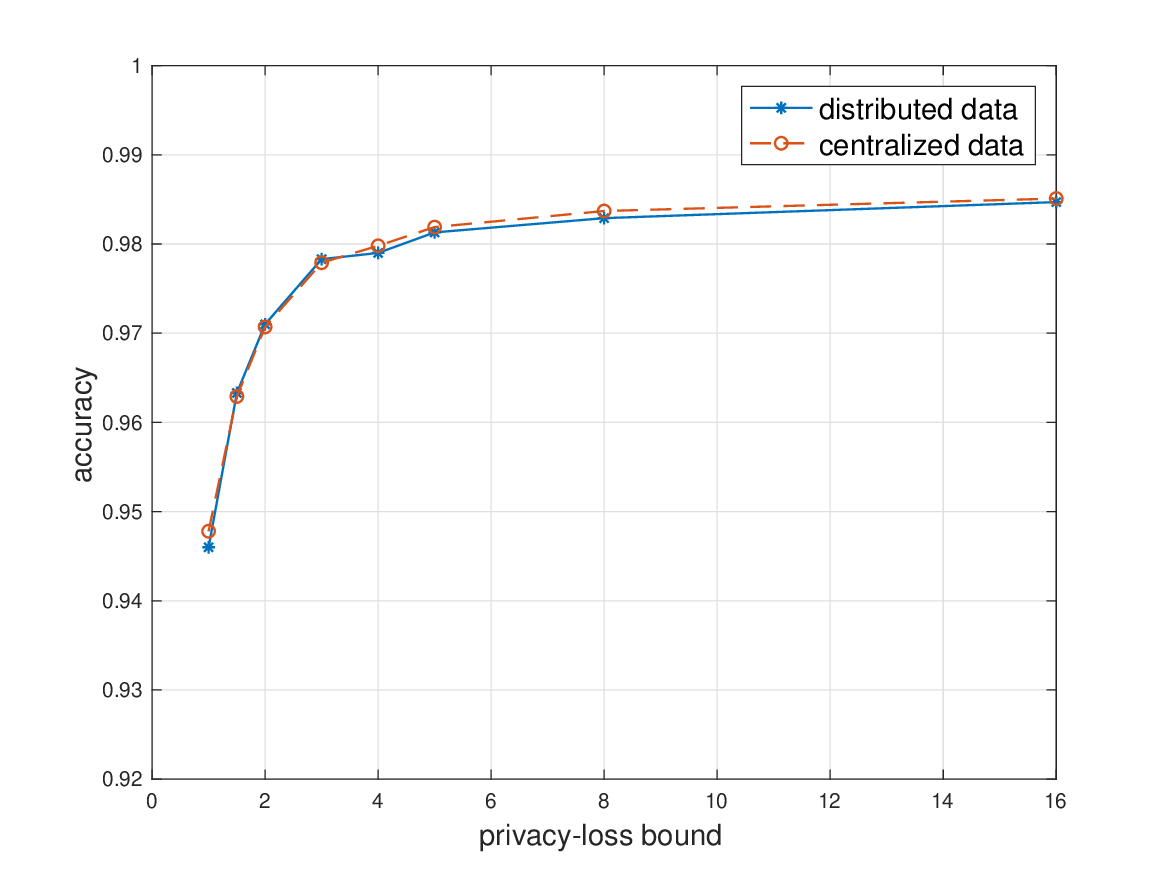}\label{results_federated_1}} \hfil \subfigure[Scenario 2]{\includegraphics[width=0.33\textwidth]{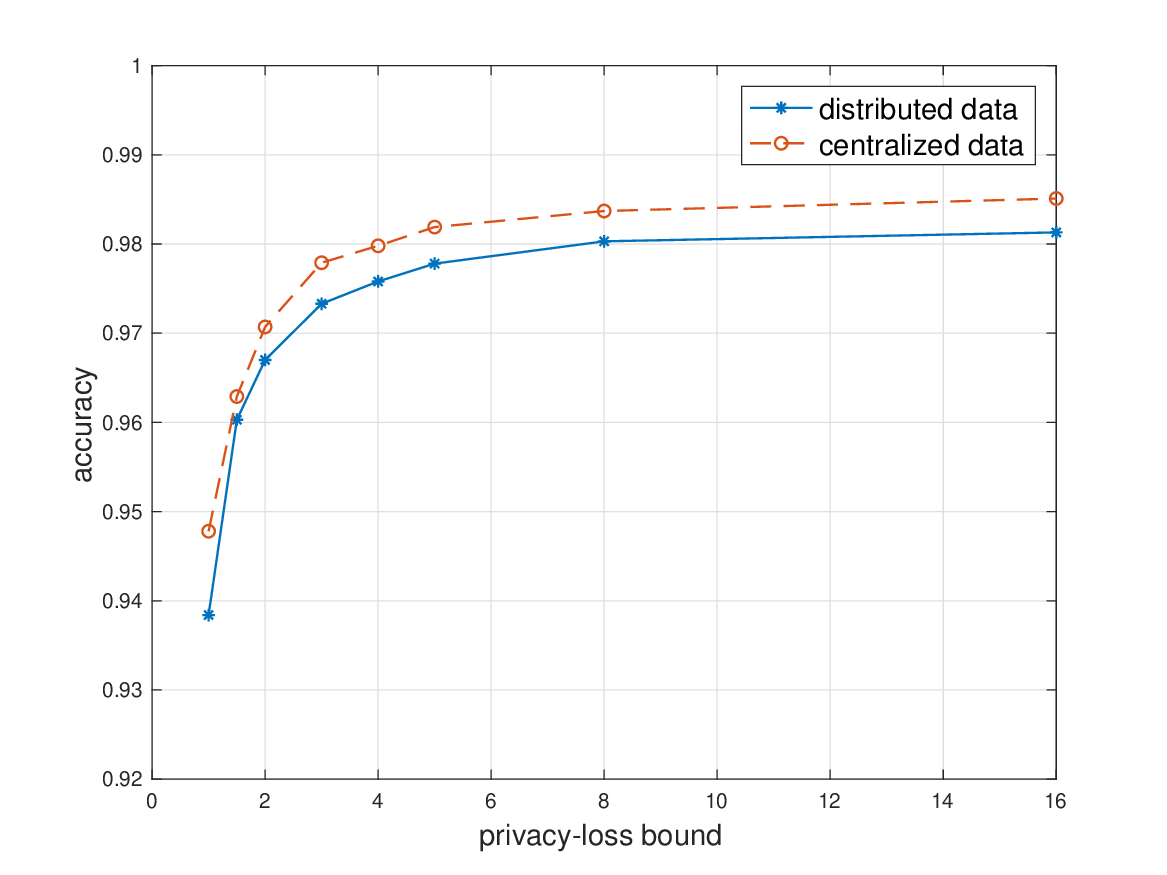}\label{results_federated_2}} \hfil \subfigure[Scenario 3]{\includegraphics[width=0.33\textwidth]{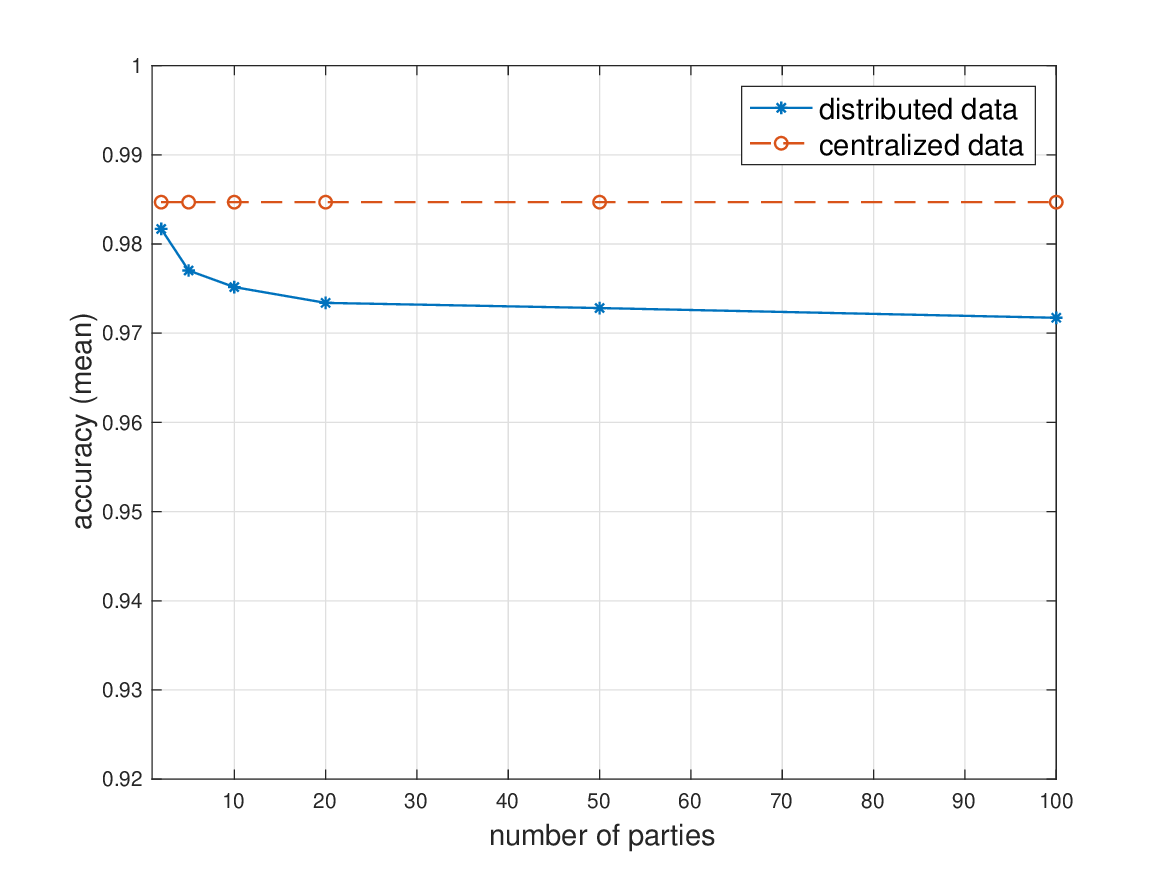}\label{results_federated_3}} }
\caption{The plots for the results of federated learning experiments on MNIST dataset.}
\label{fig_results_federated_learning}
\end{figure}

The obtained results are reported in Table~\ref{table_results_federated_1} and Fig.~\ref{results_federated_1} for Scenario 1, in Table~\ref{table_results_federated_2} and Fig.~\ref{results_federated_2} for Scenario 2, and in Table~\ref{table_results_federated_3} and Fig.~\ref{results_federated_3} for Scenario 3. Following observations are made from the obtained results: 1) In Scenario 1 (when samples of a class are not shared by parties), the federated learning performance is not different from that of learning with centralized data. This is expected, as for each class there exists only one local KAHM that serves as the global KAHM for that class as well. Since there remains no difference between the class specific global and local KAHMs, the performance in the federated setting remains unaffected. Thus, the change in accuracy due to data being distributed, as reported in Table~\ref{table_results_federated_1}, remains less than 0.0018. 2) In Scenario 2 (when samples of a class are shared by two parties), the performance under federated setting reduces slightly across the whole range of privacy-loss bound. It is observed from Table~\ref{table_results_federated_2} that the change in accuracy due to data being distributed, averaged over the considered range of privacy-loss bound, is equal to $-0.0044$. The accuracy loss of 0.0044 on an average  due to distributed data is marginal indicating a good performance. 3) In Scenario 3 (when samples of a class are shared by up to 100 parties), it is observed from Table~\ref{table_results_federated_3} that the decrease in performance due to data being distributed is not significantly high. Specifically, the data distribution among 100 parties did not cause a loss in accuracy of more than 0.013. This verifies the application potential of the proposed federated learning scheme. 
\subsection{Computational Time}\label{sec_computational_time_experiments}
We study the effect of sample size $N$ only up to $1000$ on the computational time of KAHM, as samples more than 1000 are divided into subsets and processed in parallel, as explained in Remark~\ref{rem_large_data}. Further, the effects of data dimension $p$ and subspace dimension $n$ are studied on the computational time of KAHM. For this, MATLAB R2017b simulations have been made on a MacBook Pro machine with a 2.2 GHz Intel Core i7 processor and 16 GB of memory. The simulations are made on the randomly generated data from the Gaussian distribution with mean 0 and variance 1 with
\begin{IEEEeqnarray}{rCl}
p & \in & \{10, 100, 500, 1000, 2500, 5000, 7500, 10000, 12500, 15000, 17500, 20000 \},\\
n & \in & \{5, 10, 20, 50, 75, 100, 200, 300, 400, 500\},\mbox{ and}\\
N & \in & \{100, 200, 300, 400, 500, 600, 700, 800, 900, 1000 \}.
 \end{IEEEeqnarray}   

\begin{figure}
\centerline{\subfigure[against data dimension $p$]{\includegraphics[width=0.35\textwidth]{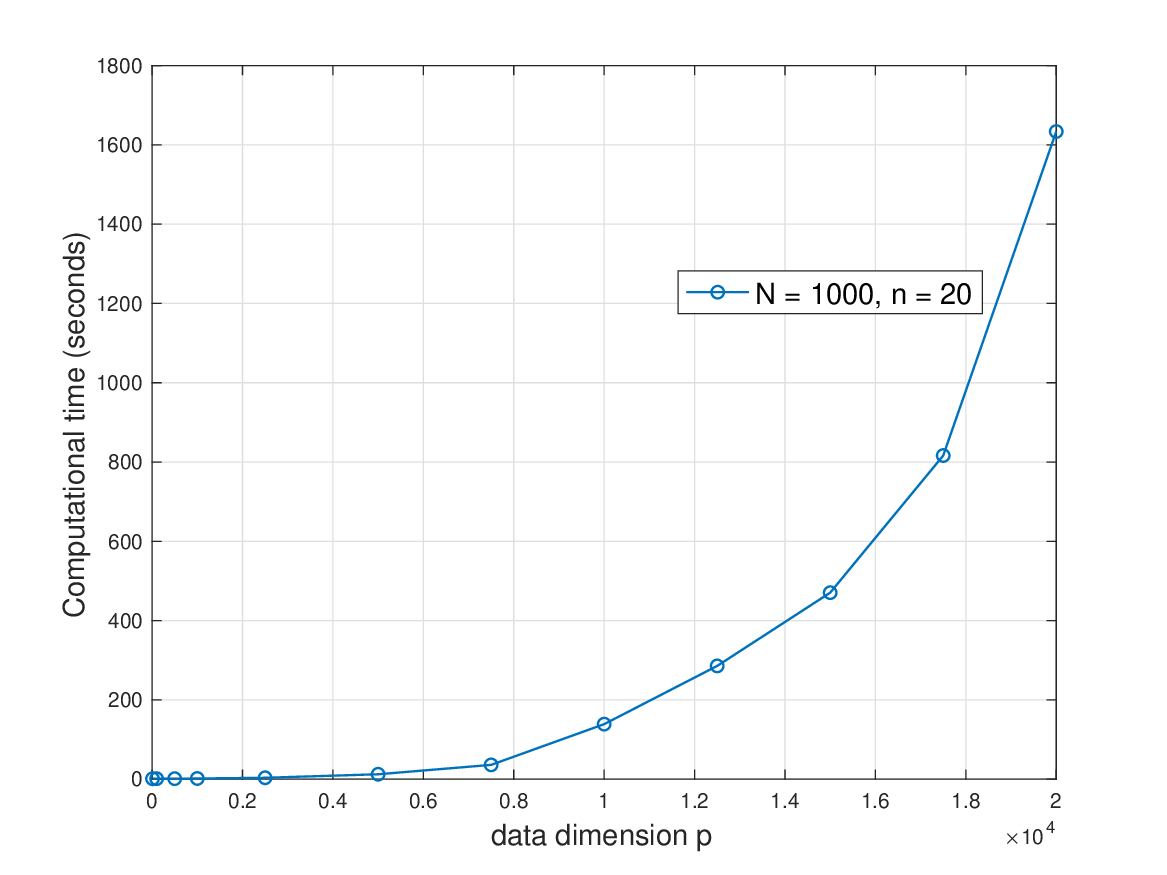}\label{results_computational_1}}  \hfil \subfigure[against subspace dimension $n$]{\includegraphics[width=0.35\textwidth]{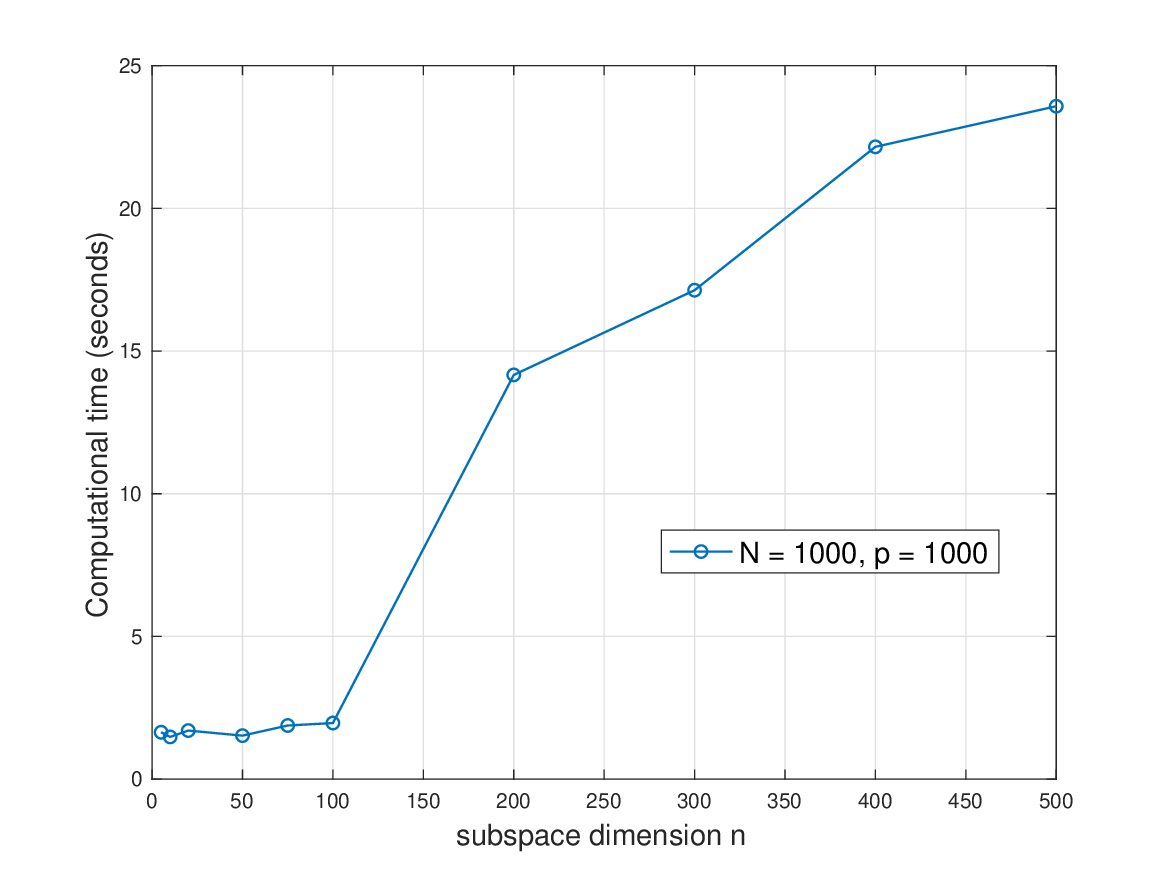}\label{results_computational_2}}}
\centerline{ \subfigure[against number of samples $N$]{\includegraphics[width=0.35\textwidth]{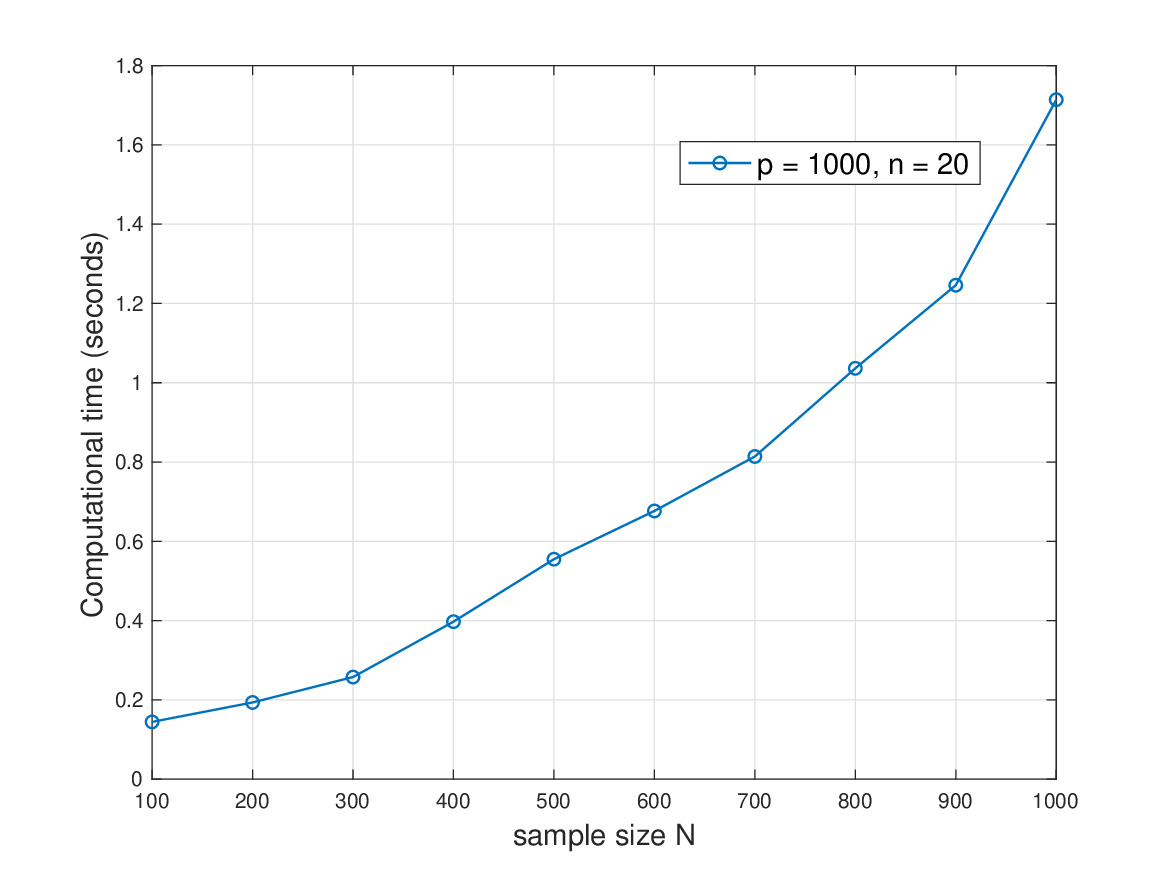}\label{results_computational_3}} \hfil  \subfigure[against data size $N \times p$]{\includegraphics[width=0.35\textwidth]{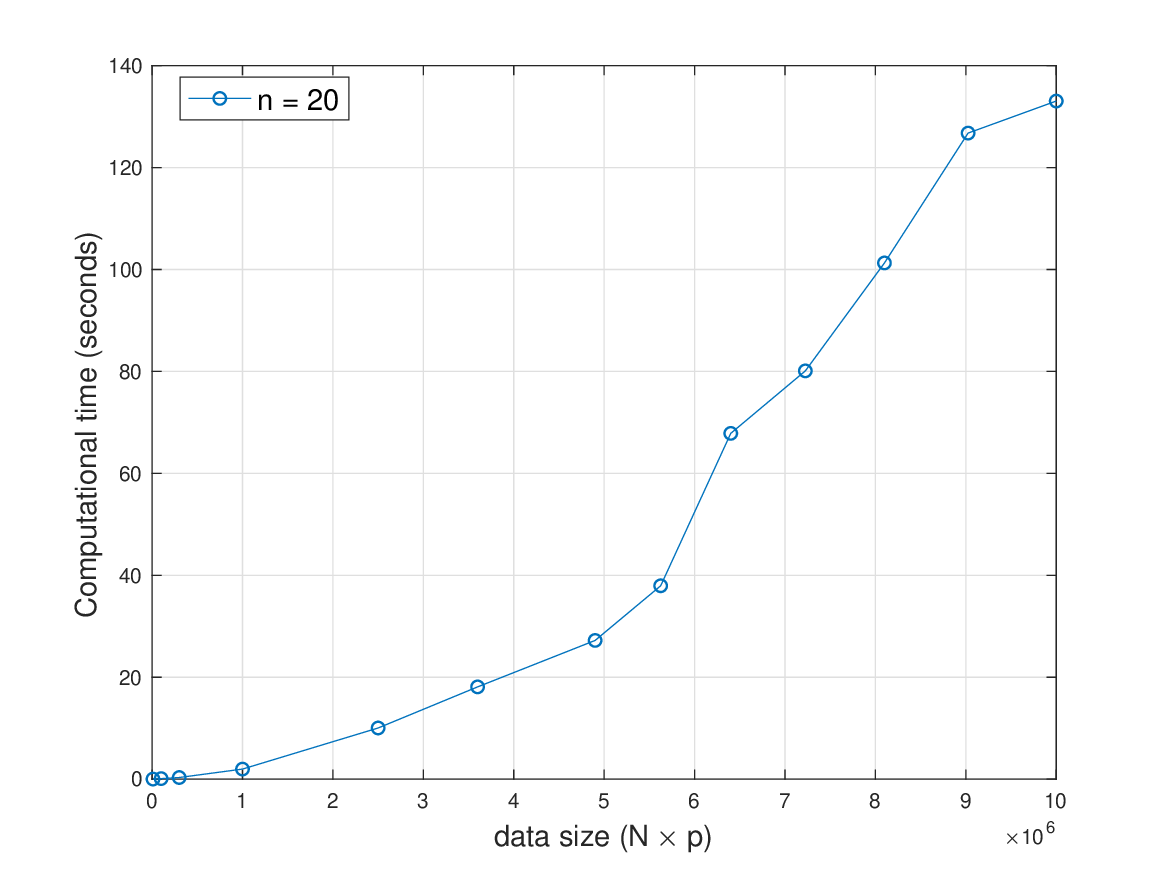}\label{results_computational_4}}}
\caption{The plots for the computational time of KAHM.}
\label{fig_results_computational}
\end{figure}

Fig.~\ref{fig_results_computational} plots the computational time of KAHM in relation to increasing data dimension $p$, subspace dimension $n$, sample size $N$, and data size $N \times p$. The results verify that KAHM remains computationally practical for a wide range of these parameters. Since the data dimension could be very high, simulations include the range of data dimension up to $20000$. The results verify that 1) a higher data dimension does not pose a major computational challenge as it took around 1600 seconds to compute a KAHM from 1000 samples of 20000-dimensional data points, and 2) KAHM is computationally practical as observed from Table~\ref{table_added_18012024} that it took around 133 seconds to process a dataset with $10^7$ entries (i.e. $N \times p = 10^7$).  
\begin{table}
\renewcommand{\arraystretch}{1.3}
\caption{Time required by KAHM to process data of varying sizes.}
\label{table_added_18012024} \centering
\begin{tabular}{c|c}
\hline
data size ($N \times p$) & processing time (in seconds)\\ \hline \hline   
10000 & 0.0193 \\
     100000 & 0.1077 \\
      300000 & 0.3232 \\
     1000000 & 1.9783 \\
     2500000 & 10.0412 \\
     3600000 & 18.1033 \\
     4900000 & 27.2175 \\
     5625000 & 37.9418 \\
     6400000 & 67.8467 \\
     7225000 & 80.1050 \\
     8100000 & 101.2839 \\
     9025000 & 126.7630 \\
    10000000 & 133.0497 \\
\hline \hline
\end{tabular}
\end{table}
\begin{remark}[Dealing with Large Data]\label{rem_large_data}
To deal with the large data when the number of data samples $N$ is large (say $N > 1000$), the data points are suggested to be divided into $S$ number of subsets (where $S$ is given as in (\ref{eq_190220231832})) and for each subset a separate conditionally deep KAHM is built leading to a wide conditionally deep KAHM (Definition~\ref{def_WDKAHM}). Since each data subset can be processed independently, a large number of data samples can be computationally managed using parallel and distributed computing. \end{remark}
\subsection{Summary of Experimental Results}\label{section_experiments_summary}
Following inferences are drawn from the experiments: 
\begin{enumerate}
\item KAHM based classifier improves the existing results on Freiburg groceries dataset indicating a highly competitive performance in modeling and thus classifying 8192-dimensional data points by means of KAHMs. 
\item The proposed fabricated data based differentially private classification reduces considerably the risk of membership inference attack with relatively much smaller loss of accuracy. This is demonstrated through experiments on three different dataset: MNIST dataset, Freiburg groceries dataset, and a real biomedical dataset. 
\item The application potential of the KAHM based differentially private federated learning scheme is verified by the observation that the accuracy-loss due to data being distributed is either marginal or not significantly high. 
\item The computational issues arising from a large number of data samples are addressed automatically by design via splitting the dataset into subsets and processing each subset independently by a branch of the wide conditionally deep KAHM. Simulations verify that KAHM remains computationally practical and a higher data dimension does not pose a major computational challenge.      
\end{enumerate}          
\section{Conclusion}\label{sec_conclusion}
Having learned the representation of data samples in RKHS via solving a kernel regularized least squares problem with a meaningful choice of regularization parameter, KAHM defines a bounded geometric structure in the affine hull of data samples. KAHM and KAHM based models (consisting of series and parallel compositions of KAHMs) induce a distance function that measures the distance of an arbitrary data point from the data samples. Modeling the region of each class in data space through a separate KAHM allows building a classifier. An optimal differentially private noise adding mechanism is applied on training data samples to build a differentially private classifier. The smoothing of noise added samples through a carefully defined transformation (that ensures reducing the geometric modeling error of smoothed samples below of that of original samples) mitigates the accuracy-loss issue of the differentially private classifier. 

The theoretical results obtained in this study are related to the determination of regularization parameter for the kernel regularized least squares problem, boundedness of KAHM, distance functions induced by KAHMs, and smoothing of data for reduction in KAHM modeling error. KAHMs can be applied to a wide range of machine learning problems and this study has considered the differentially private federated learning problem as an application example. The practical significance of the theory is demonstrated through numerous experiments performed to verify the application potential of the KAHMs. A significant feature of our approach is that mathematical analysis has been carried out in a pure deterministic setting without making any statistical assumption. 

Our future work will extend the KAHMs in several directions:
\begin{itemize}
\item A limitation of the current study is that the kernel function has been priori fixed of Gaussian type in defining the KAHM. The effect of different kernel functions and spectral properties of kernel matrix on the resulting geometric structures has not been investigated, which is the part of our future work. 
\item An advantage offered by the proposed approach is that it leverages the post-processing property of differential privacy and thus, unlike stochastic gradient descent based learning algorithms, there is no need of keeping track of the privacy loss incurred by successive iterations of an algorithm. However, in future we will also study the KAHM based iterative differentially private algorithms.
\item KAHM approach will be extended to include a feature extraction procedure for the images, allowing for KAHMs to serve as a competitive alternative to the CNNs and a testing on large-scaled image datasets for a comparison with the existing models. 
\item Finally, the potential of KAHMs as deep generative models will be investigated. 
\end{itemize}

\acks{The research reported in this paper has been supported by the Austrian Research Pro- motion Agency (FFG) COMET-Modul S3AI (Security and Safety for Shared Artificial Intelligence); FFG Grant SMiLe (Secure Machine Learning Applications with Homomorphically Encrypted Data); FFG Grant PRIMAL (Privacy Preserving Machine Learning for Industrial Applications); FFG Sub-Project PETAI (Privacy Secured Explainable and Transferable AI for Healthcare Systems); and the Austrian Ministry for Transport, Innovation and Technology, the Federal Ministry for Digital and Economic Affairs, and the State of Upper Austria in the frame of the SCCH competence center INTEGRATE [(FFG grant no. 892418)] part of the FFG COMET Competence Centers for Excellent Technologies Programme.
}

\appendix
\section{Proof of Theorem~\ref{result_definition_mse_function}}
\label{appendix1}
The proof is split into four parts.
\paragraph{Part 1:} 
Consider
\begin{IEEEeqnarray}{rCl}
(Y)_{:,j} - K_{X} \left(K_{X} + (e+\tau) I_N \right)^{-1} (Y)_{:,j} & = & (I_N + \frac{1}{(e+\tau)} K_{X}  )^{-1} (Y)_{:,j}
 \end{IEEEeqnarray}        
and thus
\begin{IEEEeqnarray}{rCl}
\mathcal{R}_{k,X,Y}(e,\tau) & = & \frac{1}{pN} \sum_{j=1}^p ((Y)_{:,j})^T ( I_N + \frac{1}{(e+\tau)}  K_{X}  )^{-2} (Y)_{:,j}
 \end{IEEEeqnarray}        
Since $ K_{X}$ is a positive definite matrix and $(e+\tau) > 0$,
\begin{IEEEeqnarray}{rCl}
\mu_{min}\left(I_N + \frac{1}{(e+\tau)} K_{X} \right) & > & 1 
\end{IEEEeqnarray} 
where ``$\mu_{min}(\cdot)$'' denotes the minimum eigenvalue. Thus,
\begin{IEEEeqnarray}{rCl}
\mu_{max}\left(\left(I_N + \frac{1}{(e+\tau)} K_{X} \right)^{-2} \right) & < & 1 
\end{IEEEeqnarray} 
where ``$\mu_{max}(\cdot)$'' denotes the maximum eigenvalue. This results in
\begin{IEEEeqnarray}{rCl}
\mathcal{R}_{k,X,Y}(e,\tau) & < & \frac{1}{pN} \sum_{j=1}^p \|(Y)_{:,j} \|^2 \\
& = &  \frac{1}{pN}  \|Y \|_F^2.
\end{IEEEeqnarray}   
It is obvious that $\mathcal{R}_{k,X,Y}(e,\tau)  > 0$, hence (\ref{eq_738818.778}) follows. 
\paragraph{Part 2:}
The derivative of $\mathcal{R}_{k,X,Y}$ w.r.t. $e$ is given as
\begin{IEEEeqnarray}{rCl}
 \frac{\dd \mathcal{R}_{k,X,Y}(e,\tau)}{\dd e}  &= & \frac{2}{pN}\sum_{j=1}^p \left \{(e+\tau) ((Y)_{:,j})^T \left((e+\tau) I_N + K_{X}   \right)^{-2}  (Y)_{:,j} \right. \\
 && \left. -   (e+\tau)^2 ((Y)_{:,j})^T \left( (e+\tau) I_N + K_{X}   \right)^{-3}  (Y)_{:,j} \right \}.
  \end{IEEEeqnarray}  
Consider
\begin{IEEEeqnarray}{rCl}
\lefteqn{(e+\tau)^2 ((Y)_{:,j})^T \left( (e+\tau) I_N + K_{X}  \right)^{-3}  (Y)_{:,j}} \\
 & \leq & (e+\tau)^2 \left \| \left( (e+\tau) I_N + K_{X}   \right)^{-1} \right \|_2 ((Y)_{:,j})^T \left( (e+\tau) I_N + K_{X}   \right)^{-2} (Y)_{:,j} \\
 & = & (e+\tau) \left \| \left( I_N + \frac{1}{(e+\tau)} K_{X}   \right)^{-1} \right \|_2 ((Y)_{:,j})^T \left( (e+\tau) I_N + K_{X}   \right)^{-2} (Y)_{:,j} \\
 & = & (e+\tau) \frac{1}{\sigma_{min}\left( I_N + \frac{1}{(e+\tau)} K_{X} \right)}((Y)_{:,j})^T \left( (e+\tau) I_N + K_{X}   \right)^{-2} (Y)_{:,j}
\end{IEEEeqnarray}
where ``$\sigma_{min}(\cdot)$'' denotes the minimum singular value. Observing that $(e+\tau) > 0$ and $K_{X}$ is a positive definite matrix, we have
\begin{IEEEeqnarray}{rCl}
\sigma_{min}\left( I_N + \frac{1}{(e+\tau)} K_{X} \right) & = & 1 + \sigma_{min}\left(\frac{1}{(e+\tau) } K_{X} \right) \\
 & > & 1.
\end{IEEEeqnarray}
Thus,
 \begin{IEEEeqnarray}{rCl} 
\nonumber \lefteqn{(e+\tau)^2 ((Y)_{:,j})^T \left( (e+\tau) I_N + K_{X}  \right)^{-3} (Y)_{:,j}}\\
  & < &(e+\tau) ((Y)_{:,j})^T \left((e+\tau) I_N + K_{X}   \right)^{-2} (Y)_{:,j} 
\end{IEEEeqnarray}      
resulting in   
\begin{IEEEeqnarray}{rCl}
 \frac{\dd \mathcal{R}_{k,X,Y}(e)}{\dd e }  & >  &0. \label{eq_738818.7858}
  \end{IEEEeqnarray}
Since $(e+\tau) > 0$ and $K_{X}$ is a positive definite matrix,
\begin{IEEEeqnarray}{rCl}
(e+\tau)^2 ((Y)_{:,j})^T \left( (e+\tau) I_N + K_{X}    \right)^{-3}   (Y)_{:,j} & > & 0,
\end{IEEEeqnarray}
and therefore
\begin{IEEEeqnarray}{rCl}
 \frac{\dd \mathcal{R}_{k,X,Y}(e,\tau)}{\dd e}  & < & \frac{2}{pN} (e+\tau) \sum_{j=1}^p  ((Y)_{:,j})^T  \left((e+\tau) I_N + K_{X}   \right)^{-2}   (Y)_{:,j} \\
 & = & \frac{2}{(e+\tau)}  \mathcal{R}_{k,X,Y}(e)\\
 & < &   \frac{2}{(e+\tau)} \frac{1}{pN}  \|Y \|_F^2. \label{eq_738818.7862}
  \end{IEEEeqnarray}
Inequalities (\ref{eq_738818.7858}) and (\ref{eq_738818.7862}) lead to (\ref{eq_738818.7872}). 
\paragraph{Part 3:} 
For a given $\tau \in \mathbf{R}_{+}$, introduce $m_{\tau}(e)  =  \mathcal{R}_{k,X,Y}(e,\tau) - e$, and observe that $m_{\tau}(0) > 0$ and $m_{\tau}(\frac{1}{pN}\|Y \|_F^2) < 0$. By the intermediate value theorem, there is a $\hat{e} \in (0, \frac{1}{pN}\|Y \|_F^2  )$ such that $m_{\tau}(\hat{e}) = 0$, i.e., $\hat{e}  =\mathcal{R}_{k,X,Y}(\hat{e},\tau)$. Thus, $\hat{e}$ is a fixed point of $\mathcal{R}_{k,X,Y}(e,\tau)$.  
\paragraph{Part 4:}
It follows from (\ref{eq_738819.4989}) and (\ref{eq_738818.7872}) that
 \begin{IEEEeqnarray}{rCl} 
  \frac{\dd \mathcal{R}_{k,X,Y}(e,\tau)}{\dd e} & \in &  (0, 1 ).
\end{IEEEeqnarray} 
That is, there exists a constant $c$ such that  
\begin{IEEEeqnarray}{rCCCCCl}
\label{eq_738570.5074}0 & < & \frac{\dd \mathcal{R}_{k,X,Y}(e|_{it},\tau)}{\dd e}  & \leq & c & < 1,\; \forall it \in \{0,1,2,\cdots \}.
  \end{IEEEeqnarray}  
Let $\hat{e}$ be a fixed point of $\mathcal{R}_{k,X,Y}(e,\tau)$. Now, consider
 \begin{IEEEeqnarray}{rCCCl} 
 \left | e |_{it} -  \hat{e} \right | & = &  \left | \mathcal{R}_{k,X,Y}(e|_{it-1},\tau) - \mathcal{R}_{k,X,Y}(\hat{e},\tau) \right | & \leq & c \left | e |_{it-1} -  \hat{e} \right | \\
 &&& \leq & c^2 \left | e|_{it-2} -  \hat{e} \right | \\
\nonumber &&&& \vdots \\
  &&& \leq & c^{it} \left | e|_0 - \hat{e} \right |,
   \end{IEEEeqnarray}  
 that leads to 
 \begin{IEEEeqnarray}{rCCCl} 
\lim_{it \to \infty} \left | e|_{it} -  \hat{e} \right | & \leq & \lim_{it \to \infty} c^{it} \left | e |_0 -  \hat{e} \right | & = & 0.
    \end{IEEEeqnarray}  
Hence the iterations (\ref{eq_738819.5207})-(\ref{eq_738819.5208}) converge to a fixed point of $\mathcal{R}_{k,X,Y}(e,\tau)$. The uniqueness of the fixed point can be seen via assuming by contradiction that there exists another fixed point, say $\tilde{e}$. Now consider
 \begin{IEEEeqnarray}{rCCCCCCl} 
 \left | \tilde{e} -  \hat{e} \right | & = & \left |\mathcal{R}_{k,X,Y}(\tilde{e},\tau) -  \mathcal{R}_{k,X,Y}(\hat{e},\tau) \right | & \leq & c \left | \tilde{e} -  \hat{e} \right | & < & \left | \tilde{e} -  \hat{e} \right |. 
    \end{IEEEeqnarray}  
This implies that $\tilde{e}  =  \hat{e} $. Hence, the result follows.  

\section{Proof of Theorem~\ref{result_kahm_bounded_function}}
\label{appendix1_5}
Define a diagonal matrix $D_y$ as
 \begin{IEEEeqnarray}{rCl}
 D_y & = & \textrm{diag}\left(k_{\theta}(Py,Py^1),\cdots,k_{\theta}(Py,Py^N)\right),\mbox{ and}\\
\overline{k}_y  & = & \max_{i \in \{1,2,\cdots,N\}}\;k_{\theta}(Py,Py^i).
   \end{IEEEeqnarray}   
Further define a vector $G_y$ as
 \begin{IEEEeqnarray}{rCl}
 G_y & = & \left[\begin{IEEEeqnarraybox*}[][c]{,c/c/c,} k_{\theta}(Py,Py^1) & \cdots & k_{\theta}(Py,Py^N)\end{IEEEeqnarraybox*} \right]^T.
  \end{IEEEeqnarray}         
It is obvious that $D_y^{-1} - (\overline{k}_y )^{-1}I$ is symmetric positive semi-definite, i.e.,
 \begin{IEEEeqnarray}{rCl}
\label{eq_291220221500}D_y^{-1} - (\overline{k}_y )^{-1}I_N & \succeq & 0.
   \end{IEEEeqnarray}         
Since $K_{YP^T}$ is symmetric positive definite and $\lambda^* > 0$,      
 \begin{IEEEeqnarray}{rCl}
\label{eq_291220221501}(K_{YP^T} + \lambda^* I_N)^{-1} & \succ & 0.
   \end{IEEEeqnarray}         
It follows from (\ref{eq_291220221500}) and (\ref{eq_291220221501}) that
 \begin{IEEEeqnarray}{rCl}
 \left(D_y^{-1} - (\overline{k}_y )^{-1}I_N \right) (K_{YP^T} + \lambda^* I_N)^{-1}    & \succeq & 0, \;\mbox{i.e.}\\
 D_y^{-1}(K_{YP^T} + \lambda^* I_N)^{-1} - (\overline{k}_y )^{-1} (K_{YP^T} + \lambda^* I_N)^{-1}   & \succeq & 0, \;\mbox{i.e.}\\
G_y^T \left( D_y^{-1}(K_{YP^T} + \lambda^* I_N)^{-1} - (\overline{k}_y )^{-1} (K_{YP^T} + \lambda^* I_N)^{-1}   \right) G_y & \geq & 0.
   \end{IEEEeqnarray}      
Thus,
 \begin{IEEEeqnarray}{rCl}
G_y^T  D_y^{-1}(K_{YP^T} + \lambda^* I_N)^{-1}  G_y & \geq & (\overline{k}_y )^{-1} G_y^T (K_{YP^T} + \lambda^* I_N)^{-1} G_y.
   \end{IEEEeqnarray}     
Also,
 \begin{IEEEeqnarray}{rCl}
\sum_{i=1}^Nh^i_{k_{\theta},YP^T,\lambda^*}(Py)  & = & (\mathbf{1}_N)^T \left(K_{YP^T} + \lambda I_N \right)^{-1} G_y \\
& = & G_y^T  D_y^{-1} \left(K_{YP^T} + \lambda I_N \right)^{-1} G_y \\
\label{eq_050120231126}& \geq & (\overline{k}_y )^{-1} G_y^T (K_{YP^T} + \lambda^* I_N)^{-1} G_y \\
& \geq & (\overline{k}_y )^{-1} \mu_{min}\left( \left(K_{YP^T} + \lambda^* I_N \right)^{-1}  \right) \| G_y\|^2.
   \end{IEEEeqnarray}   
As $\left(K_{YP^T} + \lambda^* I_N \right)^{-1} $ is real symmetric positive definite,
 \begin{IEEEeqnarray}{rCl}
\label{eq_300120231514}\sum_{i=1}^Nh^i_{k_{\theta},YP^T,\lambda^*}(Py)  
& > & 0.
   \end{IEEEeqnarray} 
Consider
 \begin{IEEEeqnarray}{rCl}
 \frac{\left \|  \left[\begin{IEEEeqnarraybox*}[][c]{,c/c/c,} h^1_{k_{\theta},YP^T,\lambda^*}(Py) & \cdots & h^N_{k_{\theta},YP^T,\lambda^*}(Py) \end{IEEEeqnarraybox*} \right]^T  \right \|}{\left | \sum_{i=1}^Nh^i_{k_{\theta},YP^T,\lambda^*}(Py) \right |} & = & \frac{ \left \| \left(K_{YP^T} + \lambda^* I_N \right)^{-1} G_y \right \|}{\sum_{i=1}^Nh^i_{k_{\theta},YP^T,\lambda^*}(Py)}\\
 &\leq& \frac{\overline{k}_y }{\| G_y\|} \frac{\left \| \left(K_{YP^T} + \lambda^* I_N \right)^{-1}  \right \|_2}{\mu_{min}\left( \left(K_{YP^T} + \lambda^* I_N \right)^{-1}  \right)}.
    \end{IEEEeqnarray} 
 Since $\left(K_{YP^T} + \lambda^* I_N \right)^{-1} $ is real symmetric positive definite,      
 \begin{IEEEeqnarray}{rCl}
\frac{\overline{k}_y }{\| G_y\|} \frac{\left \| \left(K_{YP^T} + \lambda^* I_N \right)^{-1}  \right \|_2}{\mu_{min}\left( \left(K_{YP^T} + \lambda^* I_N \right)^{-1}  \right)} & = & \frac{\overline{k}_y }{\| G_y\|} \frac{\mu_{max}\left ( \left(K_{YP^T} + \lambda^* I_N \right)^{-1}  \right )}{\mu_{min}\left( \left(K_{YP^T} + \lambda^* I_N \right)^{-1}  \right)} \\
 & = & \frac{\overline{k}_y }{\| G_y\|} \frac{\lambda^* + \mu_{max}(K_{YP^T} )}{\lambda^* + \mu_{min}(K_{YP^T} )} \\
 & = & \frac{  \max_{i \in \{1,2,\cdots,N\}}\;k_{\theta}(Py,Py^i) }{\sqrt{\sum_{i=1}^N |k_{\theta}(Py,Py^i)|^2}} \frac{\lambda^* + \mu_{max}(K_{YP^T} )}{\lambda^* + \mu_{min}(K_{YP^T} )} \\
 & < & \frac{\lambda^* + \mu_{max}(K_{YP^T} )}{\lambda^* + \mu_{min}(K_{YP^T} )}.
   \end{IEEEeqnarray} 
Thus
 \begin{IEEEeqnarray}{rCl}
\label{eq_100120231427} \frac{\left \|  \left[\begin{IEEEeqnarraybox*}[][c]{,c/c/c,} h^1_{k_{\theta},YP^T,\lambda^*}(Py) & \cdots & h^N_{k_{\theta},YP^T,\lambda^*}(Py) \end{IEEEeqnarraybox*} \right]^T  \right \|}{\left | \sum_{i=1}^Nh^i_{k_{\theta},YP^T,\lambda^*}(Py) \right |} & < &   \frac{\lambda^* + \mu_{max}(K_{YP^T} )}{\lambda^* + \mu_{min}(K_{YP^T} )}.
   \end{IEEEeqnarray} 
 Since $K_{YP^T}$ is positive definite (i.e. $\mu_{min}(K_{YP^T}) > 0$) and $\mathrm{tr}(K_{YP^T} ) = N$ (i.e. $\mu_{max}(K_{YP^T} ) < N$), we have 
 \begin{IEEEeqnarray}{rCl}
\label{eq_100120231357}\frac{\lambda^* + \mu_{max}(K_{YP^T} )}{\lambda^* + \mu_{min}(K_{YP^T} )}  & < &  \frac{\lambda^* + N}{\lambda^* }.
   \end{IEEEeqnarray}
Using (\ref{eq_010120231034}) with the observation that $\hat{e} > 0$, we have 
 \begin{IEEEeqnarray}{rCl}
\label{eq_080120231344}\lambda^* & > &  \frac{2}{pN}\|Y \|^2_F,\mbox{ leading to} 
   \end{IEEEeqnarray}
    \begin{IEEEeqnarray}{rCl}
\label{eq_100120231358}\frac{\lambda^* + \mu_{max}(K_{YP^T} )}{\lambda^* + \mu_{min}(K_{YP^T} )} & < &  1 + \frac{p N^2}{2 \|Y\|_F^2}.
   \end{IEEEeqnarray}
It is observed from (\ref{eq_301220221131}) that
 \begin{IEEEeqnarray}{rCl}
 \mathcal{A}_{Y,n}(y) & = & \frac{1}{\sum_{i=1}^Nh^i_{k_{\theta},YP^T,\lambda^*}(Py)} \left[\begin{IEEEeqnarraybox*}[][c]{,c/c/c,} y^1 & \cdots & y^N\end{IEEEeqnarraybox*} \right] \left[\begin{IEEEeqnarraybox*}[][c]{,c/c/c,} h^1_{k_{\theta},YP^T,\lambda^*}(Py) & \cdots & h^N_{k_{\theta},YP^T,\lambda^*}(Py) \end{IEEEeqnarraybox*} \right]^T. \IEEEeqnarraynumspace
   \end{IEEEeqnarray}  
Thus,
 \begin{IEEEeqnarray}{rCl}
\label{eq_100120231359} \|  \mathcal{A}_{Y,n}(y)\| & \leq & \left \| \left[\begin{IEEEeqnarraybox*}[][c]{,c/c/c,} y^1 & \cdots & y^N \end{IEEEeqnarraybox*} \right]  \right \|_2 \frac{\left \|  \left[\begin{IEEEeqnarraybox*}[][c]{,c/c/c,} h^1_{k_{\theta},YP^T,\lambda^*}(Py) & \cdots & h^N_{k_{\theta},YP^T,\lambda^*}(Py) \end{IEEEeqnarraybox*} \right]^T  \right \|}{\left | \sum_{i=1}^Nh^i_{k_{\theta},YP^T,\lambda^*}(Py) \right | }. 
   \end{IEEEeqnarray}  
Using (\ref{eq_100120231427}) and (\ref{eq_100120231358}) in (\ref{eq_100120231359}) leads to 
 \begin{IEEEeqnarray}{rCCCl}
 \|  \mathcal{A}_{Y,n}(y)\| & < & \left \| Y  \right \|_2 \frac{\lambda^* + \mu_{max}(K_{YP^T} )}{\lambda^* + \mu_{min}(K_{YP^T} )} & < &  \left \| Y  \right \|_2 \left( 1 + \frac{p N^2}{2 \|Y\|_F^2} \right).
   \end{IEEEeqnarray} 
Hence, (\ref{eq_100120231400}) follows. 
\section{Proof of Theorem~\ref{result_ratio_distances}}
\label{appendix2}      
It is observed from (\ref{eq_301220221131}) that
\begin{IEEEeqnarray}{rCl}
\nonumber \lefteqn{ y - \mathcal{A}_{Y,n}(y)}\\
 & = & \frac{1}{\sum_{i=1}^Nh^i_{k_{\theta},YP^T,\lambda^*}(Py)} \left[\begin{IEEEeqnarraybox*}[][c]{,c/c/c,} y-y^1 & \cdots & y-y^N\end{IEEEeqnarraybox*} \right] \left[\begin{IEEEeqnarraybox*}[][c]{,c/c/c,} h^1_{k_{\theta},YP^T,\lambda^*}(Py) & \cdots & h^N_{k_{\theta},YP^T,\lambda^*}(Py) \end{IEEEeqnarraybox*} \right]^T. \IEEEeqnarraynumspace
    \end{IEEEeqnarray}  
Thus,
 \begin{IEEEeqnarray}{rCl}
\|  y - \mathcal{A}_{Y,n}(y) \| & \leq & \left \| \left[\begin{IEEEeqnarraybox*}[][c]{,c/c/c,} y-y^1 & \cdots & y-y^N\end{IEEEeqnarraybox*} \right]  \right \|_2 \frac{\left \|  \left[\begin{IEEEeqnarraybox*}[][c]{,c/c/c,} h^1_{k_{\theta},YP^T,\lambda^*}(Py) & \cdots & h^N_{k_{\theta},YP^T,\lambda^*}(Py) \end{IEEEeqnarraybox*} \right]^T  \right \|}{\left | \sum_{i=1}^Nh^i_{k_{\theta},YP^T,\lambda^*}(Py) \right | }. \IEEEeqnarraynumspace
   \end{IEEEeqnarray}       
That is,
 \begin{IEEEeqnarray}{rCl}
 \frac{ \Gamma_{\mathcal{A}_{Y,n}}(y)}{\left \| \left[\begin{IEEEeqnarraybox*}[][c]{,c/c/c,} y-y^1 & \cdots & y-y^N\end{IEEEeqnarraybox*} \right]  \right \|_2} & \leq &  \frac{\left \|  \left[\begin{IEEEeqnarraybox*}[][c]{,c/c/c,} h^1_{k_{\theta},YP^T,\lambda^*}(Py) & \cdots & h^N_{k_{\theta},YP^T,\lambda^*}(Py) \end{IEEEeqnarraybox*} \right]^T  \right \|}{\left | \sum_{i=1}^Nh^i_{k_{\theta},YP^T,\lambda^*}(Py) \right | }.
   \end{IEEEeqnarray}         
Using (\ref{eq_100120231427}) and (\ref{eq_100120231358}) leads to    
 \begin{IEEEeqnarray}{rCCCl}
 \frac{ \Gamma_{\mathcal{A}_{Y,n}}(y)}{\left \| \left[\begin{IEEEeqnarraybox*}[][c]{,c/c/c,} y-y^1 & \cdots & y-y^N\end{IEEEeqnarraybox*} \right]  \right \|_2} & < & \frac{\lambda^* + \mu_{max}(K_{YP^T} )}{\lambda^* + \mu_{min}(K_{YP^T} )} & < & 1 + \frac{p N^2}{2 \|Y\|_F^2}.
   \end{IEEEeqnarray}        
Hence, (\ref{eq_100120231432}) follows.

\section{Proof of Theorem~\ref{result_ratio_distances_cond_deep_KAHM}} \label{appendix3}
It is observed from the definition of $\mathcal{D}_{Y,n,L}$ (i.e. (\ref{eq_220120231153}-\ref{eq_220120231155})) that 
\begin{IEEEeqnarray}{rCl}
\label{eq_220120231201}\Gamma_{\mathcal{D}_{Y,n,L}}(y) & \leq &  \Gamma_{\mathcal{A}_{Y,n}}(y).
  \end{IEEEeqnarray} 
Using (\ref{eq_220120231201}) in (\ref{eq_100120231432}) leads to (\ref{eq_220120231130}).

\section{Proof of Theorem~\ref{result_ratio_distances_wide_cond_deep_KAHM}} \label{appendix4}
\begin{IEEEeqnarray}{rCl}
 \Gamma_{\mathcal{W}_{Y,n,L,S}}(y) & = & \min_{s \in \{1,2,\cdots,S \}} \; \Gamma_{\mathcal{D}_{Y_s,n,L}}(y) \\
 & \leq & \min_{s \in \{1,2,\cdots,S \}} \;  \Gamma_{\mathcal{A}_{Y_s,n}}(y)   \\
 & < & \min_{s \in \{1,2,\cdots,S \}} \; \left\{ \left( 1 + \frac{p N_s^2}{2 \|Y_s\|_F^2} \right) \left \|\left[\begin{IEEEeqnarraybox*}[][c]{,c/c/c,} y - y^{1,s} & \cdots & y - y^{N_s,s} \end{IEEEeqnarraybox*} \right] \right \|_2 \right \} \\
 & \leq & \min_{s \in \{1,2,\cdots,S \}} \; \left\{ \left( 1 + \frac{p N_s^2}{2 \|Y_s\|_F^2} \right) \left \|\left[\begin{IEEEeqnarraybox*}[][c]{,c/c/c,} y - y^{1,s} & \cdots & y - y^{N_s,s} \end{IEEEeqnarraybox*} \right] \right \|_F \right \}.
   \end{IEEEeqnarray}    
Since $\{y^{1,s}, \cdots, y^{N_s,s} \} \subset \{y^1,\cdots,y^N \}$,
 \begin{IEEEeqnarray}{rCl} 
 \left  \|\left[\begin{IEEEeqnarraybox*}[][c]{,c/c/c,} y - y^{1,s} & \cdots & y - y^{N_s,s} \end{IEEEeqnarraybox*} \right] \right \|_F & < &  \left \|\left[\begin{IEEEeqnarraybox*}[][c]{,c/c/c,} y - y^1 & \cdots & y - y^N \end{IEEEeqnarraybox*} \right] \right \|_F,
  \end{IEEEeqnarray} 
and thus
 \begin{IEEEeqnarray}{rCl} 
\Gamma_{\mathcal{W}_{Y,n,L,S}}(y) & < & \left \|\left[\begin{IEEEeqnarraybox*}[][c]{,c/c/c,} y - y^1 & \cdots & y - y^N \end{IEEEeqnarraybox*} \right] \right \|_F \times \min_{s \in \{1,2,\cdots,S \}} \;  \left( 1 + \frac{p N_s^2}{2 \|Y_s\|_F^2} \right) 
  \end{IEEEeqnarray} 
leading to (\ref{eq_220120231835}).

\section{Proof of Theorem~\ref{result_transformation}}
\label{appendix5}
Define a $N \times N$ matrix $H_m$ and a $p \times N$ matrix $O_m$ as
 \begin{IEEEeqnarray}{rCl}
 H_m & = & \left[\begin{IEEEeqnarraybox*}[][c]{,c/c/c,} h_{k_{\theta_m},\hat{Y}_mP^T_m,\lambda^*_m}^1(P_m\hat{y}^{1,m})  & \cdots & h_{k_{\theta_m},\hat{Y}_mP^T_m,\lambda^*_m}^1(P_m\hat{y}^{N,m}) \\ \vdots & & \vdots \\ h_{k_{\theta_m},\hat{Y}_mP^T_m,\lambda^*_m}^N(P_m\hat{y}^{1,m}) & \cdots & h_{k_{\theta_m},\hat{Y}_mP^T_m,\lambda^*_m}^N(P_m\hat{y}^{N,m})  \end{IEEEeqnarraybox*} \right],\\
 O_m& = & \left[\begin{IEEEeqnarraybox*}[][c]{,c/c/c,} \sum_{j=1}^Nh_{k_{\theta_m},\hat{Y}_mP^T_m,\lambda^*_m}^j(P_m\hat{y}^{1,m})  & \cdots & \sum_{j=1}^Nh_{k_{\theta_m},\hat{Y}_mP^T_m,\lambda^*_m}^j(P_m\hat{y}^{N,m}) \\ \vdots & & \vdots \\ \sum_{j=1}^Nh_{k_{\theta_m},\hat{Y}_mP^T_m,\lambda^*_m}^j(P_m\hat{y}^{1,m}) & \cdots & \sum_{j=1}^Nh_{k_{\theta_m},\hat{Y}_mP^T_m,\lambda^*_m}^j(P_m\hat{y}^{N,m})  \end{IEEEeqnarraybox*} \right]_{p \times N}.
     \end{IEEEeqnarray} 
It can be seen using (\ref{eq_250220231734}) that
\begin{IEEEeqnarray}{rCl}
H_m & = & (K_{\hat{Y}_mP^T_m} + \lambda^*_m I_N)^{-1} K_{\hat{Y}_mP^T_m},\mbox{ i.e.}\\
\label{eq_290120231238}I_N - H_m & = & (I_N + \frac{1}{\lambda^*_m}K_{\hat{Y}_mP^T_m})^{-1},\mbox{ i.e.}\\
\| I_N - H_m \|_2 & = & \frac{1}{\sigma_{min}(I_N + \frac{1}{\lambda^*_m}K_{\hat{Y}_mP^T_m})}.
 \end{IEEEeqnarray}
As $\lambda^*_m > 0$ and $K_{\hat{Y}_mP^T_m}$ is a positive definite matrix, 
\begin{IEEEeqnarray}{rCCCl}
\label{eq_290120231750} \| I_N - H_m \|_2 & = & \frac{1}{1 + \sigma_{min}(\frac{1}{\lambda^*_m} K_{\hat{Y}_mP^T_m})} & < & 1.
 \end{IEEEeqnarray}
As the r.h.s. of (\ref{eq_290120231238}) is a positive definite matrix, we have 
\begin{IEEEeqnarray}{rCl}
 \mu_{min}(I_N - H_m) & > & 0,\mbox{ i.e.}\\
 \label{eq_290120231243}\mu_{max}(H_m) & < & 1.
 \end{IEEEeqnarray} 
As $H_m$ is a real symmetric matrix, it follows immediately from (\ref{eq_290120231243}) that $\sigma_{max}(H_m) < 1$, and thus
\begin{IEEEeqnarray}{rCl}
\label{eq_290120231747}\|  H_m \|_2 & < & 1,\mbox{ i.e.}\\
\|H_m \|_1 & < & \sqrt{N},\mbox{ i.e.}\\
\max_{i \in \{1,2,\cdots,N \} } \sum_{j=1}^N | h_{k_{\theta_m},\hat{Y}_mP^T_m,\lambda^*_m}^j(P_m\hat{y}^{i,m}) | & < & \sqrt{N},\mbox{ thus}\\
\sum_{j=1}^N  h_{k_{\theta_m},\hat{Y}_mP^T_m,\lambda^*_m}^j(P_m\hat{y}^{i,m})  & < & \sqrt{N},\;\forall i \in \{1,2,\cdots,N \}.
 \end{IEEEeqnarray} 
In view of (\ref{eq_300120231514}), we have
\begin{IEEEeqnarray}{rCCCl}
\label{eq_300120231553}0 &<&\sum_{j=1}^N  h_{k_{\theta_m},\hat{Y}_mP^T_m,\lambda^*_m}^j(P_m\hat{y}^{i,m})  & < & \sqrt{N},\;\forall i \in \{1,2,\cdots,N \}.
 \end{IEEEeqnarray} 
It follows from (\ref{eq_301220221131}) that
\begin{IEEEeqnarray}{rCl}
 O_m \circ \left[\begin{IEEEeqnarraybox*}[][c]{,c/c/c,} \mathcal{A}_{\hat{Y}_{m},n}(\hat{y}^{1,m})  & \cdots & \mathcal{A}_{\hat{Y}_{m},n}(\hat{y}^{N,m})  \end{IEEEeqnarraybox*} \right]  & = &  \left[\begin{IEEEeqnarraybox*}[][c]{,c/c/c,} \hat{y}^{1,m}  & \cdots & \hat{y}^{N,m}  \end{IEEEeqnarraybox*} \right]   H_m, \mbox{ i.e.}\\
\label{eq_290120231545}\hat{Y}_{m+1}^T & = & \hat{Y}_m^T H_m, \mbox{ i.e.}\\
\label{eq_290120231708} \hat{Y}_m^T -  \hat{Y}_{m+1}^T & = & \hat{Y}_m^T (I_N - H_m).
 \end{IEEEeqnarray} 
Using (\ref{eq_290120231708}) for $m = M-1$, we have
\begin{IEEEeqnarray}{rCl}
\label{eq_290120231727}\hat{Y}_{M-1}^T -  \hat{Y}_M^T & = & \hat{Y}_{M-1}^T (I_N - H_{M-1}).
 \end{IEEEeqnarray}  
Using (\ref{eq_290120231545}) recursively from $m = 0$ to $m = M-2$, we have
\begin{IEEEeqnarray}{rCl}
\label{eq_290120231728}\hat{Y}_{M-1}^T & = &  \hat{Y}_0^T H_0H_1 \cdots H_{M-2}.
 \end{IEEEeqnarray} 
Combining (\ref{eq_290120231727}) and (\ref{eq_290120231728}) leads to
\begin{IEEEeqnarray}{rCl}
\hat{Y}_{M-1}^T -  \hat{Y}_M^T & = &  \hat{Y}_0^T H_0H_1 \cdots H_{M-2} (I_N - H_{M-1}),\mbox{ i.e.}\\
\hat{Y}_{M-1} -  \hat{Y}_M & = & (I_N - H_{M-1})H_{M-2} \cdots H_1 H_0 \hat{Y}_0,\mbox{ i.e.}\\
\| \hat{Y}_{M-1} -  \hat{Y}_M \|_F & \leq & \|  (I_N - H_{M-1})H_{M-2} \cdots H_1 H_0 \|_2 \|  \hat{Y}_0\|_F, \mbox{ i.e.}\\
\label{eq_290120231754}\| \hat{Y}_{M-1} -  \hat{Y}_M \|_F & \leq & \| I_N - H_{M-1} \|_2 \| H_{M-2} \|_2 \cdots  \|H_1 \|_2  \| H_0 \|_2    \| \hat{Y}_0 \|_F.
 \end{IEEEeqnarray}  
Define
\begin{IEEEeqnarray}{rCl}
\beta & = & \max\; \left(   \| I_N - H_{M-1} \|_2, \| H_{M-2} \|_2, \cdots,  \|H_1 \|_2,  \| H_0 \|_2 \right).
 \end{IEEEeqnarray} 
Since $\|  H_m \|_2  <  1$ (i.e. (\ref{eq_290120231747})) and also $\| I_N - H_m \|_2 < 1$ (i.e. (\ref{eq_290120231750})), we must have
\begin{IEEEeqnarray}{rCCCl}
\label{eq_300120231935} 0&<&\beta & < & 1.
 \end{IEEEeqnarray} 
It follows from (\ref{eq_290120231754}) that
\begin{IEEEeqnarray}{rCl}
\| \hat{Y}_{M-1} -  \hat{Y}_M \|_F & \leq &  (\beta)^M \| \hat{Y}_0 \|_F.
 \end{IEEEeqnarray} 
Considering that $\hat{Y}_0 = Y^+_{\epsilon}$ and $\hat{y}^{i,m}$ is the $i-$th column of $\hat{Y}_m^T$, we have  
\begin{IEEEeqnarray}{rCl}
\label{eq_300120231735}\| \hat{y}^{i,M-1} -  \hat{y}^{i,M} \| & \leq &  (\beta)^M \| Y^+_{\epsilon} \|_F.
 \end{IEEEeqnarray}  
Consider
\begin{IEEEeqnarray}{rCl}
\nonumber \lefteqn{\hat{y}^{i,M} - \mathcal{A}_{\hat{Y}_{M-1},n}\left(\hat{y}^{i,M-1} \right) }\\
& = & \left( \sum_{j=1}^Nh_{k_{\theta_{M-1}},\hat{Y}_{M-1}P^T_{M-1},\lambda^*_{M-1}}^j(P_{M-1}\hat{y}^{i,M-1}) - 1 \right) \mathcal{A}_{\hat{Y}_{M-1},n}\left(\hat{y}^{i,M-1} \right),\mbox{ thus}
 \end{IEEEeqnarray}  
\begin{IEEEeqnarray}{rCl}
\nonumber \lefteqn{\| \hat{y}^{i,M} - \mathcal{A}_{\hat{Y}_{M-1},n}\left(\hat{y}^{i,M-1} \right) \|}\\
\label{eq_300120231549} &= & \left | \sum_{j=1}^Nh_{k_{\theta_{M-1}},\hat{Y}_{M-1}P^T_{M-1},\lambda^*_{M-1}}^j(P_{M-1}\hat{y}^{i,M-1}) - 1 \right | \|  \mathcal{A}_{\hat{Y}_{M-1},n}\left(\hat{y}^{i,M-1} \right) \|.
 \end{IEEEeqnarray}  
Using (\ref{eq_300120231553}) in (\ref{eq_300120231549}), we have
\begin{IEEEeqnarray}{rCl}
\| \hat{y}^{i,M} - \mathcal{A}_{\hat{Y}_{M-1},n}\left(\hat{y}^{i,M-1} \right) \| & < & (\sqrt{N}-1) \|  \mathcal{A}_{\hat{Y}_{M-1},n}\left(\hat{y}^{i,M-1} \right) \|.
 \end{IEEEeqnarray}  
The inequality (\ref{eq_100120231400}) leads to
\begin{IEEEeqnarray}{rCl}
\| \hat{y}^{i,M} - \mathcal{A}_{\hat{Y}_{M-1},n}\left(\hat{y}^{i,M-1} \right) \| & < & (\sqrt{N}-1) \|\hat{Y}_{M-1}  \|_2 \frac{\lambda^*_{M-1} + \mu_{max}(K_{\hat{Y}_{M-1}P^T_{M-1}} )}{\lambda^*_{M-1} + \mu_{min}(K_{\hat{Y}_{M-1}P^T_{M-1}} )} \\
& < &  (\sqrt{N}-1) \|\hat{Y}_{M-1}  \|_F \frac{\lambda^*_{M-1} + \mu_{max}(K_{\hat{Y}_{M-1}P^T_{M-1}} )}{\lambda^*_{M-1} + \mu_{min}(K_{\hat{Y}_{M-1}P^T_{M-1}} )}.
 \end{IEEEeqnarray}  
It follows from (\ref{eq_290120231728}) that
\begin{IEEEeqnarray}{rCl} 
\|\hat{Y}_{M-1} \|_F & \leq & \|H_{M-2} \|_2 \cdots \|H_1\|_2 \| H_0\|_2 \| \hat{Y}_0\|_F,\mbox{ i.e.}\\
\|\hat{Y}_{M-1} \|_F & \leq & (\beta)^{M-1} \| Y^+_{\epsilon} \|_F.
 \end{IEEEeqnarray}    
Thus
\begin{IEEEeqnarray}{rCl}
\label{eq_300120231736}\| \hat{y}^{i,M} - \mathcal{A}_{\hat{Y}_{M-1},n}\left(\hat{y}^{i,M-1} \right) \| & < &  (\sqrt{N}-1) (\beta)^{M-1}  \frac{\lambda^*_{M-1} + \mu_{max}(K_{\hat{Y}_{M-1}P^T_{M-1}} )}{\lambda^*_{M-1} + \mu_{min}(K_{\hat{Y}_{M-1}P^T_{M-1}} )} \| Y^+_{\epsilon} \|_F.  \IEEEeqnarraynumspace
 \end{IEEEeqnarray}  
Consider 
\begin{IEEEeqnarray}{rCl}
\label{eq_300120231737} \| \hat{y}^{i,M-1} - \mathcal{A}_{\hat{Y}_{M-1},n}(\hat{y}^{i,M-1}) \| & \leq & \| \hat{y}^{i,M-1} -  \hat{y}^{i,M} \| +  \|  \hat{y}^{i,M} - \mathcal{A}_{\hat{Y}_{M-1},n}\left(\hat{y}^{i,M-1} \right) \|. 
\end{IEEEeqnarray}  
Using (\ref{eq_300120231735}) and (\ref{eq_300120231736}) in (\ref{eq_300120231737}), we finally obtain
\begin{IEEEeqnarray}{rCl}
\nonumber \lefteqn{\| \hat{y}^{i,M-1} - \mathcal{A}_{\hat{Y}_{M-1},n}(\hat{y}^{i,M-1}) \|}\\
  & < & (\beta)^{M-1} \| Y^+ \|_F \left( \beta +   \frac{\lambda^*_{M-1} + \mu_{max}(K_{\hat{Y}_{M-1}P^T_{M-1}} )}{\lambda^*_{M-1} + \mu_{min}(K_{\hat{Y}_{M-1}P^T_{M-1}} )} (\sqrt{N}-1) \right). 
 \end{IEEEeqnarray}  
Since $ \mu_{min}(K_{\hat{Y}_{M-1}P^T_{M-1}} ) > 0$, $\mu_{max}(K_{\hat{Y}_{M-1}P^T_{M-1}})  <  N$, $\lambda^*_{M-1} > 0$, and $0<\beta<1$, we have 
\begin{IEEEeqnarray}{rCCCl}
0&<& \left( \beta +   \frac{\lambda^*_{M-1} + \mu_{max}(K_{\hat{Y}_{M-1}P^T_{M-1}} )}{\lambda^*_{M-1} + \mu_{min}(K_{\hat{Y}_{M-1}P^T_{M-1}} )} (\sqrt{N}-1) \right) & < & 1 + \frac{\lambda^*_{M-1} + N}{\lambda^*_{M-1}} (\sqrt{N}-1).
 \end{IEEEeqnarray} 
Thus,
\begin{IEEEeqnarray}{rCCCCCl}
0&\leq&\frac{\| \hat{y}^{i,M-1} - \mathcal{A}_{\hat{Y}_{M-1},n}(\hat{y}^{i,M-1}) \| }{(\beta)^{M-1}} & < & \| Y^+ \|_F \left( 1 + \frac{\lambda^*_{M-1} + N}{\lambda^*_{M-1}} (\sqrt{N}-1) \right) & < & \infty
 \end{IEEEeqnarray}  
Further, it is observed from (\ref{eq_300120231935}) that
\begin{IEEEeqnarray}{rCl}
\lim_{M \to \infty} (\beta)^{M-1} & = & 0,\mbox{ and thus}\\
\label{eq_070220230947}\lim_{M \to \infty} \| \hat{y}^{i,M-1} - \mathcal{A}_{\hat{Y}_{M-1},n}(\hat{y}^{i,M-1}) \|  & = & 0.
 \end{IEEEeqnarray}  
Since (\ref{eq_070220230947}) holds for all $i \in \{1,2,\cdots,N \}$, we must have
\begin{IEEEeqnarray}{rCl}
 \lim_{M \to \infty} \sum_{i=1}^N \| \hat{y}^{i,M-1} - \mathcal{A}_{\hat{Y}_{M-1},n}(\hat{y}^{i,M-1}) \|  & = & 0.
 \end{IEEEeqnarray}   
 Hence, the result is proved.

\vskip 0.2in
\bibliography{sample}
\bibliographystyle{theapa}

\end{document}